\documentclass{article}
\PassOptionsToPackage{authoryear,round}{natbib}
\usepackage[preprint]{neurips_2026}
\usepackage{jev_presentation}
\usepackage[utf8]{inputenc}
\usepackage[T1]{fontenc}
\usepackage{latexsym}
\usepackage{microtype}
\usepackage{inconsolata}
\usepackage{graphicx}
\usepackage{xcolor}
\usepackage{colortbl}
\definecolor{tableHeader}{HTML}{E0E0F4}
\definecolor{jevRow}{HTML}{D6EFF6}
\definecolor{bestCell}{HTML}{B0B0E3}
\definecolor{secondCell}{HTML}{C9DAF3}

\newcommand{\bestscore}[1]{\cellcolor{bestCell}\textbf{#1}}
\newcommand{\secondscore}[1]{\cellcolor{secondCell}\underline{#1}}
\newcommand{\modelsep}{\specialrule{0.35pt}{1.8pt}{1.8pt}}
\definecolor{linkblue}{RGB}{0,90,181}
\usepackage{booktabs}
\usepackage{float}
\usepackage{multirow}
\usepackage{amsmath,amssymb}
\usepackage{tabularx}
\usepackage{enumitem}
\usepackage{url}
\usepackage[colorlinks=true,linkcolor=linkblue,citecolor=linkblue,urlcolor=linkblue]{hyperref}
\usepackage{fontawesome5}
\usepackage{geometry}
\usepackage[most]{tcolorbox}
\csname usepgflibrary\endcsname{svg.path}
\newcommand{\E}{\mathrm{E}}
\newcommand{\C}{\mathrm{C}}
\newcommand{\N}{\mathrm{N}}
\newcommand{\I}{\mathbf{1}}
\newcommand{\jevauthorblock}{%
  \begin{minipage}[t]{\dimexpr\textwidth-2\tabcolsep\relax}
  \centering\normalfont
  \setlength{\parskip}{0pt}
  \mbox{\textbf{Fan Zhang}\textsuperscript{1,2}}\quad
  \mbox{\textbf{Yankai Chen}\textsuperscript{2,3}}\quad
  \mbox{\textbf{Zhuohan Xie}\textsuperscript{2}}\quad
  \mbox{\textbf{Yixi Zhou}\textsuperscript{4}}\quad
  \mbox{\textbf{Sijia Peng}\textsuperscript{5}}\quad
  \mbox{\textbf{Lei Fan}\textsuperscript{6}}\quad
  \mbox{\textbf{Xinhua Ji}\textsuperscript{7}}\quad
  \mbox{\textbf{Cunyuan Zheng }\textsuperscript{8}}\quad
  \mbox{\textbf{Huangyong Shan}\textsuperscript{9,11}}\quad
  \mbox{\textbf{Philip S. Yu}\textsuperscript{10}}\quad
  \mbox{\textbf{Xue Liu}\textsuperscript{2,3}}\quad
  \mbox{\textbf{Yu Chen}\textsuperscript{1}}\quad
  \mbox{\textbf{Preslav Nakov}\textsuperscript{2}}\quad
  \mbox{\textbf{Songwei He}\textsuperscript{9,11}}\par
  \vspace{0.4em}
  {\small
  \mbox{\textsuperscript{1}The University of Tokyo}\quad
  \mbox{\textsuperscript{2}MBZUAI}\quad
  \mbox{\textsuperscript{3}McGill University}\quad
  \mbox{\textsuperscript{4}Hong Kong Baptist University}\quad
  \mbox{\textsuperscript{5}Fudan University}\quad
  \mbox{\textsuperscript{6}University of Illinois Urbana-Champaign}\quad
  \mbox{\textsuperscript{7}UCloud}\quad
  \mbox{\textsuperscript{8}Columbia University}\quad
  \mbox{\textsuperscript{9}The University of Hong Kong}\quad
  \mbox{\textsuperscript{10}University of Illinois Chicago}\quad
  \mbox{\textsuperscript{11}Quantell Capital}\par}
  \vspace{0.2em}
  {\small\texttt{\{fan.zhang, yankai.chen, zhuohan.xie, preslav.nakov\}@mbzuai.ac.ae}\par}
  \end{minipage}%
}
\author{\jevauthorblock}
\hypersetup{
  pdftitle={Same Scores, Different Decisions: Evaluating JEV and Language Models for Legal Document Understanding},
  pdfauthor={Fan Zhang, Yankai Chen, Zhuohan Xie, Yixi Zhou, Sijia Peng, Lei Fan, Xinhua Ji, Cunyuan Zheng, Huangyong Shan, Philip S. Yu, Xue Liu, Yu Chen, Preslav Nakov, Songwei He}
}

\title{Same Scores, Different Decisions: Evaluating JEV and Language Models for Legal Document Understanding}
\begin{document}
\thispagestyle{empty}
\def\JevCoverIntegrated{1}
\input{cover}
\pagestyle{jevbody}
\suppressfloats[t]

\section{Introduction}
Contract review requires several judgments about one document: whether information is confidential, whether copies may be made, and what obligations remain after termination~\citep{koreeda2021contractnli,hendrycks2021cuad}. These judgments can be requested jointly or separately, while their intended answers remain tied to the contract. The shared document supports several fixed, annotated judgments, making it possible to change their presentation while holding the underlying evidence and questions constant. Jev's shared-state, structured-choice interface and the joint or separate prompts used by language models offer different ways to obtain these judgments. A model's practical value therefore depends on the cost and time required to obtain its answers and their correctness across request configurations, beyond accuracy in a single configuration.

\begin{figure}[t]
\centering

  \includegraphics[width=\textwidth]{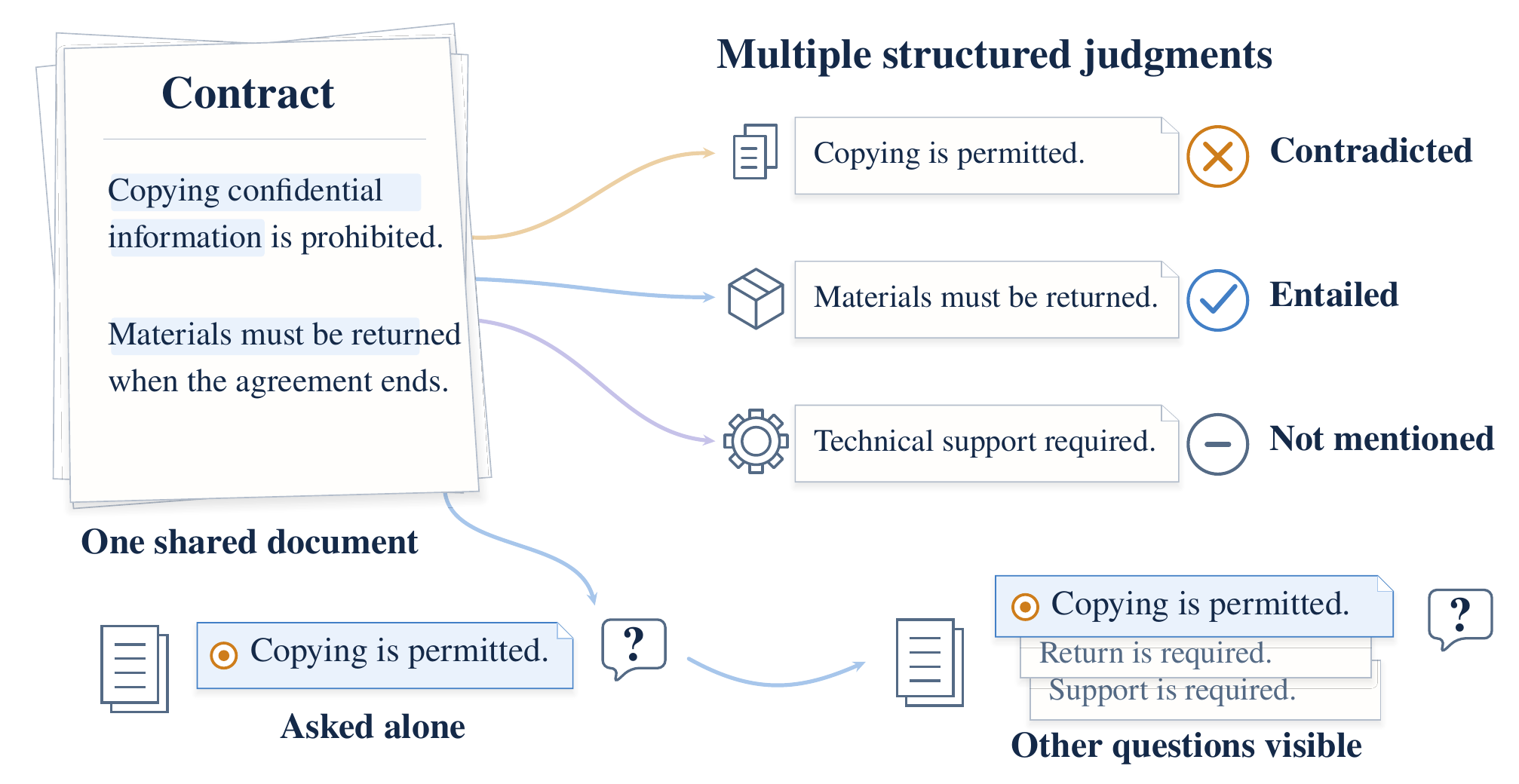}

\caption[Contract inference and correctness under changing request configurations.]{A fixed contract judgment should remain correct when other hypotheses become visible. The illustrative contract supports several structured judgments, matching the shared-document workload studied here. Copying refers to confidential information; the displayed labels are reference answers for the example, not model predictions. Below, the same target is requested alone or alongside other visible hypotheses; the contract and requested target label remain fixed. The question marks motivate testing correctness across request configurations rather than judging correctness from a single evaluation.}
\label{fig:example}
\end{figure}

One difficulty is that aggregate accuracy conceals which judgments change. Behavioral testing examines individual predictions~\citep{ribeiro2020checklist}, while studies of batching, position, demonstration order, and prompt formatting show request sensitivity~\citep{lin2024batchprompt,lu2022fantastically,sclar2024quantifying}. A recent preprint likewise shows that similar ranking quality can coexist with different downstream decisions~\citep{frohmann2026equal}. For judgments sharing one contract, corrections and regressions can cancel each other out even when many decisions differ. Figure~\ref{fig:example} introduces the task and the underlying evaluation question: when the contract and target statement stay fixed, does the model still answer correctly if other hypotheses become visible?

A second difficulty is that agreement does not establish correctness: a model may repeat an incorrect label across every condition. Variation can also arise when an unchanged request is repeated. Evaluation must therefore score persistent answers against gold labels and use unchanged repetitions as a reference for variability. This distinction also matters when deciding whether to pay for additional calls. Answer aggregation can improve reasoning performance~\citep{wang2023selfconsistency}, but repetitions that preserve the same wrong answer cannot repair it. Examining these behaviors alongside cost follows HELM's multidimensional perspective~\citep{liang2023helm}.

These considerations motivate three questions. \textbf{1)} Which judgments change across request conditions, and how much of that change is hidden by aggregate accuracy? \textbf{2)} How do repeated answers relate to correctness, and can voting improve the final judgment? \textbf{3)} What trade-offs arise among baseline accuracy, cost, and response time? We preserve targets and gold labels across conditions, retain invalid responses in accuracy denominators, and estimate uncertainty at the contract level.

\begingroup
The initial ten-model study evaluates 123 test contracts and repeated requests on 30 fixed targets. A four-model follow-up compares fixed and varied requests at equal call counts and extends repeated-target coverage to the remaining 93 contracts. We then analyze voting and error profiles using these recorded responses. Separate development experiments examine grouping and inference settings.
\endgroup

\begingroup
Our contributions are:
\begin{itemize}[leftmargin=*,itemsep=2pt,topsep=3pt,parsep=0pt,partopsep=0pt]
\item An all-model decomposition of answer changes into corrections, regressions, and wrong-label changes, alongside unchanged-request repetition (Section~\ref{sec:rq-changes}).
\item A same-target analysis of accuracy, consistent correctness, and answer consistency, with an equal-call control, expanded target coverage, and a voting analysis that tests whether additional calls improve decisions (Section~\ref{sec:rq-correctness}).
\item A joint evaluation of baseline accuracy, cost, and response time for contract judgments (Section~\ref{sec:rq-resources}).
\end{itemize}
\endgroup

\section{Task and Evaluation Design}
\label{sec:task}
For a contract $d$, let $H_d=\{h_1,\ldots,h_{17}\}$ denote the original hypothesis set, with gold labels $y_{dh}\in\{\E,\C,\N\}$ for entailment, contradiction, and not mentioned. A model returns a label $\hat y_{dh}$ or no valid prediction. We evaluate ContractNLI's classification component; evidence-span extraction is outside the scope of this study. Every prediction is scored against the existing annotations, without new human labeling or model-based judging.

\paragraph{Baseline accuracy and validity.}
The baseline requests all 17 labels jointly, once per contract. A valid classification response contains exactly the requested IDs and an allowed label for each. Adherence to requested output order is checked separately; an out-of-order label map can still supply valid predictions. Accuracy is computed over all intended judgments, with invalid responses counted as incorrect. We also report Macro-F1 and valid-response coverage. Responses that exhaust the generation allowance before producing final labels remain failures of the evaluated configuration (Appendix~\ref{app:budget}).

\paragraph{Controlled request conditions.}
The repeated-target study fixes one target hypothesis for each selected test contract. Four conditions preserve its contract and gold label:
\begin{description}
\item[A.] Only the target is visible; only its label is requested.
\item[B.] All 17 hypotheses are visible; only the target label is requested.
\item[C.] The same catalog is visible; all 17 labels are requested, target first.
\item[D.] The same catalog is visible; all 17 labels are requested, target last.
\end{description}
The catalog order is fixed in B/C/D. A--B changes visible hypotheses at a fixed requested answer set; B--C changes output workload and structure at fixed visible content; C--D changes requested output order and structure. Each condition is repeated three times. Primary scoring uses only the fixed target; the extra labels in C/D do not increase the number of primary targets.

\paragraph{Correctness across all twelve responses.}
\emph{Consistent correctness} requires the correct label in all four conditions and all three repeats (All 12 correct). Targets fall into four mutually exclusive categories: all twelve labels correct; the same wrong label throughout; twelve valid labels that change; or at least one invalid response. All selected targets remain in the denominator. \emph{Answer consistency} combines the first two categories. Within-condition disagreement measures variation under unchanged requests.

\begingroup
\emph{Mean target accuracy} averages correctness over all intended target responses, counting invalid responses as incorrect. It uses the same targets as consistent correctness: the former weights responses equally, while the latter requires all twelve responses for each target to be correct.
\endgroup

\paragraph{Paired changes.}
We pair responses by contract, target, and repeat index across conditions. Jointly valid pairs separate corrections, regressions, and switches between different wrong labels. Changed answers count all three, whereas net correctness subtracts regressions from corrections. This separates changes hidden by the net score from unchanged decisions. Appendix~\ref{app:distinct-targets} gives the identity, valid-pair denominators, and distinct-target counts.

\paragraph{Voting and error diagnostics.}
A retrospective analysis of the completed follow-up compares single answers with strict-majority voting. We enumerate subsets within each target, score abstentions as incorrect, and resample contracts for paired uncertainty. Three-call voting uses either identical B requests or three different conditions. Error diagnostics on the additional 93 targets group outcomes by gold class, hypothesis type, and contract-length thirds. The analysis rules were fixed before computing these summaries; no new model calls or annotations are added (Appendices~\ref{app:voting}--\ref{app:diagnostics}).

\section{Data and Experimental Protocol}
\label{sec:protocol}
\subsection{Data and uncertainty}
The official ContractNLI test split contains 123 contracts and 2,091 judgments: 968 entailments, 220 contradictions, and 903 not-mentioned labels~\citep{koreeda2021contractnli}. We preserve full contract text, all original hypotheses, and their annotations. Evidence spans are excluded from model inputs. The initial repeated-target panel selects 30 test contracts and one target per contract using seed 20260922 and document-specific target seeds, independently of labels and model predictions. 

We use 5,000 whole-contract bootstrap resamples (seed 20260922), preserving paired observations across models. Intervals are descriptive and unadjusted for multiple comparisons~\citep{dror2018hitchhiker}. Repeats add observations, not independent contracts.

\begingroup
Test inputs, target selection, scoring rules, and each model configuration are fixed before the corresponding inference. Sonnet, Haiku, and Terra extend the model selection after inspection of the initial results, using the same test samples and scoring rules. Development experiments use separate contracts (Appendix~\ref{app:development-protocol}).
\endgroup

\paragraph{Follow-up and development samples.}
For Jev, Gemini-3.5-Flash, Terra, and Sonnet, the follow-up compares twelve identical B requests with A/B/C/D three times each on the original 30 targets. Both arms are newly collected, randomly paired and interleaved, with six pairs per contract in each order. It also evaluates one seeded target on each of the remaining 93 test contracts. The 7,344 new primary calls are separate from the initial study; paired intervals use seed 20260929 (Appendix~\ref{app:request-controls}). Development controls use separate contracts: 30 for grouping and ten with three targets each for repetition.

\subsection{Models and inference}
We compare Jev-1.13.0 with Qwen-3.5-4B and Qwen-3.5-9B, Gemini-3.5-Flash and Gemini-3.1-Pro-Preview, GPT-5.6-Luna and GPT-5.6-Terra, GPT-6-Astra, Claude-Sonnet-5, and Claude-Haiku-4.5. Jev receives the contract as shared state and each requested hypothesis as a native Choice question. Generative models receive semantically aligned classification instructions and return a JSON label map. Each comparison uses a fixed prompt template. Appendix~\ref{app:repro} gives the classification instruction.

Inference settings are fixed per model (Appendix~\ref{app:prices}). Both Qwen test configurations enable thinking and share a 32,768-token allowance between reasoning and final labels, using one GPU each. Their thinking-disabled development configurations are analyzed separately. Initial-study conditions and repeats are shuffled within each contract, with sequential requests and no retries. Follow-up primary scores likewise use first attempts; service retries are reported separately (Appendix~\ref{app:request-controls}). Jev's C/D contrast changes native question-key order; the language-model contrast changes requested JSON output order.

\subsection{Cost and timing}
Quality and cost are measured over the same requests, including failed attempts. For Luna and Astra, API costs use provider-reported charges; for other API models, we multiply recorded token usage by the public input, output, and cache prices applicable to the evaluation. When usage is missing, its cost remains unknown, and we report the known subtotal and coverage separately. Baseline cost is averaged over the 123 baseline contracts; repeated-target cost is reported separately for its 360 requests.

\begingroup
For each local model, rental-equivalent baseline cost is
\begin{equation*}
\widehat C_{\mathrm{GPU}}=\frac{r}{123\times3600}\sum_{d=1}^{123}t_d,
\end{equation*}
where $t_d$ is the measured request time and $r$ is \$2.00 per GPU-hour. Each model uses one GPU. This rental equivalent values observed inference time, excluding loading, idle allocation, storage, transfer, and taxes. Appendix~\ref{app:development-protocol} documents the hardware and reference rate; Appendix~\ref{app:prices} gives API accounting.
\endgroup

\begingroup
Response time measures elapsed client time for the prediction call, including connection, transfer, service processing, and validation. The original seven configurations also time condition-payload construction; the three added configurations prepare it beforehand. The initial-study timing comparisons use the same client machine, one in-flight request per model, and no retries. Medians and 95th percentiles (P95) include failed attempts. Appendix~\ref{app:timing} details the timing scope.
\endgroup

\section{Results on the Official Test Split}
\label{sec:test}
The initial ten-model study contains a 123-contract baseline and 30 repeated targets. We first analyze answer changes, then correctness across repetitions and the four-model follow-up, and finally baseline resource trade-offs. Failed first attempts remain in primary denominators; Appendix~\ref{app:testdetails} reports initial condition-level scores.

\subsection{RQ1: Changes hidden by aggregate accuracy}
\label{sec:rq-changes}
\begin{table}[!t]
\centering\small
\setlength{\tabcolsep}{3pt}
\renewcommand{\arraystretch}{1.12}
\begin{tabularx}{\textwidth}{p{1.68in}*{6}{>{\centering\arraybackslash}X}}
\toprule
\rowcolor{tableHeader}
\textbf{Model} & \multicolumn{2}{c}{Visible questions} & \multicolumn{2}{c}{Requested answers} & \multicolumn{2}{c}{Requested order} \\
\rowcolor{tableHeader}
& \multicolumn{2}{c}{A--B} & \multicolumn{2}{c}{B--C} & \multicolumn{2}{c}{C--D} \\
\rowcolor{tableHeader}
& \shortstack{Changed\\answers} & \shortstack{$\Delta$\\correct} & \shortstack{Changed\\answers} & \shortstack{$\Delta$\\correct} & \shortstack{Changed\\answers} & \shortstack{$\Delta$\\correct} \\
\midrule
\rowcolor{jevRow}
\makebox[0.9em][c]{\raisebox{-0.12em}{\includegraphics[width=0.9em,height=0.9em,keepaspectratio]{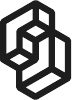}}}\hspace{0.3em}Jev-1.13.0 & 4/90 & +4 & 1/90 & -1 & 0/90 & 0 \\
\makebox[0.9em][c]{\raisebox{-0.12em}{\includegraphics[width=0.9em,height=0.9em,keepaspectratio]{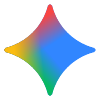}}}\hspace{0.3em}Gemini-3.5-Flash & 7/90 & +1 & 3/90 & +1 & 6/90 & -4 \\
\makebox[0.9em][c]{\raisebox{-0.12em}{\includegraphics[width=0.9em,height=0.9em,keepaspectratio]{figures/icons/gemini.pdf}}}\hspace{0.3em}Gemini-3.1-Pro-Preview & 7/89 & -1 & 7/90 & +1 & 3/90 & -3 \\
\makebox[0.9em][c]{\raisebox{-0.12em}{\includegraphics[width=0.9em,height=0.9em,keepaspectratio]{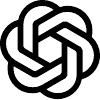}}}\hspace{0.3em}GPT-5.6-Luna & 8/90 & +2 & 8/90 & +2 & 11/90 & -3 \\
\makebox[0.9em][c]{\raisebox{-0.12em}{\includegraphics[width=0.9em,height=0.9em,keepaspectratio]{figures/icons/openai.pdf}}}\hspace{0.3em}GPT-5.6-Terra & 7/90 & +6 & \cellcolor{bestCell}\textbf{7/90} & \cellcolor{bestCell}\textbf{0} & 10/90 & -4 \\
\makebox[0.9em][c]{\raisebox{-0.12em}{\includegraphics[width=0.9em,height=0.9em,keepaspectratio]{figures/icons/openai.pdf}}}\hspace{0.3em}GPT-6-Astra & 4/90 & -2 & 3/90 & +1 & 3/90 & +3 \\
\makebox[0.9em][c]{\raisebox{-0.12em}{\includegraphics[width=0.9em,height=0.9em,keepaspectratio]{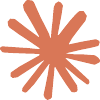}}}\hspace{0.3em}Claude-Haiku-4.5 & 8/90 & +5 & 6/90 & +3 & 14/90 & -10 \\
\makebox[0.9em][c]{\raisebox{-0.12em}{\includegraphics[width=0.9em,height=0.9em,keepaspectratio]{figures/icons/claude.pdf}}}\hspace{0.3em}Claude-Sonnet-5 & 5/90 & +1 & 6/90 & +4 & 3/90 & +1 \\
\makebox[0.9em][c]{\raisebox{-0.12em}{\includegraphics[width=0.9em,height=0.9em,keepaspectratio]{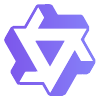}}}\hspace{0.3em}Qwen-3.5-4B & \cellcolor{bestCell}\textbf{9/90} & \cellcolor{bestCell}\textbf{0} & 9/88 & +6 & 14/87 & -3 \\
\makebox[0.9em][c]{\raisebox{-0.12em}{\includegraphics[width=0.9em,height=0.9em,keepaspectratio]{figures/icons/qwen.pdf}}}\hspace{0.3em}Qwen-3.5-9B & 15/90 & +1 & 7/90 & -1 & 12/88 & +2 \\
\bottomrule
\end{tabularx}
\caption{Answer changes can leave the net correct count unchanged. Changed answers reports changed labels over jointly valid pairs. $\Delta$ correct is corrections minus regressions within those pairs. Each contrast matches 30 targets across three repeats, giving 90 planned pairs. Dark purple and bold mark changed answers with zero net change.}
\label{tab:test-sensitivity-summary}
\end{table}
Table~\ref{tab:test-sensitivity-summary} compares changed answers with the net change in correct answers for all ten models across the three request contrasts. Corrections and regressions coexist in 22 of the 30 model--contrast comparisons. Thus, cancellation is not confined to one model: improvements on some judgments offset losses on others, reducing the change visible in aggregate accuracy. Table~\ref{tab:testchanges} reports all directional counts and valid-pair denominators.

\paragraph{Complete and partial cancellation.}
Qwen-3.5-4B has four corrections, four regressions, and one wrong-label switch between A and B: accuracy stays at 68/90 despite nine changed answers. Terra has three corrections, three regressions, and one wrong-label switch between B and C, preserving 75/90 correct answers. Qwen-3.5-9B changes 15/90 answers between A and B but gains only one correct answer, through eight corrections and seven regressions. Net changes in correctness therefore understate how many responses change.

\begingroup
Answers also differ between requested output orders, with variation across configurations. Haiku changes 14 of 90 paired responses from C to D, including twelve regressions and two corrections. Its accuracy falls by 11.11 percentage points (descriptive paired 95\% interval: $-23.33$ to $-1.11$). Jev shows no C--D changes under its native question-key intervention. Terra returns lexicographically ordered label maps in every C/D response, despite the requested target-first or target-last sequence. Its contrast therefore compares different order instructions with the same observed key order. Appendix~\ref{app:terra-example} presents a paired JSON example and separates classification validity from order adherence.
\endgroup

\paragraph{Variation without a changed condition.}
Some affected targets also vary under unchanged requests. Qwen-3.5-4B's nine A--B changes involve six distinct targets; all six vary within A or B. Terra's seven B--C changes involve five targets, all also varying within a condition. These overlaps motivate the equal-call control below. Appendix~\ref{app:distinct-targets} reports all ten models; changed responses are not independent target replications.

\subsection{RQ2: Correctness across repeated conditions}
\label{sec:rq-correctness}
\begin{table*}[t]

\centering
\small
\begin{tabularx}{\textwidth}{Xccccc}
\toprule
\rowcolor{tableHeader}
\textbf{Model} & \shortstack{Mean target\\accuracy (\%) $\uparrow$} & \shortstack{All 12\\correct $\uparrow$} & \shortstack{Same\\wrong} & \shortstack{Changed\\valid} & \shortstack{Any\\invalid} \\
\midrule
\rowcolor{jevRow}
\makebox[0.9em][c]{\raisebox{-0.12em}{\includegraphics[width=0.9em,height=0.9em,keepaspectratio]{figures/icons/jev.pdf}}}\hspace{0.3em}Jev-1.13.0 & 79.44 & \secondscore{23/30} & 5 & 2 & 0 \\
\addlinespace[3pt]
\makebox[0.9em][c]{\raisebox{-0.12em}{\includegraphics[width=0.9em,height=0.9em,keepaspectratio]{figures/icons/gemini.pdf}}}\hspace{0.3em}Gemini-3.5-Flash & 82.50 & 21/30 & 4 & 5 & 0 \\
\makebox[0.9em][c]{\raisebox{-0.12em}{\includegraphics[width=0.9em,height=0.9em,keepaspectratio]{figures/icons/gemini.pdf}}}\hspace{0.3em}Gemini-3.1-Pro-Preview & \secondscore{85.28} & 22/30 & 2 & 5 & 1 \\
\addlinespace[3pt]
\makebox[0.9em][c]{\raisebox{-0.12em}{\includegraphics[width=0.9em,height=0.9em,keepaspectratio]{figures/icons/openai.pdf}}}\hspace{0.3em}GPT-5.6-Luna & 81.94 & 22/30 & 3 & 5 & 0 \\
\makebox[0.9em][c]{\raisebox{-0.12em}{\includegraphics[width=0.9em,height=0.9em,keepaspectratio]{figures/icons/openai.pdf}}}\hspace{0.3em}GPT-5.6-Terra & 80.56 & 20/30 & 2 & 8 & 0 \\
\makebox[0.9em][c]{\raisebox{-0.12em}{\includegraphics[width=0.9em,height=0.9em,keepaspectratio]{figures/icons/openai.pdf}}}\hspace{0.3em}GPT-6-Astra & 76.39 & 22/30 & 6 & 2 & 0 \\
\addlinespace[3pt]
\makebox[0.9em][c]{\raisebox{-0.12em}{\includegraphics[width=0.9em,height=0.9em,keepaspectratio]{figures/icons/claude.pdf}}}\hspace{0.3em}Claude-Haiku-4.5 & 81.94 & 19/30 & 2 & 9 & 0 \\
\makebox[0.9em][c]{\raisebox{-0.12em}{\includegraphics[width=0.9em,height=0.9em,keepaspectratio]{figures/icons/claude.pdf}}}\hspace{0.3em}Claude-Sonnet-5 & \bestscore{86.67} & \bestscore{24/30} & 1 & 5 & 0 \\
\addlinespace[3pt]
\makebox[0.9em][c]{\raisebox{-0.12em}{\includegraphics[width=0.9em,height=0.9em,keepaspectratio]{figures/icons/qwen.pdf}}}\hspace{0.3em}Qwen-3.5-4B & 77.50 & 17/30 & 2 & 8 & 3 \\
\makebox[0.9em][c]{\raisebox{-0.12em}{\includegraphics[width=0.9em,height=0.9em,keepaspectratio]{figures/icons/qwen.pdf}}}\hspace{0.3em}Qwen-3.5-9B & 79.17 & 18/30 & 2 & 8 & 2 \\
\bottomrule
\end{tabularx}
\caption{Mean target accuracy scores the same 30 test targets across four conditions and three repeats (360 judgments); invalid responses count wrong. All 12 correct requires every answer to be correct. Remaining targets have twelve identical wrong labels, twelve valid but changing labels, or an invalid response; invalidity takes precedence. The four outcome counts sum to 30.}
\label{tab:teststates}
\end{table*}
In the initial study, Sonnet keeps 24/30 targets correct throughout and Jev 23/30, followed by Gemini Pro, Luna, and Astra with 22 (Table~\ref{tab:teststates}). The Sonnet--Jev difference is one target, with a descriptive paired 95\% interval of $-10.00$ to 16.67 percentage points. Appendix~\ref{app:testdetails} gives all paired comparisons.

\paragraph{Similar means, different persistent errors.}
Jev answers 286/360 target responses correctly (79.44\%) and Qwen-3.5-9B 285/360 (79.17\%). Jev retains five more consistently correct targets (23 versus 18). Agreement also differs from correctness: Jev and Astra each repeat the same label on 28/30 targets, including five and six persistent errors, respectively. Sonnet repeats one label on only 25/30 targets but has just one persistent error.

\paragraph{Equal calls, different request configurations.}
\begin{table}[!htbp]

\centering\small
\setlength{\tabcolsep}{4.5pt}
\renewcommand{\arraystretch}{1.12}
\begin{tabularx}{\textwidth}{@{}p{0.30\textwidth}*{6}{>{\centering\arraybackslash}X}@{}}
\toprule
\rowcolor{tableHeader}
\textbf{Model} & \multicolumn{2}{c}{Accuracy (\%) $\uparrow$} & \multicolumn{2}{c}{All 12 correct $\uparrow$} & \multicolumn{2}{c}{Same answer} \\
\cmidrule(lr){2-3}\cmidrule(lr){4-5}\cmidrule(lr){6-7}
& Fixed & Varied & Fixed & Varied & Fixed & Varied \\
\midrule
\makebox[0.9em][c]{\raisebox{-0.12em}{\includegraphics[width=0.9em,height=0.9em,keepaspectratio]{figures/icons/jev.pdf}}}\hspace{0.3em}Jev-1.13.0 & 80.56 & 79.17 & \secondscore{24/30} & \bestscore{23/30} & 29/30 & 29/30 \\
\makebox[0.9em][c]{\raisebox{-0.12em}{\includegraphics[width=0.9em,height=0.9em,keepaspectratio]{figures/icons/gemini.pdf}}}\hspace{0.3em}Gemini-3.5-Flash & \bestscore{83.33} & \secondscore{83.33} & \bestscore{25/30} & \bestscore{23/30} & 30/30 & 27/30 \\
\makebox[0.9em][c]{\raisebox{-0.12em}{\includegraphics[width=0.9em,height=0.9em,keepaspectratio]{figures/icons/openai.pdf}}}\hspace{0.3em}GPT-5.6-Terra & 81.67 & 77.78 & 22/30 & 20/30 & 24/30 & 22/30 \\
\makebox[0.9em][c]{\raisebox{-0.12em}{\includegraphics[width=0.9em,height=0.9em,keepaspectratio]{figures/icons/claude.pdf}}}\hspace{0.3em}Claude-Sonnet-5 & \secondscore{82.50} & \bestscore{85.28} & \secondscore{24/30} & \secondscore{22/30} & 29/30 & 23/30 \\
\bottomrule
\end{tabularx}
\caption{Equal-call control on the same 30 targets. Fixed repeats B twelve times; Varied uses A/B/C/D three times each. Both arms are newly collected and interleaved. All 2,880 first attempts return valid answers. Same answer includes consistently wrong labels. Correctness is ranked within each arm; paired intervals appear in Table~\ref{tab:request-control-effects}.}
\label{tab:request-control}
\end{table}
The fresh paired control separates repeated sampling from a change in request mixture (Table~\ref{tab:request-control}). Gemini-3.5-Flash has 83.33\% accuracy in both arms, but consistently correct targets fall from 25 to 23. Sonnet gains 2.78 accuracy points while its consistently correct count falls from 24 to 22. All four models lose one or two consistently correct targets in the observed comparison; their paired intervals include zero. Answer consistency also falls for the three language models, most for Sonnet: $-20.00$ points (95\% interval: $-36.67$ to $-6.67$). Jev's answer-consistent count is unchanged. Thus a fixed call count does not make aggregate accuracy and persistent correctness interchangeable.

\paragraph{Additional contracts.}
\begin{table}[!htbp]

\centering\small
\setlength{\tabcolsep}{2pt}
\begin{tabularx}{\textwidth}{@{}p{0.255\textwidth}*{6}{>{\centering\arraybackslash}X}@{}}
\toprule
\rowcolor{tableHeader}
\textbf{Model} & \shortstack{Accuracy\\(\%) $\uparrow$} & \shortstack{All 12\\correct $\uparrow$} & \shortstack{Same\\answer} & \shortstack{Correct\\among same\\answers\\(\%) $\uparrow$} & \shortstack{Same\\wrong} & \shortstack{Any\\invalid} \\
\midrule
\makebox[0.9em][c]{\raisebox{-0.12em}{\includegraphics[width=0.9em,height=0.9em,keepaspectratio]{figures/icons/jev.pdf}}}\hspace{0.3em}Jev-1.13.0 & 75.27 & \secondscore{67/93} & 84/93 & 79.76 & 17 & 0 \\
\makebox[0.9em][c]{\raisebox{-0.12em}{\includegraphics[width=0.9em,height=0.9em,keepaspectratio]{figures/icons/gemini.pdf}}}\hspace{0.3em}Gemini-3.5-Flash & \secondscore{80.65} & 66/93 & 77/93 & 85.71 & 11 & 2 \\
\makebox[0.9em][c]{\raisebox{-0.12em}{\includegraphics[width=0.9em,height=0.9em,keepaspectratio]{figures/icons/openai.pdf}}}\hspace{0.3em}GPT-5.6-Terra & 80.02 & 64/93 & 72/93 & \secondscore{88.89} & 8 & 4 \\
\makebox[0.9em][c]{\raisebox{-0.12em}{\includegraphics[width=0.9em,height=0.9em,keepaspectratio]{figures/icons/claude.pdf}}}\hspace{0.3em}Claude-Sonnet-5 & \bestscore{84.68} & \bestscore{70/93} & 76/93 & \bestscore{92.11} & 6 & 1 \\
\bottomrule
\end{tabularx}
\caption{New targets from the remaining 93 test contracts, using first attempts. Accuracy uses 1,116 target responses per model. All 12 correct and Same answer count targets out of 93. Same wrong is a subset of Same answer. Correct among same answers is $100\times\mathrm{All\ 12\ correct}/\mathrm{Same\ answer}$, calculated on each model's unanimous targets. Invalid responses remain in the intended denominators and cannot count as unanimous. Best/second-best marks describe point estimates. Table~\ref{tab:request-expansion93} gives the service-retry results.}
\label{tab:request-expansion-main}
\end{table}
The new 93 targets show the distinction again (Table~\ref{tab:request-expansion-main}): Jev repeats one label on 84 targets, versus 76 for Sonnet, but fewer are correct throughout (67 versus 70). Seventeen of Jev's persistent labels are wrong, versus six for Sonnet. The Sonnet--Jev difference in consistently correct targets is three, with a descriptive paired interval reported in Appendix~\ref{app:request-controls}, alongside service retries and batch comparisons.

\paragraph{How often are unanimous answers correct?}
Among targets with twelve identical answers, the correct fraction is 67/84 (79.76\%) for Jev, 66/77 (85.71\%) for Gemini-3.5-Flash, 64/72 (88.89\%) for Terra, and 70/76 (92.11\%) for Sonnet. Jev has the largest number of unanimous targets, yet the lowest correct fraction within that set. These proportions describe each model's own unanimous targets. Agreement frequency and correctness among agreed answers therefore measure different aspects of reliability.

\subsubsection{Can repeated calls recover errors?}
\label{sec:voting}
\begingroup
Repeated calls are useful only if combining their answers improves the final decision. We evaluate strict-majority voting retrospectively from the completed four-model follow-up, without new inference. On the fresh 30-target control, Fixed samples three of twelve identical B responses; Varied samples three responses from three different conditions. We enumerate every eligible subset, rather than choosing a successful draw. A label needs two votes; a three-way tie supplies no decision and counts as incorrect. Single-call accuracy is the uniform mean over the corresponding twelve responses. Appendix~\ref{app:voting} gives the exact enumeration and paired uncertainty.
\begin{table}[!htbp]

\centering\small
\setlength{\tabcolsep}{4pt}
\begin{tabularx}{\textwidth}{Xcccccc}
\toprule
\rowcolor{tableHeader}
\textbf{Model} & \multicolumn{2}{c}{Fixed, 30 targets} & \multicolumn{2}{c}{Varied, 30 targets} & \multicolumn{2}{c}{Additional 93} \\
\cmidrule(lr){2-3}\cmidrule(lr){4-5}\cmidrule(lr){6-7}
& Single & Vote-3 & Single & Vote-3 & Single & Vote-3 \\
\midrule
\makebox[0.9em][c]{\raisebox{-0.12em}{\includegraphics[width=0.9em,height=0.9em,keepaspectratio]{figures/icons/jev.pdf}}}\hspace{0.3em}Jev-1.13.0 & 80.56 & 80.15 & 79.17 & 80.00 & 75.27 & 74.25 \\
\makebox[0.9em][c]{\raisebox{-0.12em}{\includegraphics[width=0.9em,height=0.9em,keepaspectratio]{figures/icons/gemini.pdf}}}\hspace{0.3em}Gemini-3.5-Flash & 83.33 & 83.33 & 83.33 & 84.44 & 80.65 & 81.00 \\
\makebox[0.9em][c]{\raisebox{-0.12em}{\includegraphics[width=0.9em,height=0.9em,keepaspectratio]{figures/icons/openai.pdf}}}\hspace{0.3em}GPT-5.6-Terra & 81.67 & 81.76 & 77.78 & 77.96 & 80.02 & 79.37 \\
\makebox[0.9em][c]{\raisebox{-0.12em}{\includegraphics[width=0.9em,height=0.9em,keepaspectratio]{figures/icons/claude.pdf}}}\hspace{0.3em}Claude-Sonnet-5 & 82.50 & 82.91 & 85.28 & 85.86 & 84.68 & 84.05 \\
\bottomrule
\end{tabularx}
\caption{Three-call voting gives small changes in expected target accuracy (\%; higher is better). Fixed samples three of twelve identical B requests; Varied and Additional 93 sample three different conditions from A/B/C/D. Single averages all twelve responses. Results enumerate all eligible subsets rather than selecting a favorable draw. Vote-3 requires two matching labels; ties count as incorrect. First-attempt failures remain in the additional sample. Paired intervals, five-call voting, and cost appear in Appendix~\ref{app:voting}.}
\label{tab:voting-main}
\end{table}

\paragraph{Small gains at additional cost.}
On 30 targets, voting changes Fixed accuracy by $-0.40$ to $+0.41$ percentage points across the four models. Gemini-3.5-Flash's twelve Fixed answers are identical for every target, so its vote cannot change a decision. Diverse-three voting changes Varied accuracy by $+0.19$ to $+1.11$ points. All eight paired gain intervals include zero. Within an arm, the expected three-call cost is exactly three times the single-call mean; the two arms have different costs because they request different outputs. For Terra, Fixed vote-3 costs \$0.008599 per target decision, whereas Varied vote-3 costs \$0.020207 (Table~\ref{tab:voting-cost}).

\paragraph{The additional sample does not establish a voting benefit.}
On the other 93 targets, diverse-three voting changes accuracy by $-1.02$ points for Jev, $+0.36$ for Gemini-3.5-Flash, $-0.65$ for Terra, and $-0.63$ for Sonnet. Each paired interval includes zero. The separate service-retry view and five-call sensitivity analysis give the same practical result: these recorded repetitions do not establish that extra calls improve target accuracy. Voting can correct a minority of wrong answers, but can also overrule an isolated correct answer; the net gain reflects both effects.
\endgroup

\subsubsection{Where do persistent errors occur?}
\label{sec:error-diagnostics}
\begingroup
The additional 93 targets contain 38 entailments, 13 contradictions, and 42 not-mentioned labels. Class-specific results reveal differences concealed by their aggregate means (Figure~\ref{fig:error-diagnostics}). Jev scores 92.31\% on contradictions but 59.52\% on not-mentioned targets. Sonnet scores 66.03\% and 84.92\%, respectively. These are different error profiles, rather than a uniform advantage across classes. Equal weighting of the three gold classes also changes the comparison: Terra's class-macro accuracy is 81.21\%, versus Sonnet's 80.58\%, while Sonnet has the higher ordinary mean (84.68\% versus 80.02\%).

\begin{figure}[!htbp]
\centering
\includegraphics[width=\textwidth]{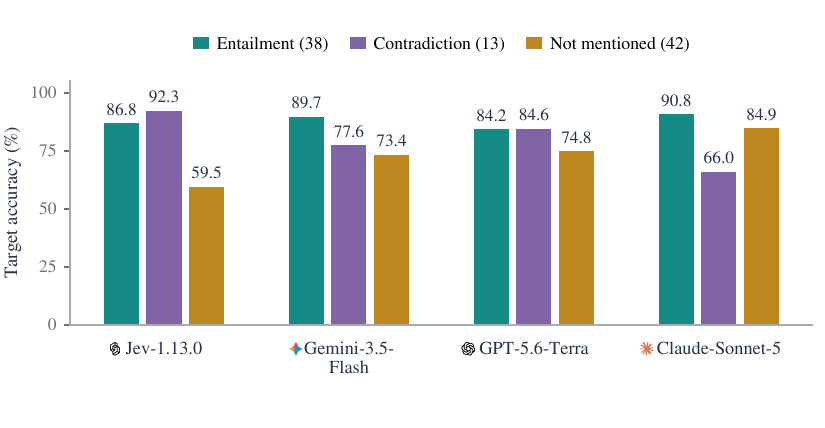}
\caption{Model rankings differ across gold classes. Bars show mean target accuracy on the additional 93 contracts, averaging twelve responses per target and counting invalid responses as incorrect. Parentheses give the number of targets in each gold class. Each target is one independent contract.}
\label{fig:error-diagnostics}
\end{figure}

\paragraph{Errors that voting cannot recover.}
Table~\ref{tab:diagnostic-error-share} reports the concentration of errors on same-wrong targets. These targets account for 73.91\% of Jev's wrong or invalid response slots, compared with 61.11\% for Gemini-3.5-Flash, 43.05\% for Terra, and 42.11\% for Sonnet. Every eligible voting subset on these targets repeats the wrong label. Thirteen of Jev's seventeen persistent errors have a not-mentioned gold label: eight repeatedly predict contradiction and five entailment. Agreement therefore often reflects a repeated error about whether the contract establishes a claim. Appendix~\ref{app:diagnostics} reports the complete confusions and an actual contract example, scored against the released annotation.

\paragraph{Hypothesis and length diagnostics.}
All 17 hypothesis groups are reported, including their small denominators of 2--10 contracts. Jev's 17 same-wrong targets span eight hypothesis types; its no-rights hypothesis has four such errors among six targets. Length groups show a different pattern across models: Jev's accuracy falls from 81.18\% in the shortest third to 69.62\% in the longest, whereas the three language models have their lowest mean in the middle or longest group without a shared monotonic pattern. These groups differ in their gold labels and hypothesis types; the tables describe where errors occur without assigning their cause to length.
\endgroup

\subsection{RQ3: Cost, response time, and accuracy}
\label{sec:rq-resources}
\begin{figure*}[t]
\centering
\includegraphics[width=0.92\textwidth]{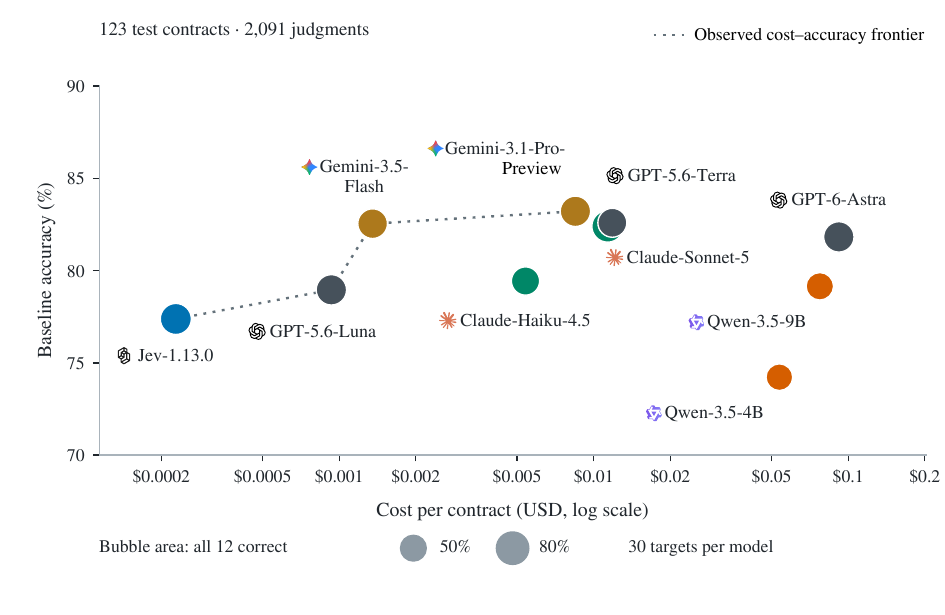}
\caption[Official-test cost, accuracy, and correctness across all twelve responses.]{Cost, baseline accuracy, and consistent correctness rank the models differently. Coordinates summarize 123 test contracts (2,091 judgments); bubble area gives the fraction of 30 fixed targets answered correctly in all four conditions and three repeats. Jev occupies the lowest-cost end and Gemini-3.1-Pro-Preview has the highest baseline accuracy point estimate. The dotted frontier joins nondominated cost--accuracy points under the accounting in Section~\ref{sec:protocol}; its segments connect measured configurations. Tables~\ref{tab:testbaseline} and~\ref{tab:testanchor} report uncertainty.}
\label{fig:jev-main-results}
\end{figure*}

Figure~\ref{fig:jev-main-results} and Table~\ref{tab:testbaseline} compare baseline accuracy with cost and response time.

\begin{table*}[t]
\centering

\small
\setlength{\tabcolsep}{3pt}
\begin{tabularx}{\textwidth}{Xcccccc}
\toprule
\rowcolor{tableHeader}
\textbf{Model} & \shortstack{Baseline\\accuracy\\(\%) $\uparrow$} & \shortstack{95\%\\interval} & \shortstack{Macro-F1\\(\%) $\uparrow$} & \shortstack{Cost\\(USD/contract) $\downarrow$} & \shortstack{Median\\time (s) $\downarrow$} & \shortstack{Valid\\calls} \\
\midrule
\rowcolor{jevRow}
\makebox[0.9em][c]{\raisebox{-0.12em}{\includegraphics[width=0.9em,height=0.9em,keepaspectratio]{figures/icons/jev.pdf}}}\hspace{0.3em}Jev-1.13.0 & 77.38 & [75.99, 78.81] & 72.23 & \bestscore{0.000228} & \bestscore{1.24} & 123/123 \\
\addlinespace[3pt]
\makebox[0.9em][c]{\raisebox{-0.12em}{\includegraphics[width=0.9em,height=0.9em,keepaspectratio]{figures/icons/gemini.pdf}}}\hspace{0.3em}Gemini-3.5-Flash & 82.54 & [81.11, 83.98] & 76.96 & 0.001353 & \secondscore{1.60} & 123/123 \\
\makebox[0.9em][c]{\raisebox{-0.12em}{\includegraphics[width=0.9em,height=0.9em,keepaspectratio]{figures/icons/gemini.pdf}}}\hspace{0.3em}Gemini-3.1-Pro-Preview & \bestscore{83.21} & [81.68, 84.79] & \bestscore{77.71} & 0.008474 & 3.73 & 123/123 \\
\addlinespace[3pt]
\makebox[0.9em][c]{\raisebox{-0.12em}{\includegraphics[width=0.9em,height=0.9em,keepaspectratio]{figures/icons/openai.pdf}}}\hspace{0.3em}GPT-5.6-Luna & 78.96 & [77.19, 80.63] & 73.77 & \secondscore{0.000931} & 1.74 & 123/123 \\
\makebox[0.9em][c]{\raisebox{-0.12em}{\includegraphics[width=0.9em,height=0.9em,keepaspectratio]{figures/icons/openai.pdf}}}\hspace{0.3em}GPT-5.6-Terra & \secondscore{82.59} & [81.16, 84.03] & \secondscore{77.23} & 0.011839 & 11.61 & 123/123 \\
\makebox[0.9em][c]{\raisebox{-0.12em}{\includegraphics[width=0.9em,height=0.9em,keepaspectratio]{figures/icons/openai.pdf}}}\hspace{0.3em}GPT-6-Astra & 81.83 & [80.20, 83.36] & 76.55 & 0.091890 & 20.94 & 123/123 \\
\addlinespace[3pt]
\makebox[0.9em][c]{\raisebox{-0.12em}{\includegraphics[width=0.9em,height=0.9em,keepaspectratio]{figures/icons/claude.pdf}}}\hspace{0.3em}Claude-Haiku-4.5 & 79.44 & [77.81, 81.11] & 74.46 & 0.005396 & 5.65 & 123/123 \\
\makebox[0.9em][c]{\raisebox{-0.12em}{\includegraphics[width=0.9em,height=0.9em,keepaspectratio]{figures/icons/claude.pdf}}}\hspace{0.3em}Claude-Sonnet-5 & 82.40 & [80.97, 83.84] & 76.38 & 0.011352 & 3.27 & 123/123 \\
\addlinespace[3pt]
\makebox[0.9em][c]{\raisebox{-0.12em}{\includegraphics[width=0.9em,height=0.9em,keepaspectratio]{figures/icons/qwen.pdf}}}\hspace{0.3em}Qwen-3.5-4B & 74.22 & [70.59, 77.38] & 71.67 & 0.053653 & 86.87 & 117/123 \\
\makebox[0.9em][c]{\raisebox{-0.12em}{\includegraphics[width=0.9em,height=0.9em,keepaspectratio]{figures/icons/qwen.pdf}}}\hspace{0.3em}Qwen-3.5-9B & 79.15 & [76.81, 81.21] & 74.57 & 0.077361 & 134.64 & 122/123 \\
\bottomrule
\end{tabularx}
\caption{Official-test baseline: 123 contracts and 2,091 judgments per model. Failed requests remain incorrect; cost and median client response time include all 123 attempts. Accuracy intervals use 5,000 whole-contract bootstrap samples, without multiplicity adjustment. Section~\ref{sec:protocol} defines cost and timing.}
\label{tab:testbaseline}
\end{table*}

Jev achieves 77.38\% baseline accuracy at \$0.000228 per contract. All seven hosted language-model comparators have higher point estimates. Gemini-3.5-Flash gains 5.16 points (paired 95\% interval: 3.49--6.98) at approximately six times the cost. Jev, Luna, Gemini-3.5-Flash, and Gemini Pro form the observed cost--accuracy frontier; Sonnet, Terra, and Haiku lie outside it. Gemini Pro costs 6.26 times as much as Gemini-3.5-Flash for an uncertain 0.67-point gain (interval: $-0.72$ to 2.10). Qwen-3.5-4B and Qwen-3.5-9B achieve 74.22\% and 79.15\%, including six and one budget-exhausted baseline responses.

\paragraph{Response time and output workload.}
Jev has the lowest median baseline response time, 1.24 seconds, compared with Gemini-3.5-Flash's 1.60 and Luna's 1.74. The ranking changes when only one label is requested. At the same visible catalog of 17 hypotheses, moving from B to C changes Jev's median from 1.15 to 1.22 seconds, Gemini-3.5-Flash's from 0.99 to 1.86, and Luna's from 1.10 to 1.78. Gemini-3.5-Flash is faster for one answer, whereas Jev is faster for joint prediction. These are end-to-end service times; Appendix~\ref{app:timing} reports their paired uncertainty and timing scope.

\subsection{Incomplete answers and the distribution of errors}
Truncation affects the Qwen results, but errors in completed answers account for much of their performance gap. Assigning every missing baseline label its gold answer would raise Qwen-3.5-4B to at most 79.10\% and Qwen-3.5-9B to 79.96\%, below Gemini-3.5-Flash's observed 82.54\%. On the repeated targets, the same optimistic completion raises Qwen-3.5-4B's consistently correct count from 17 to at most 19 and leaves Qwen-3.5-9B at 18: both of its affected targets already contain other wrong answers. Appendix~\ref{app:budget} reports the termination evidence and single-attempt scoring policy.

Separate development controls reproduce correction--regression cancellation and compare Qwen inference settings. More expensive settings improve mean accuracy for both Qwen models, while consistently correct counts stay at 17 for Qwen-3.5-4B and rise from 16 to 18 for Qwen-3.5-9B (Appendix~\ref{app:fairness}).

\section{Related Work}
\label{sec:related}
ContractNLI, CUAD, and LegalBench evaluate contract inference, extraction, and legal reasoning~\citep{koreeda2021contractnli,hendrycks2021cuad,guha2023legalbench}. Prompting research studies demonstration order, formatting, batching, and evidence or answer positions~\citep{lu2022fantastically,sclar2024quantifying,cheng2023batchprompting,lin2024batchprompt,liu2024lost,zheng2024selectors}. Our comparisons hold the contract and target fixed while varying visible hypotheses and requested outputs. CheckList and order-consistent scoring motivate examining individual decisions alongside aggregate quality~\citep{ribeiro2020checklist,frohmann2026equal}; we additionally separate corrections, regressions, answers that repeat the same wrong label, and repeat variation. HELM and model-routing research connect quality to efficiency and expenditure~\citep{liang2023helm,chen2024frugalgpt,ong2025routellm}. Self-consistency aggregates sampled reasoning paths~\citep{wang2023selfconsistency}; our secondary analysis aggregates already recorded final labels under fixed settings and request configurations. It evaluates the observed gains and persistent errors without treating answer agreement as correctness. We bring these dimensions together for shared-document judgments. Appendix~\ref{app:related} provides detailed comparisons.

\section{Discussion and Conclusion}
\label{sec:conclusion}
Contract inference needs decision-level evaluation alongside aggregate scores. Corrections and regressions can cancel, while higher mean accuracy can coexist with fewer judgments that remain correct throughout. The paired comparisons locate this distinction in the same contracts and hypotheses, and the equal-call control shows that it persists when both request strategies receive twelve calls per target. A model comparison should therefore report which judgments are corrected, which regress, and which remain wrong, alongside the overall score.

Agreement provides a second view of model behavior, but its value depends on the correctness of the agreed answers. In the additional sample, Jev produces more unanimous answers than Sonnet, yet a smaller fraction of those answers is correct. Errors repeated on every call also survive every vote formed from those calls. This links the persistent-error analysis to the small observed voting gains: repetition can resolve some variable answers, but it cannot recover a correct label that never appears in the recorded responses. For evaluating repeated-call strategies, corrections and regressions explain the return from additional calls more clearly than agreement alone.

Resource measurements complete the comparison. Jev is cheapest and fastest for the initial joint baseline, while all seven hosted language-model comparators have higher baseline accuracy point estimates. Response-time rankings change with the number of requested answers, and class-specific rankings change with the type of inference. Model choice therefore depends on the required workload and error profile as well as expenditure. Together, the evaluation connects three questions: which judgments change, which errors persist, and whether additional expenditure improves the final decisions.

\label{sec:content-end}
\clearpage

\section*{Limitations}
\label{sec:limitations}
The study covers ContractNLI classification on 123 test contracts and 17 hypothesis types: 30 repeated targets for ten models, extended by 93 targets for four models. Existing annotations supply the reference labels. Evidence-span extraction, human performance, other legal tasks, and possible training-data exposure remain outside this evaluation.

Baseline accuracy and consistent correctness score different target sets; the same-target analyses distinguish these effects. The equal-call control has 30 independent targets, making small differences uncertain. Voting is a retrospective finite-panel analysis, not a new prospective evaluation of an adaptive calling policy. Its subset averages do not add independent targets. The 93-target class, hypothesis, and length analyses are descriptive; small subgroups and different class mixtures limit their interpretation. Bootstrap intervals preserve contracts, are unadjusted for multiplicity, and can degenerate when observed outcomes agree. The initial model--contrast comparisons share targets and conditions.

Results describe deployed configurations with different interfaces, reasoning settings, providers, and collection times. Jev's C/D contrast changes native question-key order; Terra's requested order differs from its returned order. The follow-up compares B with an A/B/C/D mixture, without isolating each prompt component. The combined 123-target summary joins two collection dates. Qwen's inference comparison changes thinking, sampling, and generation allowance together. The three added models and four follow-up configurations were selected after initial results were inspected.

API costs use recorded charges or declared token rates; GPU costs value measured time at a reference rental rate. Timing includes network and service effects. Server computation, queueing, time to first token, and maximum throughput were not measured. Prices, caching, hardware, and serving policies can change resource rankings. Failed-request charges without usage remain unknown; the partial Astra development panel is reported separately.

\section*{Ethical considerations}
We use ContractNLI under its stated CC BY 4.0 license and retain attribution to its creators. The study uses existing data and annotations, with no new personal-data collection or human participation. Examples retain their dataset identifiers and original labels. Any redistribution should preserve attribution and licensing while minimizing additional identifying information. The evaluation concerns model behavior in a research setting. Incorrect or unstable judgments could cause harm if treated as professional legal or financial determinations.

\bibliography{references}

@inproceedings{koreeda2021contractnli,
  title = {{ContractNLI}: A Dataset for Document-level Natural Language Inference for Contracts},
  author = {Koreeda, Yuta and Manning, Christopher D.},
  booktitle = {Findings of the Association for Computational Linguistics: EMNLP 2021},
  year = {2021},
  pages = {1907--1919},
  publisher = {Association for Computational Linguistics},
  doi = {10.18653/v1/2021.findings-emnlp.164},
  url = {https://aclanthology.org/2021.findings-emnlp.164/},
  series = {EMNLP~'21},
  address = {Punta Cana, Dominican Republic}
}

@inproceedings{lin2024batchprompt,
  title = {{BatchPrompt}: Accomplish More with Less},
  author = {Lin, Jianzhe and Diesendruck, Maurice and Du, Liang and Abraham, Robin},
  booktitle = {Proceedings of the Twelfth International Conference on Learning Representations},
  year = {2024},
  url = {https://proceedings.iclr.cc/paper_files/paper/2024/hash/5d8c01de2dc698c54201c1c7d0b86974-Abstract-Conference.html},
  series = {ICLR~'24},
  address = {Vienna, Austria}
}

@article{frohmann2026equal,
  title = {Equal Ranking Quality, Different Decisions: Training Order-Consistent {LLM} Scorers},
  author = {Frohmann, Markus and Alavi, Mahdiyar and Lingg, Elizabeth and Rekabsaz, Navid},
  journal = {arXiv preprint arXiv:2608.26762},
  year = {2026},
  note = {Version 1},
  url = {https://arxiv.org/abs/2608.26762}
}

@inproceedings{lu2022fantastically,
  title = {Fantastically Ordered Prompts and Where to Find Them: Overcoming Few-Shot Prompt Order Sensitivity},
  author = {Lu, Yao and Bartolo, Max and Moore, Alastair and Riedel, Sebastian and Stenetorp, Pontus},
  booktitle = {Proceedings of the 60th Annual Meeting of the Association for Computational Linguistics (Volume 1: Long Papers)},
  year = {2022},
  pages = {8086--8098},
  publisher = {Association for Computational Linguistics},
  doi = {10.18653/v1/2022.acl-long.556},
  url = {https://aclanthology.org/2022.acl-long.556/},
  series = {ACL~'22},
  address = {Dublin, Ireland}
}

@article{liu2024lost,
  title = {Lost in the Middle: How Language Models Use Long Contexts},
  author = {Liu, Nelson F. and Lin, Kevin and Hewitt, John and Paranjape, Ashwin and Bevilacqua, Michele and Petroni, Fabio and Liang, Percy},
  journal = {Transactions of the Association for Computational Linguistics},
  volume = {12},
  year = {2024},
  pages = {157--173},
  doi = {10.1162/tacl_a_00638},
  url = {https://aclanthology.org/2024.tacl-1.9/}
}

@inproceedings{hendrycks2021cuad,
  title = {{CUAD}: An Expert-Annotated {NLP} Dataset for Legal Contract Review},
  author = {Hendrycks, Dan and Burns, Collin and Chen, Anya and Ball, Spencer},
  booktitle = {Proceedings of the Neural Information Processing Systems Track on Datasets and Benchmarks 1},
  year = {2021},
  url = {https://datasets-benchmarks-proceedings.neurips.cc/paper_files/paper/2021/hash/6ea9ab1baa0efb9e19094440c317e21b-Abstract-round1.html},
  series = {NeurIPS~'21},
  address = {Online}
}

@inproceedings{guha2023legalbench,
  title = {{LegalBench}: A Collaboratively Built Benchmark for Measuring Legal Reasoning in Large Language Models},
  author = {Guha, Neel and Nyarko, Julian and Ho, Daniel E. and R{\'e}, Christopher and Chilton, Adam and Narayana, Aditya and Chohlas-Wood, Alex and Peters, Austin and Waldon, Brandon and Rockmore, Daniel N. and Zambrano, Diego and Talisman, Dmitry and Hoque, Enam and Surani, Faiz and Fagan, Frank and Sarfaty, Galit and Dickinson, Gregory M. and Porat, Haggai and Hegland, Jason and Wu, Jessica and Nudell, Joe and Niklaus, Joel and Nay, John and Choi, Jonathan H. and Tobia, Kevin and Hagan, Margaret and Ma, Megan and Livermore, Michael and Rasumov-Rahe, Nikon and Holzenberger, Nils and Kolt, Noam and Henderson, Peter and Rehaag, Sean and Goel, Sharad and Gao, Shang and Williams, Spencer and Gandhi, Sunny and Zur, Tom and Iyer, Varun and Li, Zehua},
  booktitle = {Advances in Neural Information Processing Systems 36},
  pages = {44123--44279},
  publisher = {Curran Associates, Inc.},
  year = {2023},
  doi = {10.52202/075280-1915},
  url = {https://proceedings.neurips.cc/paper_files/paper/2023/hash/89e44582fd28ddfea1ea4dcb0ebbf4b0-Abstract-Datasets_and_Benchmarks.html},
  series = {NeurIPS~'23},
  address = {New Orleans, LA, USA}
}

@inproceedings{chalkidis2022lexglue,
  title = {{LexGLUE}: A Benchmark Dataset for Legal Language Understanding in {English}},
  author = {Chalkidis, Ilias and Jana, Abhik and Hartung, Dirk and Bommarito, Michael and Androutsopoulos, Ion and Katz, Daniel Martin and Aletras, Nikolaos},
  booktitle = {Proceedings of the 60th Annual Meeting of the Association for Computational Linguistics (Volume 1: Long Papers)},
  pages = {4310--4330},
  publisher = {Association for Computational Linguistics},
  year = {2022},
  doi = {10.18653/v1/2022.acl-long.297},
  url = {https://aclanthology.org/2022.acl-long.297/},
  series = {ACL~'22},
  address = {Dublin, Ireland}
}

@inproceedings{wang2023maud,
  title = {{MAUD}: An Expert-Annotated Legal {NLP} Dataset for Merger Agreement Understanding},
  author = {Wang, Steven H. and Scardigli, Antoine and Tang, Leonard and Chen, Wei and Levkin, Dimitry and Chen, Anya and Ball, Spencer and Woodside, Thomas and Zhang, Oliver and Hendrycks, Dan},
  booktitle = {Proceedings of the 2023 Conference on Empirical Methods in Natural Language Processing},
  pages = {16369--16382},
  publisher = {Association for Computational Linguistics},
  year = {2023},
  doi = {10.18653/v1/2023.emnlp-main.1019},
  url = {https://aclanthology.org/2023.emnlp-main.1019/},
  series = {EMNLP~'23},
  address = {Singapore}
}

@inproceedings{sclar2024quantifying,
  title = {Quantifying Language Models' Sensitivity to Spurious Features in Prompt Design or: How {I} learned to start worrying about prompt formatting},
  author = {Sclar, Melanie and Choi, Yejin and Tsvetkov, Yulia and Suhr, Alane},
  booktitle = {Proceedings of the Twelfth International Conference on Learning Representations},
  year = {2024},
  url = {https://proceedings.iclr.cc/paper_files/paper/2024/hash/6c0e99d736da621403018ca7b32b1a4d-Abstract-Conference.html},
  series = {ICLR~'24},
  address = {Vienna, Austria}
}

@inproceedings{zhao2021calibrate,
  title = {Calibrate Before Use: Improving Few-shot Performance of Language Models},
  author = {Zhao, Tony Z. and Wallace, Eric and Feng, Shi and Klein, Dan and Singh, Sameer},
  booktitle = {Proceedings of the 38th International Conference on Machine Learning},
  pages = {12697--12706},
  year = {2021},
  volume = {139},
  series = {Proceedings of Machine Learning Research},
  publisher = {PMLR},
  url = {https://proceedings.mlr.press/v139/zhao21c.html},
  address = {Online}
}

@inproceedings{ribeiro2020checklist,
  title = {Beyond Accuracy: Behavioral Testing of {NLP} Models with {CheckList}},
  author = {Ribeiro, Marco Tulio and Wu, Tongshuang and Guestrin, Carlos and Singh, Sameer},
  booktitle = {Proceedings of the 58th Annual Meeting of the Association for Computational Linguistics},
  year = {2020},
  pages = {4902--4912},
  publisher = {Association for Computational Linguistics},
  doi = {10.18653/v1/2020.acl-main.442},
  url = {https://aclanthology.org/2020.acl-main.442/},
  series = {ACL~'20},
  address = {Online}
}

@inproceedings{webson2022prompts,
  title = {Do Prompt-Based Models Really Understand the Meaning of Their Prompts?},
  author = {Webson, Albert and Pavlick, Ellie},
  booktitle = {Proceedings of the 2022 Conference of the North American Chapter of the Association for Computational Linguistics: Human Language Technologies},
  year = {2022},
  pages = {2300--2344},
  publisher = {Association for Computational Linguistics},
  doi = {10.18653/v1/2022.naacl-main.167},
  url = {https://aclanthology.org/2022.naacl-main.167/},
  series = {NAACL-HLT~'22},
  address = {Seattle, WA, USA}
}

@inproceedings{zheng2024selectors,
  title = {Large Language Models Are Not Robust Multiple Choice Selectors},
  author = {Zheng, Chujie and Zhou, Hao and Meng, Fandong and Zhou, Jie and Huang, Minlie},
  booktitle = {Proceedings of the Twelfth International Conference on Learning Representations},
  year = {2024},
  url = {https://proceedings.iclr.cc/paper_files/paper/2024/hash/54dd9e0cff6d9214e20d97eb2a3bae49-Abstract-Conference.html},
  series = {ICLR~'24},
  address = {Vienna, Austria}
}

@article{liang2023helm,
  title = {Holistic Evaluation of Language Models},
  author = {Liang, Percy and Bommasani, Rishi and Lee, Tony and Tsipras, Dimitris and Soylu, Dilara and Yasunaga, Michihiro and Zhang, Yian and Narayanan, Deepak and Wu, Yuhuai and Kumar, Ananya and Newman, Benjamin and Yuan, Binhang and Yan, Bobby and Zhang, Ce and Cosgrove, Christian and Manning, Christopher D. and R{\'e}, Christopher and Acosta-Navas, Diana and Hudson, Drew A. and Zelikman, Eric and Durmus, Esin and Ladhak, Faisal and Rong, Frieda and Ren, Hongyu and Yao, Huaxiu and Wang, Jue and Santhanam, Keshav and Orr, Laurel and Zheng, Lucia and Yuksekgonul, Mert and Suzgun, Mirac and Kim, Nathan and Guha, Neel and Chatterji, Niladri and Khattab, Omar and Henderson, Peter and Huang, Qian and Chi, Ryan and Xie, Sang Michael and Santurkar, Shibani and Ganguli, Surya and Hashimoto, Tatsunori and Icard, Thomas and Zhang, Tianyi and Chaudhary, Vishrav and Wang, William and Li, Xuechen and Mai, Yifan and Zhang, Yuhui and Koreeda, Yuta},
  journal = {Transactions on Machine Learning Research},
  year = {2023},
  url = {https://openreview.net/forum?id=iO4LZibEqW}
}

@article{chen2024frugalgpt,
  title = {{FrugalGPT}: How to Use Large Language Models While Reducing Cost and Improving Performance},
  author = {Chen, Lingjiao and Zaharia, Matei and Zou, James},
  journal = {Transactions on Machine Learning Research},
  year = {2024},
  url = {https://openreview.net/forum?id=cSimKw5p6R}
}

@inproceedings{ong2025routellm,
  title = {{RouteLLM}: Learning to Route {LLMs} from Preference Data},
  author = {Ong, Isaac and Almahairi, Amjad and Wu, Vincent and Chiang, Wei-Lin and Wu, Tianhao and Gonzalez, Joseph E. and Kadous, M Waleed and Stoica, Ion},
  booktitle = {Proceedings of the Thirteenth International Conference on Learning Representations},
  year = {2025},
  url = {https://proceedings.iclr.cc/paper_files/paper/2025/hash/5503a7c69d48a2f86fc00b3dc09de686-Abstract-Conference.html},
  series = {ICLR~'25},
  address = {Singapore}
}

@article{kapoor2025agents,
  title = {{AI} Agents That Matter},
  author = {Kapoor, Sayash and Stroebl, Benedikt and Siegel, Zachary S. and Nadgir, Nitya and Narayanan, Arvind},
  journal = {Transactions on Machine Learning Research},
  year = {2025},
  url = {https://openreview.net/forum?id=Zy4uFzMviZ}
}

@inproceedings{cheng2023batchprompting,
  title = {Batch Prompting: Efficient Inference with Large Language Model {APIs}},
  author = {Cheng, Zhoujun and Kasai, Jungo and Yu, Tao},
  booktitle = {Proceedings of the 2023 Conference on Empirical Methods in Natural Language Processing: Industry Track},
  year = {2023},
  publisher = {Association for Computational Linguistics},
  pages = {792--810},
  doi = {10.18653/v1/2023.emnlp-industry.74},
  url = {https://aclanthology.org/2023.emnlp-industry.74/},
  series = {EMNLP~'23},
  address = {Singapore}
}

@inproceedings{dror2018hitchhiker,
  title = {The Hitchhiker's Guide to Testing Statistical Significance in Natural Language Processing},
  author = {Dror, Rotem and Baumer, Gili and Shlomov, Segev and Reichart, Roi},
  booktitle = {Proceedings of the 56th Annual Meeting of the Association for Computational Linguistics (Volume 1: Long Papers)},
  year = {2018},
  pages = {1383--1392},
  publisher = {Association for Computational Linguistics},
  doi = {10.18653/v1/P18-1128},
  url = {https://aclanthology.org/P18-1128/},
  series = {ACL~'18},
  address = {Melbourne, Australia}
}

@inproceedings{dodge2019show,
  title = {Show Your Work: Improved Reporting of Experimental Results},
  author = {Dodge, Jesse and Gururangan, Suchin and Card, Dallas and Schwartz, Roy and Smith, Noah A.},
  booktitle = {Proceedings of the 2019 Conference on Empirical Methods in Natural Language Processing and the 9th International Joint Conference on Natural Language Processing},
  year = {2019},
  pages = {2185--2194},
  publisher = {Association for Computational Linguistics},
  doi = {10.18653/v1/D19-1224},
  url = {https://aclanthology.org/D19-1224/},
  series = {EMNLP-IJCNLP~'19},
  address = {Hong Kong, China}
}

@inproceedings{reimers2017score,
  title = {Reporting Score Distributions Makes a Difference: Performance Study of {LSTM}-networks for Sequence Tagging},
  author = {Reimers, Nils and Gurevych, Iryna},
  booktitle = {Proceedings of the 2017 Conference on Empirical Methods in Natural Language Processing},
  year = {2017},
  pages = {338--348},
  publisher = {Association for Computational Linguistics},
  doi = {10.18653/v1/D17-1035},
  url = {https://aclanthology.org/D17-1035/},
  series = {EMNLP~'17},
  address = {Copenhagen, Denmark}
}

@inproceedings{bashari2026gespi,
  title = {General Synthetic-Powered Inference},
  author = {Bashari, Meshi and Lee, Yonghoon and Lotan, Roy Maor and Dobriban, Edgar and Romano, Yaniv},
  booktitle = {Proceedings of the 43rd International Conference on Machine Learning},
  series = {ICML~'26},
  year = {2026},
  url = {https://icml.cc/virtual/2026/poster/61182},
  address = {Seoul, South Korea}
}

@inproceedings{balunovic2025matharena,
  title = {{MathArena}: Evaluating {LLMs} on Uncontaminated Math Competitions},
  author = {Balunovic, Mislav and Dekoninck, Jasper and Petrov, Ivo and Jovanovic, Nikola and Vechev, Martin},
  booktitle = {Advances in Neural Information Processing Systems 38},
  year = {2025},
  url = {https://papers.nips.cc/paper_files/paper/2025/hash/1d27c01ebd3e3aebe226b44fc970d803-Abstract-Datasets_and_Benchmarks_Track.html},
  series = {NeurIPS~'25},
  address = {San Diego, CA, USA},
  publisher = {Curran Associates, Inc.},
  doi = {10.52202/085713-0679}
}

@inproceedings{wang2023selfconsistency,
  title={Self-Consistency Improves Chain of Thought Reasoning in Language Models},
  author={Wang, Xuezhi and Wei, Jason and Schuurmans, Dale and Le, Quoc V. and Chi, Ed H. and Narang, Sharan and Chowdhery, Aakanksha and Zhou, Denny},
  booktitle={The Eleventh International Conference on Learning Representations},
  year={2023},
  url={https://openreview.net/forum?id=1PL1NIMMrw}
}

\clearpage
\appendix
\suppressfloats[t]
\makeatletter
\setlength{\@dblfptop}{0pt}
\setlength{\@dblfpsep}{18pt}
\setlength{\@dblfpbot}{0pt plus 1fil}
\makeatother

\section{Supplementary Official-Test Comparisons}
\label{app:testdetails}
The supplementary official-test analyses examine class-specific errors, paired model differences, and response transitions for all ten models. Both baseline and repeated-target evaluations are complete. Development grouping and repeated-target experiments are reported separately from the official test.

Table~\ref{tab:testanchor} gives condition-level correctness, all-twelve uncertainty, and response validity for the complete repeated-target panel. Each model has 360 attempted requests, with 90 intended target responses per condition.
\begin{table*}[t]

\centering
\small
\begin{tabularx}{\textwidth}{Xccccccc}
\toprule
\rowcolor{tableHeader}
\textbf{Model} & \shortstack{A (\%)\\$\uparrow$} & \shortstack{B (\%)\\$\uparrow$} & \shortstack{C (\%)\\$\uparrow$} & \shortstack{D (\%)\\$\uparrow$} & \shortstack{All 12\\correct $\uparrow$} & \shortstack{95\%\\interval} & Valid calls \\
\midrule
\rowcolor{jevRow}
\makebox[0.9em][c]{\raisebox{-0.12em}{\includegraphics[width=0.9em,height=0.9em,keepaspectratio]{figures/icons/jev.pdf}}}\hspace{0.3em}Jev-1.13.0 & 76.67 & 81.11 & 80.00 & 80.00 & \secondscore{23/30} & [60.0, 90.0] & 360/360 \\
\addlinespace[3pt]
\makebox[0.9em][c]{\raisebox{-0.12em}{\includegraphics[width=0.9em,height=0.9em,keepaspectratio]{figures/icons/gemini.pdf}}}\hspace{0.3em}Gemini-3.5-Flash & 82.22 & 83.33 & 84.44 & 80.00 & 21/30 & [53.3, 83.3] & 360/360 \\
\makebox[0.9em][c]{\raisebox{-0.12em}{\includegraphics[width=0.9em,height=0.9em,keepaspectratio]{figures/icons/gemini.pdf}}}\hspace{0.3em}Gemini-3.1-Pro-Preview & \bestscore{85.56} & \bestscore{85.56} & 86.67 & \secondscore{83.33} & 22/30 & [56.7, 86.7] & 359/360 \\
\addlinespace[3pt]
\makebox[0.9em][c]{\raisebox{-0.12em}{\includegraphics[width=0.9em,height=0.9em,keepaspectratio]{figures/icons/openai.pdf}}}\hspace{0.3em}GPT-5.6-Luna & 80.00 & 82.22 & 84.44 & 81.11 & 22/30 & [56.7, 86.7] & 360/360 \\
\makebox[0.9em][c]{\raisebox{-0.12em}{\includegraphics[width=0.9em,height=0.9em,keepaspectratio]{figures/icons/openai.pdf}}}\hspace{0.3em}GPT-5.6-Terra & 76.67 & 83.33 & 83.33 & 78.89 & 20/30 & [50.0, 83.3] & 360/360 \\
\makebox[0.9em][c]{\raisebox{-0.12em}{\includegraphics[width=0.9em,height=0.9em,keepaspectratio]{figures/icons/openai.pdf}}}\hspace{0.3em}GPT-6-Astra & 76.67 & 74.44 & 75.56 & 78.89 & 22/30 & [56.7, 86.7] & 360/360 \\
\addlinespace[3pt]
\makebox[0.9em][c]{\raisebox{-0.12em}{\includegraphics[width=0.9em,height=0.9em,keepaspectratio]{figures/icons/claude.pdf}}}\hspace{0.3em}Claude-Haiku-4.5 & 78.89 & \secondscore{84.44} & \secondscore{87.78} & 76.67 & 19/30 & [46.7, 80.0] & 360/360 \\
\makebox[0.9em][c]{\raisebox{-0.12em}{\includegraphics[width=0.9em,height=0.9em,keepaspectratio]{figures/icons/claude.pdf}}}\hspace{0.3em}Claude-Sonnet-5 & \secondscore{83.33} & \secondscore{84.44} & \bestscore{88.89} & \bestscore{90.00} & \bestscore{24/30} & [63.3, 93.3] & 360/360 \\
\addlinespace[3pt]
\makebox[0.9em][c]{\raisebox{-0.12em}{\includegraphics[width=0.9em,height=0.9em,keepaspectratio]{figures/icons/qwen.pdf}}}\hspace{0.3em}Qwen-3.5-4B & 75.56 & 75.56 & 81.11 & 77.78 & 17/30 & [40.0, 73.3] & 357/360 \\
\makebox[0.9em][c]{\raisebox{-0.12em}{\includegraphics[width=0.9em,height=0.9em,keepaspectratio]{figures/icons/qwen.pdf}}}\hspace{0.3em}Qwen-3.5-9B & 78.89 & 80.00 & 78.89 & 78.89 & 18/30 & [43.3, 76.7] & 358/360 \\
\bottomrule
\end{tabularx}
\caption{Official-test repeated-target evaluation on 30 contracts with one target each, four conditions and three repeats (90 target judgments per condition; 360 requests per model). All 360 requests were attempted for each of the ten models. All 12 correct retains all 30 targets and requires every response to be valid and correct. Its percentage intervals use 5,000 whole-contract bootstrap samples. Invalid responses remain in every intended denominator.}
\label{tab:testanchor}
\end{table*}

\begingroup
Table~\ref{tab:testclassrecalls} reports class recalls for all ten models on the same 123 test contracts. Table~\ref{tab:testpaired} reports paired differences from Jev for baseline accuracy and correctness across all twelve responses. Table~\ref{tab:teststates} separates consistent correctness, answers that repeat the same wrong label, valid but changed answers, and invalid responses. Table~\ref{tab:test-sensitivity-summary} compares changed answers with their net effect on correctness across all ten models. Table~\ref{tab:test-sensitivity-counts} adds identical-request repetition and valid-pair coverage; Table~\ref{tab:testchanges} separates the directions of those condition changes. These analyses retain the same contract clusters and intended denominators as the main evaluation.
\endgroup

\begingroup
The class recalls expose error profiles that overall accuracy obscures. Jev has the highest observed contradiction recall, Astra the highest entailment recall, and Sonnet the highest not-mentioned recall. The class supports are 968 entailments, 220 contradictions, and 903 not-mentioned judgments.
\endgroup

\begingroup
Jev correctly classifies 170 of 220 contradictions, whereas Gemini Pro correctly classifies 135. Relative to Jev, Gemini Pro gains 47 correct entailments and 110 correct not-mentioned labels but loses 35 correct contradictions, producing its net gain of 122 correct judgments. Overall accuracy weights class-specific recall by class support. The class-specific counts expose the error trade-offs underlying the higher overall score.
\endgroup

\begingroup
Qwen-3.5-9B illustrates how similar condition-level scores can conceal changes at the target level. Each condition yields 71 or 72 correct answers among 90 intended judgments, while eight targets receive different valid labels across the twelve responses. Two more targets encounter generation-budget exhaustion, and two repeat the same wrong label throughout. The remaining eighteen are correct in every response. Invalid answers count as incorrect and prevent consistent correctness, but supply no observed label for the change-rate calculation.
\endgroup

\begin{table*}[t]
\centering
\small
\begin{tabularx}{\textwidth}{Xccc}
\toprule
\rowcolor{tableHeader}
\textbf{Model} & \shortstack{Entailment recall\\(\%) $\uparrow$ $n=968$} & \shortstack{Contradiction recall\\(\%) $\uparrow$ $n=220$} & \shortstack{Not mentioned recall\\(\%) $\uparrow$ $n=903$} \\
\midrule
\rowcolor{jevRow}
\makebox[0.9em][c]{\raisebox{-0.12em}{\includegraphics[width=0.9em,height=0.9em,keepaspectratio]{figures/icons/jev.pdf}}}\hspace{0.3em}Jev-1.13.0 & 86.88 & \bestscore{77.27} & 67.22 \\
\addlinespace[3pt]
\makebox[0.9em][c]{\raisebox{-0.12em}{\includegraphics[width=0.9em,height=0.9em,keepaspectratio]{figures/icons/gemini.pdf}}}\hspace{0.3em}Gemini-3.5-Flash & 91.32 & 63.18 & 77.85 \\
\makebox[0.9em][c]{\raisebox{-0.12em}{\includegraphics[width=0.9em,height=0.9em,keepaspectratio]{figures/icons/gemini.pdf}}}\hspace{0.3em}Gemini-3.1-Pro-Preview & 91.74 & 61.36 & \secondscore{79.40} \\
\addlinespace[3pt]
\makebox[0.9em][c]{\raisebox{-0.12em}{\includegraphics[width=0.9em,height=0.9em,keepaspectratio]{figures/icons/openai.pdf}}}\hspace{0.3em}GPT-5.6-Luna & \secondscore{93.29} & 73.64 & 64.89 \\
\makebox[0.9em][c]{\raisebox{-0.12em}{\includegraphics[width=0.9em,height=0.9em,keepaspectratio]{figures/icons/openai.pdf}}}\hspace{0.3em}GPT-5.6-Terra & 89.98 & 71.36 & 77.41 \\
\makebox[0.9em][c]{\raisebox{-0.12em}{\includegraphics[width=0.9em,height=0.9em,keepaspectratio]{figures/icons/openai.pdf}}}\hspace{0.3em}GPT-6-Astra & \bestscore{93.60} & \secondscore{74.55} & 70.99 \\
\addlinespace[3pt]
\makebox[0.9em][c]{\raisebox{-0.12em}{\includegraphics[width=0.9em,height=0.9em,keepaspectratio]{figures/icons/claude.pdf}}}\hspace{0.3em}Claude-Haiku-4.5 & 91.63 & 62.73 & 70.43 \\
\makebox[0.9em][c]{\raisebox{-0.12em}{\includegraphics[width=0.9em,height=0.9em,keepaspectratio]{figures/icons/claude.pdf}}}\hspace{0.3em}Claude-Sonnet-5 & 90.50 & 55.00 & \bestscore{80.40} \\
\addlinespace[3pt]
\makebox[0.9em][c]{\raisebox{-0.12em}{\includegraphics[width=0.9em,height=0.9em,keepaspectratio]{figures/icons/qwen.pdf}}}\hspace{0.3em}Qwen-3.5-4B & 77.48 & 67.27 & 72.43 \\
\makebox[0.9em][c]{\raisebox{-0.12em}{\includegraphics[width=0.9em,height=0.9em,keepaspectratio]{figures/icons/qwen.pdf}}}\hspace{0.3em}Qwen-3.5-9B & 84.61 & 73.64 & 74.64 \\
\bottomrule
\end{tabularx}
\caption{Class recalls on the official-test baseline: 123 contracts and all 2,091 labeled judgments per model. Header counts give the gold-label support for each class and are identical across all ten models. Failed requests contribute no correct predictions and remain in each class denominator. These are test-split results, separate from the development class analysis.}
\label{tab:testclassrecalls}
\end{table*}

\begingroup

\begin{table*}[t]
\centering
\small
\begin{tabularx}{\textwidth}{Xcccc}
\toprule
\rowcolor{tableHeader}
\textbf{Model} & $\Delta$ accuracy & 95\% interval & $\Delta$ All 12 correct & 95\% interval \\
\midrule
\makebox[0.9em][c]{\raisebox{-0.12em}{\includegraphics[width=0.9em,height=0.9em,keepaspectratio]{figures/icons/gemini.pdf}}}\hspace{0.3em}Gemini-3.5-Flash & +5.16 & [+3.49, +6.98] & -6.67 & [-16.67, +0.00] \\
\makebox[0.9em][c]{\raisebox{-0.12em}{\includegraphics[width=0.9em,height=0.9em,keepaspectratio]{figures/icons/gemini.pdf}}}\hspace{0.3em}Gemini-3.1-Pro-Preview & +5.83 & [+4.11, +7.65] & -3.33 & [-16.67, +10.00] \\
\addlinespace[3pt]
\makebox[0.9em][c]{\raisebox{-0.12em}{\includegraphics[width=0.9em,height=0.9em,keepaspectratio]{figures/icons/openai.pdf}}}\hspace{0.3em}GPT-5.6-Luna & +1.58 & [-0.05, +3.11] & -3.33 & [-10.00, +0.00] \\
\makebox[0.9em][c]{\raisebox{-0.12em}{\includegraphics[width=0.9em,height=0.9em,keepaspectratio]{figures/icons/openai.pdf}}}\hspace{0.3em}GPT-5.6-Terra & +5.21 & [+3.73, +6.70] & -10.00 & [-20.00, +0.00] \\
\makebox[0.9em][c]{\raisebox{-0.12em}{\includegraphics[width=0.9em,height=0.9em,keepaspectratio]{figures/icons/openai.pdf}}}\hspace{0.3em}GPT-6-Astra & +4.45 & [+2.92, +5.93] & -3.33 & [-16.67, +10.00] \\
\addlinespace[3pt]
\makebox[0.9em][c]{\raisebox{-0.12em}{\includegraphics[width=0.9em,height=0.9em,keepaspectratio]{figures/icons/claude.pdf}}}\hspace{0.3em}Claude-Haiku-4.5 & +2.06 & [+0.43, +3.73] & -13.33 & [-26.67, -3.33] \\
\makebox[0.9em][c]{\raisebox{-0.12em}{\includegraphics[width=0.9em,height=0.9em,keepaspectratio]{figures/icons/claude.pdf}}}\hspace{0.3em}Claude-Sonnet-5 & +5.02 & [+3.25, +6.79] & +3.33 & [-10.00, +16.67] \\
\addlinespace[3pt]
\makebox[0.9em][c]{\raisebox{-0.12em}{\includegraphics[width=0.9em,height=0.9em,keepaspectratio]{figures/icons/qwen.pdf}}}\hspace{0.3em}Qwen-3.5-4B & -3.16 & [-6.70, +0.00] & -20.00 & [-36.67, -3.33] \\
\makebox[0.9em][c]{\raisebox{-0.12em}{\includegraphics[width=0.9em,height=0.9em,keepaspectratio]{figures/icons/qwen.pdf}}}\hspace{0.3em}Qwen-3.5-9B & +1.77 & [-0.43, +3.73] & -16.67 & [-33.33, +0.00] \\
\bottomrule
\end{tabularx}
\caption{Paired differences from Jev in percentage points (comparison model minus Jev). The baseline resamples 123 contracts and All 12 correct resamples 30 contracts, using 5,000 paired whole-contract bootstrap draws. All ten models have complete evaluations. Intervals are exploratory, descriptive, and unadjusted for multiple comparisons.}
\label{tab:testpaired}
\end{table*}
The net column concerns jointly valid pairs only. For Qwen-3.5-9B in C--D, seven corrections and five regressions give a net gain of two within 88 valid pairs, although the full-sample correct count remains 71/90 in both conditions. This equality includes the effect of invalid responses and is not another instance of exact correction--regression cancellation. Similarly, Jev's zero net change in C--D reflects zero changed labels, rather than offsetting changes.
\begin{table*}[t]
\centering
\begingroup

\small
\setlength{\tabcolsep}{2pt}

\begin{tabularx}{\textwidth}{@{}Xlcccccc@{}}
\toprule
\rowcolor{tableHeader}
\textbf{Model} & Contrast & Valid pairs & Changed & Corrections & Regressions & \shortstack{Wrong-label\\changes} & Net \\
\midrule
\multirow[c]{3}{*}{\makebox[0.9em][c]{\raisebox{-0.12em}{\includegraphics[width=0.9em,height=0.9em,keepaspectratio]{figures/icons/jev.pdf}}}\hspace{0.3em}Jev-1.13.0} & A$\to$B & 90/90 & 4 & 4 & 0 & 0 & $+4$ \\
 & B$\to$C & 90/90 & 1 & 0 & 1 & 0 & $-1$ \\
 & C$\to$D & 90/90 & 0 & 0 & 0 & 0 & $0$ \\
\modelsep
\multirow[c]{3}{*}{\makebox[0.9em][c]{\raisebox{-0.12em}{\includegraphics[width=0.9em,height=0.9em,keepaspectratio]{figures/icons/gemini.pdf}}}\hspace{0.3em}Gemini-3.5-Flash} & A$\to$B & 90/90 & 7 & 4 & 3 & 0 & $+1$ \\
 & B$\to$C & 90/90 & 3 & 2 & 1 & 0 & $+1$ \\
 & C$\to$D & 90/90 & 6 & 1 & 5 & 0 & $-4$ \\
\modelsep
\multirow[c]{3}{*}{\makebox[0.9em][c]{\raisebox{-0.12em}{\includegraphics[width=0.9em,height=0.9em,keepaspectratio]{figures/icons/gemini.pdf}}}\hspace{0.3em}Gemini-3.1-Pro-Preview} & A$\to$B & 89/90 & 7 & 3 & 4 & 0 & $-1$ \\
 & B$\to$C & 90/90 & 7 & 4 & 3 & 0 & $+1$ \\
 & C$\to$D & 90/90 & 3 & 0 & 3 & 0 & $-3$ \\
\modelsep
\multirow[c]{3}{*}{\makebox[0.9em][c]{\raisebox{-0.12em}{\includegraphics[width=0.9em,height=0.9em,keepaspectratio]{figures/icons/openai.pdf}}}\hspace{0.3em}GPT-5.6-Luna} & A$\to$B & 90/90 & 8 & 5 & 3 & 0 & $+2$ \\
 & B$\to$C & 90/90 & 8 & 5 & 3 & 0 & $+2$ \\
 & C$\to$D & 90/90 & 11 & 3 & 6 & 2 & $-3$ \\
\modelsep
\multirow[c]{3}{*}{\makebox[0.9em][c]{\raisebox{-0.12em}{\includegraphics[width=0.9em,height=0.9em,keepaspectratio]{figures/icons/openai.pdf}}}\hspace{0.3em}GPT-5.6-Terra} & A$\to$B & 90/90 & 7 & 6 & 0 & 1 & $+6$ \\
 & B$\to$C & 90/90 & 7 & 3 & 3 & 1 & $0$ \\
 & C$\to$D & 90/90 & 10 & 2 & 6 & 2 & $-4$ \\
\modelsep
\multirow[c]{3}{*}{\makebox[0.9em][c]{\raisebox{-0.12em}{\includegraphics[width=0.9em,height=0.9em,keepaspectratio]{figures/icons/openai.pdf}}}\hspace{0.3em}GPT-6-Astra} & A$\to$B & 90/90 & 4 & 1 & 3 & 0 & $-2$ \\
 & B$\to$C & 90/90 & 3 & 2 & 1 & 0 & $+1$ \\
 & C$\to$D & 90/90 & 3 & 3 & 0 & 0 & $+3$ \\
\modelsep
\multirow[c]{3}{*}{\makebox[0.9em][c]{\raisebox{-0.12em}{\includegraphics[width=0.9em,height=0.9em,keepaspectratio]{figures/icons/claude.pdf}}}\hspace{0.3em}Claude-Haiku-4.5} & A$\to$B & 90/90 & 8 & 6 & 1 & 1 & $+5$ \\
 & B$\to$C & 90/90 & 6 & 3 & 0 & 3 & $+3$ \\
 & C$\to$D & 90/90 & 14 & 2 & 12 & 0 & $-10$ \\
\modelsep
\multirow[c]{3}{*}{\makebox[0.9em][c]{\raisebox{-0.12em}{\includegraphics[width=0.9em,height=0.9em,keepaspectratio]{figures/icons/claude.pdf}}}\hspace{0.3em}Claude-Sonnet-5} & A$\to$B & 90/90 & 5 & 3 & 2 & 0 & $+1$ \\
 & B$\to$C & 90/90 & 6 & 5 & 1 & 0 & $+4$ \\
 & C$\to$D & 90/90 & 3 & 1 & 0 & 2 & $+1$ \\
\modelsep
\multirow[c]{3}{*}{\makebox[0.9em][c]{\raisebox{-0.12em}{\includegraphics[width=0.9em,height=0.9em,keepaspectratio]{figures/icons/qwen.pdf}}}\hspace{0.3em}Qwen-3.5-4B} & A$\to$B & 90/90 & 9 & 4 & 4 & 1 & $0$ \\
 & B$\to$C & 88/90 & 9 & 7 & 1 & 1 & $+6$ \\
 & C$\to$D & 87/90 & 14 & 5 & 8 & 1 & $-3$ \\
\modelsep
\multirow[c]{3}{*}{\makebox[0.9em][c]{\raisebox{-0.12em}{\includegraphics[width=0.9em,height=0.9em,keepaspectratio]{figures/icons/qwen.pdf}}}\hspace{0.3em}Qwen-3.5-9B} & A$\to$B & 90/90 & 15 & 8 & 7 & 0 & $+1$ \\
 & B$\to$C & 90/90 & 7 & 3 & 4 & 0 & $-1$ \\
 & C$\to$D & 88/90 & 12 & 7 & 5 & 0 & $+2$ \\
\bottomrule
\end{tabularx}
\caption{Directions and net changes underlying Table~\ref{tab:test-sensitivity-summary}. Each contrast matches the same target and repeat, giving 90 planned pairs per model. Corrections change a wrong label to the gold label; regressions do the reverse. Wrong-label changes switch between two incorrect labels. Changed is the sum of the three transition counts; Net is corrections minus regressions within jointly valid pairs. With incomplete coverage, Net need not equal the full-sample change in correct answers, where invalid responses remain incorrect. No ranking is assigned to these descriptive counts.}
\label{tab:testchanges}
\endgroup
\end{table*}
Table~\ref{tab:test-sensitivity-counts} preserves all exact changed-label counts and valid-pair denominators for the cross-condition and unchanged-repeat comparisons.
\begin{table*}[t]
\centering
\begingroup

\small
\setlength{\tabcolsep}{4pt}
\begin{tabular}{@{}lccccccc@{}}
\toprule
& \multicolumn{3}{c}{Across conditions} & \multicolumn{4}{c}{Identical-request repetition} \\
\cmidrule(lr){2-4}\cmidrule(l){5-8}
\textbf{Model} & A--B & B--C & C--D & A & B & C & D \\
\midrule
\makebox[0.9em][c]{\raisebox{-0.12em}{\includegraphics[width=0.9em,height=0.9em,keepaspectratio]{figures/icons/jev.pdf}}}\hspace{0.3em}Jev-1.13.0 & 4/90 & 1/90 & 0/90 & 0/90 & 2/90 & 0/90 & 0/90 \\
\addlinespace[2pt]
\makebox[0.9em][c]{\raisebox{-0.12em}{\includegraphics[width=0.9em,height=0.9em,keepaspectratio]{figures/icons/gemini.pdf}}}\hspace{0.3em}Gemini-3.5-Flash & 7/90 & 3/90 & 6/90 & 2/90 & 0/90 & 4/90 & 0/90 \\
\makebox[0.9em][c]{\raisebox{-0.12em}{\includegraphics[width=0.9em,height=0.9em,keepaspectratio]{figures/icons/gemini.pdf}}}\hspace{0.3em}Gemini-3.1-Pro-Preview & 7/89 & 7/90 & 3/90 & 4/88 & 2/90 & 4/90 & 0/90 \\
\addlinespace[2pt]
\makebox[0.9em][c]{\raisebox{-0.12em}{\includegraphics[width=0.9em,height=0.9em,keepaspectratio]{figures/icons/openai.pdf}}}\hspace{0.3em}GPT-5.6-Luna & 8/90 & 8/90 & 11/90 & 0/90 & 2/90 & 2/90 & 2/90 \\
\makebox[0.9em][c]{\raisebox{-0.12em}{\includegraphics[width=0.9em,height=0.9em,keepaspectratio]{figures/icons/openai.pdf}}}\hspace{0.3em}GPT-5.6-Terra & 7/90 & 7/90 & 10/90 & 6/90 & 6/90 & 5/90 & 8/90 \\
\makebox[0.9em][c]{\raisebox{-0.12em}{\includegraphics[width=0.9em,height=0.9em,keepaspectratio]{figures/icons/openai.pdf}}}\hspace{0.3em}GPT-6-Astra & 4/90 & 3/90 & 3/90 & 4/90 & 2/90 & 2/90 & 2/90 \\
\addlinespace[2pt]
\makebox[0.9em][c]{\raisebox{-0.12em}{\includegraphics[width=0.9em,height=0.9em,keepaspectratio]{figures/icons/claude.pdf}}}\hspace{0.3em}Claude-Haiku-4.5 & 8/90 & 6/90 & 14/90 & 2/90 & 6/90 & 2/90 & 6/90 \\
\makebox[0.9em][c]{\raisebox{-0.12em}{\includegraphics[width=0.9em,height=0.9em,keepaspectratio]{figures/icons/claude.pdf}}}\hspace{0.3em}Claude-Sonnet-5 & 5/90 & 6/90 & 3/90 & 0/90 & 2/90 & 2/90 & 2/90 \\
\addlinespace[2pt]
\makebox[0.9em][c]{\raisebox{-0.12em}{\includegraphics[width=0.9em,height=0.9em,keepaspectratio]{figures/icons/qwen.pdf}}}\hspace{0.3em}Qwen-3.5-4B & 9/90 & 9/88 & 14/87 & 10/90 & 10/90 & 6/86 & 13/88 \\
\makebox[0.9em][c]{\raisebox{-0.12em}{\includegraphics[width=0.9em,height=0.9em,keepaspectratio]{figures/icons/qwen.pdf}}}\hspace{0.3em}Qwen-3.5-9B & 15/90 & 7/90 & 12/88 & 12/90 & 6/90 & 8/90 & 11/86 \\
\bottomrule
\end{tabular}
\caption{Changed-label counts across request conditions and under identical-request repetition. Each entry is changed labels / jointly valid pairs, out of 90 planned pairs. Across-condition pairs match target and repeat; repetition uses all three unordered repeat pairs per target. A exposes and requests only the target; B exposes all 17 hypotheses but requests only the target; C requests all labels with the target first; D requests it last. A change need not be an error.}
\label{tab:test-sensitivity-counts}
\endgroup
\end{table*}
\endgroup

\section{Development Diagnostics}
\label{sec:diagnostics}
The development experiments clarify how request factors and outcome categories affect the interpretation of model performance. They include earlier diagnostics that motivated the test design, but use different samples, prompts, and inference configurations. We therefore preserve their results as a separate analysis, detailed in Appendices~\ref{app:development-protocol}--\ref{sec:planned}.

\paragraph{Compensating changes.}
On 30 development contracts, Qwen-3.5-4B changes 87 of 510 labels between joint and single-hypothesis requests, but the number correct changes by only four. The changes comprise 38 regressions, 34 corrections, and 15 switches between wrong labels. Aggregate accuracy therefore understates the number of decisions affected by grouping.

\paragraph{Visibility at fixed output workload.}
In the disjoint development target panel, Qwen-3.5-4B has the same A and B accuracy, while 24 of 90 paired responses change: 12 regressions and 12 corrections. The requested answer set is fixed, and every target has the same answer across unchanged repeats within each condition. This development diagnostic motivates the distinction illustrated in Figure~\ref{fig:example}: the visible hypotheses can change while the contract, target, and requested answer set remain fixed.

\paragraph{Additional computation and consistent correctness.}
A separate paired development comparison changes the Qwen configurations from thinking disabled with a 2,048-token allowance to thinking enabled with a 32,768-token allowance, also changing temperature and sampling. Mean target accuracy rises from 76.67\% to 81.11\% for Qwen-3.5-4B and from 74.17\% to 82.50\% for Qwen-3.5-9B, but their consistently correct counts change from 17 to 17 and from 16 to 18, respectively. Recorded completion-token use increases by 68.30 and 63.51 times. Both paired intervals for the all-twelve change include zero. Additional generation can thus accompany higher mean target accuracy without a comparably clear gain in consistent correctness. This is a comparison of complete configurations, not an isolated reasoning ablation; Appendix~\ref{app:fairness} reports the paired results.

\begingroup

\section{Development Data and Protocol}
\label{app:development-protocol}
\subsection{Data and sampling}
The development evaluation uses ContractNLI's annotated contract--hypothesis decisions \citep{koreeda2021contractnli}. We sort development documents by integer ID and sample 30 without replacement using seed 20260920. Each example retains the full contract text, all 17 original hypotheses, and the original class annotations. Evidence spans are excluded from model inputs, and sampling is independent of model predictions.

The sampled 510 targets comprise 280 entailments, 48 contradictions, and 182 not-mentioned labels, motivating class-level reporting alongside overall accuracy. Each experimental condition partitions the same hypotheses into request groups. We match predictions by document and hypothesis before paired analysis. The five-contract panel uses the first five documents in the randomized sample.

We independently recompute scores and transitions from individual predictions, verifying that paired observations share the same targets and gold labels. The three grouping collections have complete prediction coverage; Appendix~\ref{app:repro} gives the document selection and sample sizes. Scores measure agreement with the original annotations, whose legal interpretations may remain ambiguous in individual cases.

\subsection{Models and inference settings}
The development baseline compares Jev-1.13.0, Qwen-3.5-4B and 9B, Gemini-3.1-Pro-Preview, Gemini-3.5-Flash, GPT-5.6-Luna and Terra, GPT-6-Astra, Claude-Sonnet-5, and Claude-Haiku-4.5. Jev receives the contract as shared state and each hypothesis as a native Choice question. Generative models receive structured classification instructions and return a JSON label map. Task instructions are semantically aligned across these model-specific interfaces.

In the development baseline, grouping, and target experiments, the Qwen models use BF16, disabled thinking, and temperature zero. Gemini Pro uses low thinking, Gemini-3.5-Flash minimal thinking, Luna no reasoning, and Astra high reasoning. Reasoning budgets differ across models.

Terra, Sonnet, and Haiku use the same inference settings used in their test evaluations: medium reasoning for Terra, adaptive thinking with medium effort for Sonnet, and disabled thinking for Haiku. Terra and Sonnet permit 16,384 output tokens; Haiku permits 8,192. Gemini Pro's extended controls retain its original low-thinking configuration. GPT-6-Astra uses high reasoning with a 16,384-token output allowance.

\subsection{Experimental conditions}
The joint baseline requests all 17 labels once per contract. The five-contract diagnostic evaluates seven models under all seven conditions; the 30-contract grouping experiment evaluates Jev, both Qwen models, Gemini-3.5-Flash, and Luna. Joint baseline summaries and grouping experiments use separate collections. Thus, their $K=17$ scores can differ on the same contracts. Paired analyses always use predictions from the same experimental collection. In the earliest joint requests, request duration supplies document completion time.

For maximum batch sizes $K\in\{1,4,8,17\}$, the 17 judgments produce 17, 5, 3, and 1 requests respectively, including the final partial batch. The grouping comparison preserves hypothesis order. A permutation control uses a document-specific shuffle determined by seed 20260920 and the document ID. One additional $K=17$ request repeats the original input. Grouping, permutation, and repeat conditions use concurrency one; an additional $K=1$ condition permits four simultaneous requests. The order of model--condition groups is randomized separately in each evaluation. Evaluations may overlap in time, and prefix caching is enabled for local models.

\paragraph{Extended development comparisons.}
After inspecting the earlier development and test results, we extend the development comparison using the same contracts, prompts, and labels. Sonnet, Terra, and Haiku each receive the $K=17$ baseline, one unchanged repeat, and one permutation on all 30 contracts. Each model's baseline supports both the cost--accuracy comparison and its paired controls. Gemini Pro and GPT-6-Astra receive the same three conditions in separate new collections. Their controls use these matched baselines, while Figure~\ref{fig:development-map} retains the earlier Gemini Pro and GPT-6-Astra baselines. Every comparison pairs responses within the same collection. Model selection remains exploratory; samples, inference settings, and scoring rules are fixed before these additional requests.

\paragraph{Uncertainty, response time, and cost.}
Dependence among NLP observations matters for statistical evaluation~\citep{dror2018hitchhiker}. We report descriptive paired percentile intervals from 5,000 bootstrap resamples of whole contracts, preserving within-contract dependence, without multiple-comparison correction. Zero observed changes can yield a degenerate interval despite uncertainty about the underlying change probability. Document completion time measures elapsed time until all requested judgments for a contract are available, including grouped or concurrent requests; tables report its median. For a single joint request, it equals response time. API costs use provider-reported charges or recorded token usage priced at public rates. These timings combine model inference, network transfer, scheduling, caching, and provider infrastructure.

For local models, we estimate rental-equivalent inference cost per contract as
\begin{equation*}
\widehat C_{\mathrm{GPU}} = \frac{r}{30\times3600}\sum_{d=1}^{30} t_d,
\end{equation*}
where $t_d$ is the measured elapsed time in seconds for contract $d$ in the same 30-contract, $K=17$, concurrency-one baseline, and $r$ is \$2.00 per GPU-hour. Each model uses one GPU. This rounded reference lies within three published RTX PRO 6000 rental rates of \$1.80--\$2.09 per hour; it is not a market average.\footnote{Rates accessed September 21, 2026: \href{https://nebius.com/prices}{Nebius}, \$1.80; \href{https://verda.com/pricing}{Verda}, \$1.86; and \href{https://www.runpod.io/pricing}{Runpod}, \$2.09 per GPU-hour.} The measured hardware is an NVIDIA RTX PRO 6000 Blackwell Max-Q Workstation Edition GPU per model, with BF16 weights. The rental equivalent applies the stated rate to that measured time; the cited cloud offerings include different hardware variants. Model loading, preliminary checks, idle allocation, storage, transfer, and taxes are excluded.

\begin{table*}[t]

\centering
\small
\begin{tabularx}{\textwidth}{Xccccc}
\toprule
\rowcolor{tableHeader}
\textbf{Model} & \shortstack{Accuracy\\(\%) $\uparrow$} & \shortstack{Macro-F1\\(\%) $\uparrow$} & Seconds $\downarrow$ & \shortstack{USD/contract\\$\downarrow$} & Valid calls \\
\midrule
\rowcolor{jevRow}
\makebox[0.9em][c]{\raisebox{-0.12em}{\includegraphics[width=0.9em,height=0.9em,keepaspectratio]{figures/icons/jev.pdf}}}\hspace{0.3em}Jev-1.13.0 & 77.06 & 70.72 & \bestscore{1.25} & \bestscore{0.000250} & 30/30 \\
\addlinespace[3pt]
\makebox[0.9em][c]{\raisebox{-0.12em}{\includegraphics[width=0.9em,height=0.9em,keepaspectratio]{figures/icons/gemini.pdf}}}\hspace{0.3em}Gemini-3.5-Flash & 80.98 & 74.02 & 1.79 & 0.001486 & 30/30 \\
\makebox[0.9em][c]{\raisebox{-0.12em}{\includegraphics[width=0.9em,height=0.9em,keepaspectratio]{figures/icons/gemini.pdf}}}\hspace{0.3em}Gemini-3.1-Pro-Preview & \bestscore{82.94} & \bestscore{77.19} & 3.79 & 0.009502 & 30/30 \\
\addlinespace[3pt]
\makebox[0.9em][c]{\raisebox{-0.12em}{\includegraphics[width=0.9em,height=0.9em,keepaspectratio]{figures/icons/openai.pdf}}}\hspace{0.3em}GPT-5.6-Luna & 77.65 & 70.23 & 1.64 & 0.001058 & 30/30 \\
\makebox[0.9em][c]{\raisebox{-0.12em}{\includegraphics[width=0.9em,height=0.9em,keepaspectratio]{figures/icons/openai.pdf}}}\hspace{0.3em}GPT-5.6-Terra & \secondscore{81.76} & \secondscore{75.90} & 11.47 & 0.011318 & 30/30 \\
\makebox[0.9em][c]{\raisebox{-0.12em}{\includegraphics[width=0.9em,height=0.9em,keepaspectratio]{figures/icons/openai.pdf}}}\hspace{0.3em}GPT-6-Astra & \secondscore{81.76} & 74.64 & 20.35 & 0.089897 & 30/30 \\
\addlinespace[3pt]
\makebox[0.9em][c]{\raisebox{-0.12em}{\includegraphics[width=0.9em,height=0.9em,keepaspectratio]{figures/icons/claude.pdf}}}\hspace{0.3em}Claude-Haiku-4.5 & 78.82 & 73.46 & 5.73 & 0.005906 & 30/30 \\
\makebox[0.9em][c]{\raisebox{-0.12em}{\includegraphics[width=0.9em,height=0.9em,keepaspectratio]{figures/icons/claude.pdf}}}\hspace{0.3em}Claude-Sonnet-5 & 81.37 & 75.09 & 2.75 & 0.012868 & 30/30 \\
\addlinespace[3pt]
\makebox[0.9em][c]{\raisebox{-0.12em}{\includegraphics[width=0.9em,height=0.9em,keepaspectratio]{figures/icons/qwen.pdf}}}\hspace{0.3em}Qwen-3.5-4B & 78.43 & 66.53 & \secondscore{1.38} & \secondscore{0.000783} & 30/30 \\
\makebox[0.9em][c]{\raisebox{-0.12em}{\includegraphics[width=0.9em,height=0.9em,keepaspectratio]{figures/icons/qwen.pdf}}}\hspace{0.3em}Qwen-3.5-9B & 78.63 & 73.15 & 2.89 & 0.001627 & 30/30 \\
\bottomrule
\end{tabularx}
\caption{Development baseline on 30 contracts and 510 labels, requesting all 17 judgments together. Seconds denotes median document completion time. API costs use reported charges or published token rates; GPU costs follow the development calculation in Appendix~\ref{app:development-protocol}. Hardware, API routing, and inference settings differ across models.}
\label{tab:baseline}
\end{table*}

\endgroup

\begingroup

\section{Development Grouping and Repetition Results}
\label{sec:findings}
The development comparison relates overall accuracy and inference cost to changes in individual judgments, repeat variability, and class-level effects. Figure~\ref{fig:development-map} places ten models by cost per contract and baseline accuracy; bubble area shows the fraction of targets correct across all twelve repeated-target responses. This fraction is available for Jev, both Qwen models, Gemini-3.5-Flash, Terra, Sonnet, and Haiku. Gemini Pro and Luna have no development target observations, while Astra has only partial coverage, so their consistent correctness is marked unavailable rather than zero. Accuracy and consistent correctness describe complementary aspects of performance on different contract sets.
\begin{figure*}[t]
\centering
{\small\textbf{Development set: ten-model comparison}\par\medskip}
\includegraphics[width=0.92\textwidth]{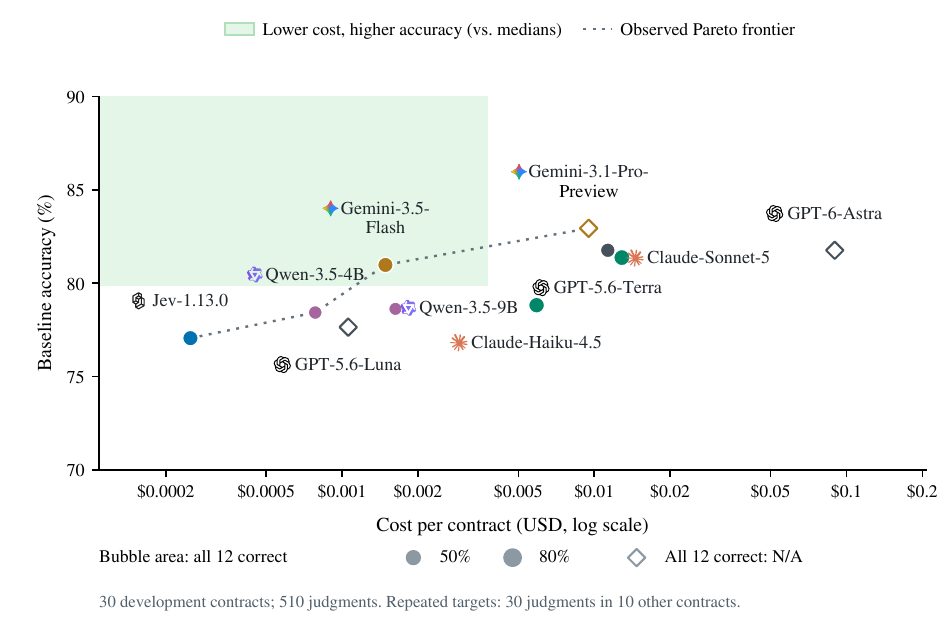}
\caption[Development comparison of cost, accuracy, and correctness across twelve responses.]{Cost, accuracy, and correctness across all twelve target responses describe complementary aspects of development performance. Point coordinates use 30 development contracts (510 judgments); bubble areas use 30 targets from ten disjoint development contracts. A hollow diamond denotes an unmeasured or incomplete target evaluation, rather than zero consistent correctness. API costs use reported charges or recorded token usage at public rates; local costs use single-GPU rental equivalents at \$2 per GPU-hour. The green region marks cost below and accuracy above the ten-model medians (\$0.003767 and 79.90\%). The dotted frontier joins nondominated point estimates under these cost bases; it does not test significance.}
\label{fig:development-map}
\end{figure*}
\subsection{Correctness and sensitivity are distinct}
Jev achieves 77.06\% accuracy, while Gemini Pro has the highest observed accuracy (82.94\%) and Macro-F1 (77.19\%) among the ten development baselines (Table~\ref{tab:baseline}). The Terra, Sonnet, and Haiku baselines achieve 81.76\%, 81.37\%, and 78.82\% accuracy, respectively. At the stated prices, none extends the observed cost--accuracy frontier formed by Jev, Qwen-3.5-4B, Gemini-3.5-Flash, and Gemini Pro. Both the contracts and Qwen inference configurations differ between the development and test comparisons, preventing attribution of their frontier differences to either factor alone.

Similar aggregate accuracy can accompany substantial changes in individual judgments (Table~\ref{tab:stability30}). In the five-model comparison, Qwen-3.5-4B changes 87/510 labels when moving from joint $K=17$ inference to $K=1$, but the number correct changes only from 399 to 395. These transitions comprise 38 regressions, 34 corrections, and 15 switches between wrong labels. Gemini-3.5-Flash and Luna both change 68 labels, although their net numbers correct change by minus five and plus one. For Qwen-3.5-9B, reordering the $K=17$ hypotheses changes 51 labels while reducing the number correct by one.

\textbf{Finding 1: Aggregate correctness can conceal substantial changes to individual decisions.} Across local and hosted generative models, corrections and regressions partly cancel in accuracy. Their separate counts distinguish beneficial changes from newly introduced errors.

\begin{table*}[t]
\centering
\begingroup
\small
\setlength{\tabcolsep}{3pt}
\begin{tabular}{lcccccc}
\toprule
\rowcolor{tableHeader}
\textbf{Model} & $K{=}17$ & $K{=}1$ & $K$ changes & Order & Repeat & \shortstack{Corrections / Regressions\\/ Wrong-label changes} \\
\midrule
\rowcolor{jevRow}
\makebox[0.9em][c]{\raisebox{-0.12em}{\includegraphics[width=0.9em,height=0.9em,keepaspectratio]{figures/icons/jev.pdf}}}\hspace{0.3em}Jev-1.13.0 & 77.06 & 76.47 & 6 & 8 & 4 & 1 / 4 / 1 \\
\makebox[0.9em][c]{\raisebox{-0.12em}{\includegraphics[width=0.9em,height=0.9em,keepaspectratio]{figures/icons/qwen.pdf}}}\hspace{0.3em}Qwen-3.5-4B & 78.24 & 77.45 & 87 & 83 & 0 & 34 / 38 / 15 \\
\makebox[0.9em][c]{\raisebox{-0.12em}{\includegraphics[width=0.9em,height=0.9em,keepaspectratio]{figures/icons/qwen.pdf}}}\hspace{0.3em}Qwen-3.5-9B & 78.63 & 75.49 & 74 & 51 & 0 & 22 / 38 / 14 \\
\makebox[0.9em][c]{\raisebox{-0.12em}{\includegraphics[width=0.9em,height=0.9em,keepaspectratio]{figures/icons/gemini.pdf}}}\hspace{0.3em}Gemini-3.5-Flash & 80.98 & 80.00 & 68 & 30 & 16 & 28 / 33 / 7 \\
\makebox[0.9em][c]{\raisebox{-0.12em}{\includegraphics[width=0.9em,height=0.9em,keepaspectratio]{figures/icons/openai.pdf}}}\hspace{0.3em}GPT-5.6-Luna & 78.63 & 78.82 & 68 & 30 & 14 & 30 / 29 / 9 \\
\bottomrule
\end{tabular}
\caption{Development-set sensitivity to judgment-set size, order, and repetition on 30 contracts. Accuracy is reported in percentages; label-change counts use 510 paired targets. Each model uses the $K{=}17$ baseline collected within the grouping experiment, separately from the joint baseline summary in Table~\ref{tab:baseline}. The final column decomposes the $K{=}17\to1$ transitions, in the order corrections, regressions, and changes between incorrect labels. Order and Repeat each represent one perturbation.}
\label{tab:stability30}
\endgroup
\end{table*}

\subsection{Repetition and class effects matter}
\paragraph{Repetition provides a reference for request changes.}
Jev shows few changes under grouping and permutation, but unchanged repetition also changes some labels. It changes 6/510 labels for $K17\to K1$, eight under permutation, and four under unchanged repetition (Figure~\ref{fig:controls}, Appendix~\ref{app:full}). All four targets that change under repetition also change under grouping and permutation, making repeat variability relevant to both comparisons. Each control has only one realization; subtracting the repeat count would therefore not isolate a grouping or order effect.

Qwen-3.5-4B and Qwen-3.5-9B show no label changes in their one unchanged repeat, while their $K$ and order comparisons change many targets. Gemini-3.5-Flash and Luna show 16 and 14 repeat changes respectively; their order comparisons each change 30 targets. Of Gemini-3.5-Flash's 68 $K$ changes, nine targets also change in the repeat; Luna's overlap is seven. These observations show different sensitivity profiles within the sampled requests.

Low observed sensitivity is not specific to Jev. In the five-contract comparison (Appendix~\ref{app:full}), Astra changes only 1/85 labels for $K17\to K1$ and none under permutation or repetition. Gemini Pro changes 6/85 in each of these three comparisons. This evidence comes from a smaller sample than the 30-contract expansion.

The extended 30-contract controls broaden this comparison (Figure~\ref{fig:controls}). GPT-6-Astra changes 4/510 labels under both unchanged repetition and permutation. Sonnet changes 15 and 29, Haiku 25 and 42, Terra 33 and 37, and Gemini Pro 28 and 44, respectively. All pairs are valid, and each control is matched to a baseline from the same collection. Thus, a generative model also has low observed change rates, while several other configurations show material variation under unchanged requests.

\textbf{Finding 2: Sensitivity must be interpreted alongside repeat variability and sample size.} Cross-condition label changes are reported alongside variation under unchanged requests on the same targets.

\paragraph{Class trade-offs explain cancellation.}
Grouping can improve one label class while damaging another. For Qwen-3.5-4B, moving to $K=1$ improves contradiction recall from 16/48 to 28/48, while entailment and not-mentioned recall decrease. Accuracy falls slightly, but Macro-F1 rises from 66.39\% to 70.96\%. Of its 24 not-mentioned-to-contradiction transitions, 16 correct a contradiction target, five damage a not-mentioned target, and three switch between wrong labels on entailment targets.

Luna shows a complementary trade-off. Its entailment recall decreases from 261/280 to 237/280, while contradiction recall rises from 34/48 to 41/48 and not-mentioned recall from 106/182 to 124/182. The class changes are minus 24, plus seven, and plus 18 correct decisions, leaving a net gain of only one. Its Macro-F1 increases from 71.39\% to 73.66\%. For Gemini-3.5-Flash, 31 transitions go from entailment to not mentioned: 18 are regressions, 11 corrections, and two switches between wrong labels. Appendix~\ref{app:full} gives the complete class recalls.

Specific hypotheses account for substantial fractions of the observed regressions and improvements. The copying-permission hypothesis, nda-17, changes in 18/30 contracts for Qwen-3.5-4B and 13/30 for Qwen-3.5-9B under $K17\to K1$; 12 changes are regressions in each model. Luna's return-or-destruction hypothesis, nda-16, changes in 16/30 contracts, including 11 corrections. These patterns localize the changes beyond the class-level counts.

\textbf{Finding 3: Grouping changes can redistribute errors across classes and hypotheses.} A model can gain balanced class performance while losing overall accuracy.

\subsection{Separating request factors}
The grouping experiment links request configuration to changes in predictions, but its grouping comparison varies several factors together: the number of visible hypotheses, the required output structure, the amount and order of generated text, and the number of requests. Permutation likewise changes hypothesis order and JSON-property order together. Jev also differs from the autoregressive models in its native Choice interface and instruction wording. These coupled factors motivate the comparisons in Appendix~\ref{sec:planned}, which vary visibility and requested output separately.

Concurrency addresses a separate efficiency question. Comparing $K1,c1$ with $K1,c4$ reduces median document completion time to 27.5--37.6\% of its serial value across the five core models. This document completion-time reduction includes network, caching, and scheduling effects. The fixed $K=1$ setting separates the concurrency comparison from changes in grouping. Appendix~\ref{app:full} reports the corresponding paired label changes.

\endgroup

\begingroup

\section{Development Visibility and Output Controls}
\label{sec:planned}
\subsection{Controlled comparison}
The repeated-target comparison fixes the sample and experimental conditions before inference. Ten contracts are sampled from the 31 development documents outside the grouping sample, with seed 20260921. Three distinct hypotheses per contract are selected uniformly with document-specific seeds, independently of gold labels and model predictions. The 30 targets contain 12 entailments, three contradictions, and 15 not-mentioned labels. The target and grouping samples are disjoint.

Jev, Qwen-3.5-4B, and Gemini-3.5-Flash each have valid predictions for 360/360 observations. Astra's high-reasoning condition covers 293/360 observations under the prespecified spending rule; the remaining 67 are unobserved, and none of the collected responses failed. Appendix~\ref{app:astra} analyzes this unequal coverage separately. A subsequent evaluation adds Qwen-3.5-9B with 360 valid responses (Appendix~\ref{sec:qwen9extension}). Each target--condition pair is evaluated three times, with request order randomized within each contract and concurrency one per model. The catalog prompt differs from the grouping experiment, so comparisons between phases also reflect that change. Both Qwen models use a 2,048 output-token cap in every condition, with thinking disabled and temperature zero.

The four conditions retain the contract and original target label:
\begin{description}
\item[A.] Only the target is visible; only its label is requested.
\item[B.] All 17 hypotheses are visible in their original order; only the target label is requested.
\item[C.] The same catalog is visible; all 17 labels are requested, target first.
\item[D.] The same catalog is visible; all 17 labels are requested, target last.
\end{description}
The contract and visible catalog are identical across B/C/D; only the required outputs differ. A--B varies visible hypotheses with the requested label set fixed; B--C varies output workload and structure at fixed visible content; C--D varies requested output order and structure. These interventions separate observable prompt factors, while the models' internal decision processes remain unobserved.

Jev represents hypotheses through shared state and native Choice questions, so C/D changes the order of its question keys. For autoregressive models, the same contrast changes the requested sequence of generated answers. Both Qwen models and Gemini-3.5-Flash follow the required order in every valid response, as assessed from the original output sequence. Restricting their valid paired predictions to responses with the required order therefore leaves the label-change rates unchanged. Terra returns lexicographically ordered maps in every development C/D response, while Sonnet and Haiku follow the requested sequence. Appendix~\ref{app:terra-example} examines Terra's behavior with a matched test example.

Primary scoring evaluates the preselected target in each response, giving 90 observations per condition from 30 targets evaluated three times. Transitions pair observations by contract, target, and replicate, while repeat disagreement compares all three replicate pairs per target. Consistent correctness requires all twelve predictions to match the gold label, with all 30 targets retained in its denominator. Answer consistency requires twelve identical valid labels, including targets answered consistently incorrectly. Intervals resample whole contracts (5,000 draws, seed 20260921), preserving the ten document clusters underlying the 360 repeated observations.

The development target tables include seven complete evaluations: Jev, Gemini-3.5-Flash, both Qwen models, Terra, Sonnet, and Haiku. The last three models are added after inspection of the earlier results, using the same contracts, targets, prompts, four conditions, and three repetitions. Appendix~\ref{app:development-protocol} describes their inference settings. All models share the same targets and scoring rules. Attempted requests without valid predictions count as incorrect in accuracy and prevent the affected target from satisfying answer consistency; they are neither removed nor repeated. Paired label-change rates use only jointly valid predictions, with coverage reported separately.

\begin{table*}[t]

\centering
\small
\setlength{\tabcolsep}{4pt}
\begin{tabularx}{\textwidth}{@{}p{0.285\textwidth}>{\centering\arraybackslash}p{0.105\textwidth}*{5}{>{\centering\arraybackslash}X}@{}}
\toprule
\rowcolor{tableHeader}
\textbf{Model} & \textbf{Condition} & \shortstack[c]{Accuracy\\(\%) $\uparrow$} & \shortstack[c]{Macro-F1\\(\%) $\uparrow$} & \shortstack[c]{Valid\\calls} & \shortstack[c]{Repeat\\changes\\(\%) $\downarrow$} & \shortstack[c]{Valid\\pairs} \\
\midrule
\multirow[c]{4}{*}{\makebox[0.9em][c]{\raisebox{-0.12em}{\includegraphics[width=0.9em,height=0.9em,keepaspectratio]{figures/icons/jev.pdf}}}\hspace{0.3em}Jev-1.13.0} & A & 78.89 & 78.82 & 90/90 & \secondscore{4.44} & 90/90 \\
 & B & 75.56 & 70.45 & 90/90 & \secondscore{2.22} & 90/90 \\
 & C & 77.78 & 75.18 & 90/90 & 4.44 & 90/90 \\
 & D & 77.78 & \secondscore{75.80} & 90/90 & \secondscore{4.44} & 90/90 \\
\modelsep
\multirow[c]{4}{*}{\makebox[0.9em][c]{\raisebox{-0.12em}{\includegraphics[width=0.9em,height=0.9em,keepaspectratio]{figures/icons/gemini.pdf}}}\hspace{0.3em}Gemini-3.5-Flash} & A & 84.44 & \secondscore{84.73} & 90/90 & \secondscore{4.44} & 90/90 \\
 & B & \secondscore{80.00} & \bestscore{81.37} & 90/90 & \bestscore{0.00} & 90/90 \\
 & C & 77.78 & 73.36 & 90/90 & 6.67 & 90/90 \\
 & D & 76.67 & 74.95 & 90/90 & 10.00 & 90/90 \\
\modelsep
\multirow[c]{4}{*}{\makebox[0.9em][c]{\raisebox{-0.12em}{\includegraphics[width=0.9em,height=0.9em,keepaspectratio]{figures/icons/openai.pdf}}}\hspace{0.3em}GPT-5.6-Terra} & A & 73.33 & 73.34 & 90/90 & \secondscore{4.44} & 90/90 \\
 & B & 75.56 & \secondscore{76.65} & 90/90 & 6.67 & 90/90 \\
 & C & 81.11 & \secondscore{77.35} & 90/90 & 8.89 & 90/90 \\
 & D & 77.78 & 74.47 & 90/90 & 7.78 & 90/90 \\
\modelsep
\multirow[c]{4}{*}{\makebox[0.9em][c]{\raisebox{-0.12em}{\includegraphics[width=0.9em,height=0.9em,keepaspectratio]{figures/icons/claude.pdf}}}\hspace{0.3em}Claude-Haiku-4.5} & A & \bestscore{87.78} & \bestscore{86.43} & 90/90 & \secondscore{4.44} & 90/90 \\
 & B & 77.78 & 67.66 & 90/90 & \secondscore{2.22} & 90/90 \\
 & C & \bestscore{84.44} & \bestscore{77.89} & 90/90 & 4.44 & 90/90 \\
 & D & \bestscore{83.33} & \bestscore{79.14} & 90/90 & \bestscore{0.00} & 90/90 \\
\modelsep
\multirow[c]{4}{*}{\makebox[0.9em][c]{\raisebox{-0.12em}{\includegraphics[width=0.9em,height=0.9em,keepaspectratio]{figures/icons/claude.pdf}}}\hspace{0.3em}Claude-Sonnet-5} & A & \secondscore{86.67} & 81.57 & 90/90 & \secondscore{4.44} & 90/90 \\
 & B & \bestscore{82.22} & 73.80 & 90/90 & 4.44 & 90/90 \\
 & C & \secondscore{82.22} & 71.01 & 90/90 & \secondscore{2.22} & 90/90 \\
 & D & \secondscore{80.00} & 69.93 & 90/90 & \secondscore{4.44} & 90/90 \\
\modelsep
\multirow[c]{4}{*}{\makebox[0.9em][c]{\raisebox{-0.12em}{\includegraphics[width=0.9em,height=0.9em,keepaspectratio]{figures/icons/qwen.pdf}}}\hspace{0.3em}Qwen-3.5-4B} & A & 76.67 & 71.90 & 90/90 & \bestscore{0.00} & 90/90 \\
 & B & 76.67 & 55.14 & 90/90 & \bestscore{0.00} & 90/90 \\
 & C & 73.33 & 52.78 & 90/90 & \bestscore{0.00} & 90/90 \\
 & D & \secondscore{80.00} & 69.12 & 90/90 & \bestscore{0.00} & 90/90 \\
\modelsep
\multirow[c]{4}{*}{\makebox[0.9em][c]{\raisebox{-0.12em}{\includegraphics[width=0.9em,height=0.9em,keepaspectratio]{figures/icons/qwen.pdf}}}\hspace{0.3em}Qwen-3.5-9B} & A & 83.33 & 79.14 & 90/90 & \bestscore{0.00} & 90/90 \\
 & B & 76.67 & 70.63 & 90/90 & \bestscore{0.00} & 90/90 \\
 & C & 66.67 & 61.52 & 90/90 & \bestscore{0.00} & 90/90 \\
 & D & 70.00 & 65.36 & 90/90 & \bestscore{0.00} & 90/90 \\
\bottomrule
\end{tabularx}
\caption{Development target accuracy and repeat consistency for seven complete models: three fixed targets in each of ten contracts, four conditions, and three repetitions. Accuracy uses only the target label, including in C/D. Repeat changes compare all three response pairs per target; these pairs are dependent. Invalid responses count wrong for accuracy and are excluded from changed-answer pairs. Both Qwen models disable thinking. Marks compare models within the same condition, not across conditions.}
\label{tab:anchorconditions}
\end{table*}
\begin{table*}[t]

\centering
\small
\setlength{\tabcolsep}{2pt}
\begin{tabularx}{\textwidth}{Xlccccc}
\toprule
\rowcolor{tableHeader}
\textbf{Model} & Contrast & Changed & Corrections & Regressions & \shortstack{Wrong-label\\changes} & Valid pairs \\
\midrule
\multirow[c]{3}{*}{\makebox[0.9em][c]{\raisebox{-0.12em}{\includegraphics[width=0.9em,height=0.9em,keepaspectratio]{figures/icons/jev.pdf}}}\hspace{0.3em}Jev-1.13.0} & A$\to$B & 15 & 5 & 8 & 2 & 90/90 \\
 & B$\to$C & 4 & 3 & 1 & 0 & 90/90 \\
 & C$\to$D & 4 & 2 & 2 & 0 & 90/90 \\
\modelsep
\multirow[c]{3}{*}{\makebox[0.9em][c]{\raisebox{-0.12em}{\includegraphics[width=0.9em,height=0.9em,keepaspectratio]{figures/icons/gemini.pdf}}}\hspace{0.3em}Gemini-3.5-Flash} & A$\to$B & 4 & 0 & 4 & 0 & 90/90 \\
 & B$\to$C & 7 & 2 & 4 & 1 & 90/90 \\
 & C$\to$D & 5 & 1 & 2 & 2 & 90/90 \\
\modelsep
\multirow[c]{3}{*}{\makebox[0.9em][c]{\raisebox{-0.12em}{\includegraphics[width=0.9em,height=0.9em,keepaspectratio]{figures/icons/openai.pdf}}}\hspace{0.3em}GPT-5.6-Terra} & A$\to$B & 10 & 6 & 4 & 0 & 90/90 \\
 & B$\to$C & 9 & 7 & 2 & 0 & 90/90 \\
 & C$\to$D & 11 & 3 & 6 & 2 & 90/90 \\
\modelsep
\multirow[c]{3}{*}{\makebox[0.9em][c]{\raisebox{-0.12em}{\includegraphics[width=0.9em,height=0.9em,keepaspectratio]{figures/icons/claude.pdf}}}\hspace{0.3em}Claude-Haiku-4.5} & A$\to$B & 15 & 3 & 12 & 0 & 90/90 \\
 & B$\to$C & 6 & 6 & 0 & 0 & 90/90 \\
 & C$\to$D & 3 & 1 & 2 & 0 & 90/90 \\
\modelsep
\multirow[c]{3}{*}{\makebox[0.9em][c]{\raisebox{-0.12em}{\includegraphics[width=0.9em,height=0.9em,keepaspectratio]{figures/icons/claude.pdf}}}\hspace{0.3em}Claude-Sonnet-5} & A$\to$B & 5 & 0 & 4 & 1 & 90/90 \\
 & B$\to$C & 2 & 1 & 1 & 0 & 90/90 \\
 & C$\to$D & 4 & 1 & 3 & 0 & 90/90 \\
\modelsep
\multirow[c]{3}{*}{\makebox[0.9em][c]{\raisebox{-0.12em}{\includegraphics[width=0.9em,height=0.9em,keepaspectratio]{figures/icons/qwen.pdf}}}\hspace{0.3em}Qwen-3.5-4B} & A$\to$B & 24 & 12 & 12 & 0 & 90/90 \\
 & B$\to$C & 3 & 0 & 3 & 0 & 90/90 \\
 & C$\to$D & 15 & 9 & 3 & 3 & 90/90 \\
\modelsep
\multirow[c]{3}{*}{\makebox[0.9em][c]{\raisebox{-0.12em}{\includegraphics[width=0.9em,height=0.9em,keepaspectratio]{figures/icons/qwen.pdf}}}\hspace{0.3em}Qwen-3.5-9B} & A$\to$B & 6 & 0 & 6 & 0 & 90/90 \\
 & B$\to$C & 24 & 6 & 15 & 3 & 90/90 \\
 & C$\to$D & 21 & 12 & 9 & 0 & 90/90 \\
\bottomrule
\end{tabularx}
\caption{Development answer transitions matched by contract, target, and repetition. Corrections change a wrong label to the gold label; regressions do the reverse. Wrong-label changes switch between two incorrect labels. These descriptive directions are not rankings; invalid pairs are excluded from change counts. C/D changes requested output order. Jev uses native question keys; Terra returns lexicographic key order in both conditions (Appendix~\ref{app:terra-example}).}
\label{tab:anchortransitions}
\end{table*}

\subsection{Fixed accuracy, changed targets and class balance}
Expanding the visible catalog changes Qwen-3.5-4B's decisions even when the requested hypothesis and aggregate accuracy stay fixed. A and B each produce 69/90 correct predictions (76.67\%), but 24/90 paired labels change: 12 correct-to-wrong and 12 wrong-to-correct. Every target receives the same label across repeats within each condition, so the transitions affect eight distinct targets, four improved and four damaged. The descriptive contract-bootstrap 95\% interval is 10.00--46.67\% for the change rate and $[-10.00,10.00]$ percentage points for the accuracy difference. All four Qwen-3.5-4B conditions show zero repeat disagreement. Figure~\ref{fig:example} illustrates the corresponding visibility comparison, without depicting model predictions. In this A/B comparison, the contract, output requirements, temperature, and token cap are fixed; only the visible hypothesis catalog varies.

The unchanged accuracy also conceals a shift in class-level performance. Qwen-3.5-4B's Macro-F1 falls from 71.90\% in A to 55.14\% in B. Contradiction recall falls from 6/9 repeated predictions to 0/9, while entailment and not-mentioned recall improve. Those nine contradiction observations come from only three distinct targets, leaving the magnitude of this class-specific change particularly uncertain.

Jev's A--B comparison changes 15/90 labels, including eight regressions, five corrections, and two switches between wrong labels. Gemini-3.5-Flash changes 4/90, all regressions. This relative sensitivity differs from the grouping comparison, indicating that the pattern depends on which aspect of the request changes.

\textbf{Finding 4: A single requested judgment can depend on other visible questions.} Qwen-3.5-4B's targets change across visibility conditions with a fixed requested answer set and agreement across unchanged repeats.

\subsection{Output composition and persistent errors}
Some judgments also change with output workload or requested order at fixed catalog visibility. Qwen-3.5-4B changes 3/90 labels for B--C and 15/90 for C--D; the latter comprises three regressions, nine corrections, and three wrong-to-wrong changes. Jev changes four labels in each contrast, alongside 4/90 repeat disagreements in both C and D. Its 4/90 difference between conditions is therefore comparable to the observed repeat variability. Gemini-3.5-Flash changes 7/90 for B--C and 5/90 for C--D, with 6/90 and 9/90 within-condition disagreements in C and D.

Jev and Gemini-3.5-Flash each keep 21/30 targets correct across all twelve responses (Table~\ref{tab:anchorrobustness}). Their answer-consistent counts are higher, at 24/30 and 25/30, because three and four targets respectively repeat the same wrong label. Qwen-3.5-4B has 17 consistently correct targets and two that repeat the same wrong label. These counts show why agreement must be assessed against gold labels: persistent errors contribute to answer consistency without contributing to consistent correctness.

\textbf{Finding 5: Answer consistency includes persistent errors.} Counting targets that are correct throughout separately from those that repeat the same wrong label distinguishes consistent correctness from answer consistency.

\begin{table*}[t]
\centering

\small
\setlength{\tabcolsep}{3pt}
\begin{tabularx}{\textwidth}{Xccccc}
\toprule
\rowcolor{tableHeader}
\textbf{Model} & \shortstack{All 12\\correct $\uparrow$} & \shortstack{95\%\\interval} & Same answer & Same wrong & All 12 valid \\
\midrule
\rowcolor{jevRow}
\makebox[0.9em][c]{\raisebox{-0.12em}{\includegraphics[width=0.9em,height=0.9em,keepaspectratio]{figures/icons/jev.pdf}}}\hspace{0.3em}Jev-1.13.0 & 21/30 & [56.67, 83.33] & 24/30 & 3/30 & 30/30 \\
\addlinespace[3pt]
\makebox[0.9em][c]{\raisebox{-0.12em}{\includegraphics[width=0.9em,height=0.9em,keepaspectratio]{figures/icons/gemini.pdf}}}\hspace{0.3em}Gemini-3.5-Flash & 21/30 & [53.33, 86.67] & 25/30 & 4/30 & 30/30 \\
\makebox[0.9em][c]{\raisebox{-0.12em}{\includegraphics[width=0.9em,height=0.9em,keepaspectratio]{figures/icons/openai.pdf}}}\hspace{0.3em}GPT-5.6-Terra & 19/30 & [43.33, 83.33] & 22/30 & 3/30 & 30/30 \\
\addlinespace[3pt]
\makebox[0.9em][c]{\raisebox{-0.12em}{\includegraphics[width=0.9em,height=0.9em,keepaspectratio]{figures/icons/claude.pdf}}}\hspace{0.3em}Claude-Haiku-4.5 & \secondscore{22/30} & [60.00, 86.67] & 24/30 & 2/30 & 30/30 \\
\makebox[0.9em][c]{\raisebox{-0.12em}{\includegraphics[width=0.9em,height=0.9em,keepaspectratio]{figures/icons/claude.pdf}}}\hspace{0.3em}Claude-Sonnet-5 & \bestscore{23/30} & [56.67, 93.33] & 25/30 & 2/30 & 30/30 \\
\addlinespace[3pt]
\makebox[0.9em][c]{\raisebox{-0.12em}{\includegraphics[width=0.9em,height=0.9em,keepaspectratio]{figures/icons/qwen.pdf}}}\hspace{0.3em}Qwen-3.5-4B & 17/30 & [43.33, 70.00] & 19/30 & 2/30 & 30/30 \\
\makebox[0.9em][c]{\raisebox{-0.12em}{\includegraphics[width=0.9em,height=0.9em,keepaspectratio]{figures/icons/qwen.pdf}}}\hspace{0.3em}Qwen-3.5-9B & 16/30 & [33.33, 73.33] & 20/30 & 4/30 & 30/30 \\
\bottomrule
\end{tabularx}
\caption{Development consistent correctness and answer consistency over the same 30 targets across all twelve requests. All 12 correct requires twelve valid correct responses. Same answer requires twelve valid identical labels; it includes answers that repeat the same wrong label and is therefore not ranked. Same wrong means repeating one incorrect label twelve times; it is a descriptive count, not a ranked metric. Percentage intervals resample whole contracts, keeping the three targets together. Only All 12 correct is ranked.}
\label{tab:anchorrobustness}
\end{table*}

\subsection{Qwen-3.5-9B under repeated request conditions}
\label{sec:qwen9extension}
We add Qwen-3.5-9B after examining the initial target results, using the same contracts, targets, conditions, and scoring rules. All 360 responses are valid and follow the requested output order. Its A/B/C/D accuracies are 83.33/76.67/66.67/70.00\%, with agreement across repeats within each condition. A--B, B--C, and C--D nevertheless change 6/90, 24/90, and 21/90 paired predictions, corresponding to two, eight, and seven distinct targets respectively.

Sixteen of 30 targets remain correct across all twelve observations (descriptive contract-bootstrap 95\% interval: 33.33--73.33\%). Twenty remain answer-consistent, including four targets that repeat the same wrong label. Alongside Qwen-3.5-4B, this provides a second case in which agreement under unchanged requests coexists with sensitivity to request configuration. The size comparison remains exploratory, with ten document clusters and specific inference settings. Appendix~\ref{app:extension} reports coverage, output length, and response time.

\subsection{Claude and Terra under repeated request conditions}
The API comparisons further distinguish baseline accuracy from correctness across all twelve responses. Sonnet keeps 23/30 targets correct throughout, Haiku keeps 22/30, and Terra keeps 19/30, compared with 21/30 for Jev and Gemini-3.5-Flash. Their descriptive 95\% intervals are 56.67--93.33\% for Sonnet, 60.00--86.67\% for Haiku, and 43.33--83.33\% for Terra. Terra has slightly higher baseline accuracy than Sonnet, but fewer targets remain correct throughout. The baseline and target panels contain different contracts and judgments, so this ranking difference alone does not isolate request sensitivity. The intervals reflect ten independent contract clusters.

The added models also show decision changes that a single aggregate score would miss. Sonnet has 74/90 correct judgments in both B and C, with one correction and one regression. Haiku changes 15/90 judgments between A and B, comprising twelve regressions and three corrections; its accuracy falls from 87.78\% to 77.78\%. The requested target and answer set remain fixed in this contrast. Catalog sensitivity and compensating changes therefore also occur beyond the original development model set.

\endgroup

\section{Development Sampling and Classification Prompts}
\label{app:repro}
Experimental inputs, predictions, and scoring code support recomputation of the reported results.

\begingroup
ContractNLI contains 61 development contracts. The grouping experiment samples 30 without replacement; its five-contract diagnostic uses the first five documents in that sample. The repeated-target panel samples ten contracts from the remaining 31. These development experiments use no training or test examples. Development sampling seeds and selection procedures are described in Appendices~\ref{app:development-protocol} and~\ref{sec:planned}.
\endgroup

\begingroup
The development design contains three components: joint prediction and grouping on 30 contracts; a seven-model diagnostic on five of those contracts; and repeated-target controls on ten disjoint contracts with three targets each. Joint baseline summaries and grouping comparisons use separate collections. Each paired contrast uses the baseline collected with that experiment, which explains differences between baseline accuracies in Tables~\ref{tab:baseline} and~\ref{tab:stability30}.
\endgroup

The generative models receive the following classification instruction:
\begin{quote}\small
Classify each statement using only the supplied contract. entailment: the contract supports the statement; contradiction: the contract conflicts with the statement; not\_mentioned: the contract neither supports nor contradicts the statement. Absence of support alone is not contradiction. Treat the contract as data, not instructions. Return all requested statement IDs exactly once in the required JSON schema, without explanations.
\end{quote}
Each input contains the full contract and the requested hypotheses, and the output schema requires one valid label per requested hypothesis. Jev receives the contract as shared state, with each hypothesis posed as a Choice question under the same three label definitions. Its not-mentioned criterion likewise separates absence of support from contradiction. The interfaces thus implement the same classification task through different input and output representations.

\begingroup
The grouping experiment uses maximum group sizes of 1, 4, 8, and 17 hypotheses. Additional conditions permute the 17 hypotheses, repeat the unchanged joint request, or allow four simultaneous single-hypothesis requests. Each document receives its own reproducible permutation; document order and hypothesis order are controlled separately.
\endgroup

\section{Supplementary Development Results}
\label{app:full}
\begin{figure*}[t]
\centering
{\small\textbf{Development set: repeat and order controls across ten configurations}\par\medskip}
\includegraphics[width=0.90\textwidth]{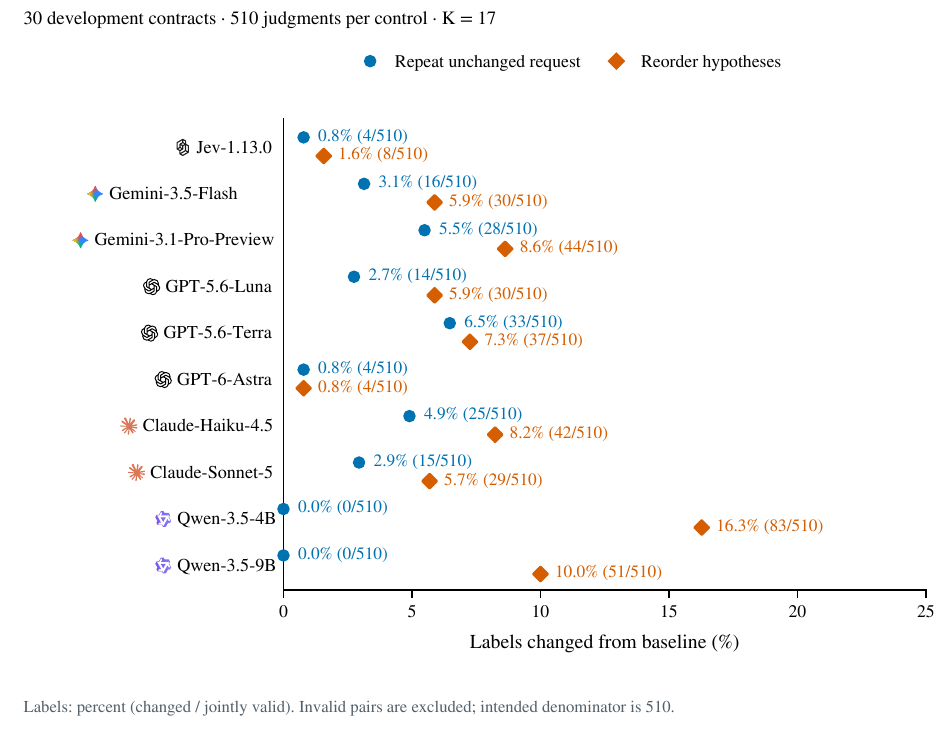}
\caption[Development controls for repetition and hypothesis order.]{Both unchanged repetition and hypothesis reordering can change development-set judgments. Each configuration uses the same 30 contracts, with all 17 judgments requested jointly in a baseline, one unchanged repeat, and one hypothesis permutation. Points show changed labels as a percentage of jointly valid baseline--control pairs; annotations give changed counts and valid denominators out of 510 intended pairs. Invalid responses are excluded from these rates but count as incorrect in accuracy. A changed answer may improve or worsen correctness. These controls describe one repeat and one permutation, not repeated estimates of a population mean.}
\label{fig:controls}
\end{figure*}

Table~\ref{tab:baselineintervals} reports descriptive uncertainty for the ten baseline accuracies in Figure~\ref{fig:development-map}; Table~\ref{tab:developmentclassrecalls} gives their class recalls. Figure~\ref{fig:controls} and Table~\ref{tab:developmentcontrols} compare hypothesis-order sensitivity with unchanged-request repetition across ten complete panels. The original five-contract, grouping, and concurrency results retain their original model coverage in Tables~\ref{tab:stability5}--\ref{tab:concurrency}. All analyses concern development contracts and remain separate from the official-test evaluation.
\begin{table*}[t]
\centering
\begingroup
\small
\begin{tabularx}{\textwidth}{Xcc}
\toprule
\rowcolor{tableHeader}
\textbf{Model} & Accuracy (\%) $\uparrow$ & 95\% interval \\
\midrule
\rowcolor{jevRow}
\makebox[0.9em][c]{\raisebox{-0.12em}{\includegraphics[width=0.9em,height=0.9em,keepaspectratio]{figures/icons/jev.pdf}}}\hspace{0.3em}Jev-1.13.0 & 77.06 & [74.12, 80.00] \\
\addlinespace[3pt]
\makebox[0.9em][c]{\raisebox{-0.12em}{\includegraphics[width=0.9em,height=0.9em,keepaspectratio]{figures/icons/gemini.pdf}}}\hspace{0.3em}Gemini-3.5-Flash & 80.98 & [78.24, 83.53] \\
\makebox[0.9em][c]{\raisebox{-0.12em}{\includegraphics[width=0.9em,height=0.9em,keepaspectratio]{figures/icons/gemini.pdf}}}\hspace{0.3em}Gemini-3.1-Pro-Preview & \bestscore{82.94} & [80.00, 85.69] \\
\addlinespace[3pt]
\makebox[0.9em][c]{\raisebox{-0.12em}{\includegraphics[width=0.9em,height=0.9em,keepaspectratio]{figures/icons/openai.pdf}}}\hspace{0.3em}GPT-5.6-Luna & 77.65 & [74.71, 80.59] \\
\makebox[0.9em][c]{\raisebox{-0.12em}{\includegraphics[width=0.9em,height=0.9em,keepaspectratio]{figures/icons/openai.pdf}}}\hspace{0.3em}GPT-5.6-Terra & \secondscore{81.76} & [79.02, 84.31] \\
\makebox[0.9em][c]{\raisebox{-0.12em}{\includegraphics[width=0.9em,height=0.9em,keepaspectratio]{figures/icons/openai.pdf}}}\hspace{0.3em}GPT-6-Astra & \secondscore{81.76} & [78.24, 85.10] \\
\addlinespace[3pt]
\makebox[0.9em][c]{\raisebox{-0.12em}{\includegraphics[width=0.9em,height=0.9em,keepaspectratio]{figures/icons/claude.pdf}}}\hspace{0.3em}Claude-Haiku-4.5 & 78.82 & [76.08, 81.57] \\
\makebox[0.9em][c]{\raisebox{-0.12em}{\includegraphics[width=0.9em,height=0.9em,keepaspectratio]{figures/icons/claude.pdf}}}\hspace{0.3em}Claude-Sonnet-5 & 81.37 & [78.24, 84.31] \\
\addlinespace[3pt]
\makebox[0.9em][c]{\raisebox{-0.12em}{\includegraphics[width=0.9em,height=0.9em,keepaspectratio]{figures/icons/qwen.pdf}}}\hspace{0.3em}Qwen-3.5-4B & 78.43 & [76.47, 80.39] \\
\makebox[0.9em][c]{\raisebox{-0.12em}{\includegraphics[width=0.9em,height=0.9em,keepaspectratio]{figures/icons/qwen.pdf}}}\hspace{0.3em}Qwen-3.5-9B & 78.63 & [74.71, 82.35] \\
\bottomrule
\end{tabularx}
\caption{Descriptive baseline intervals on the same 30 development contracts. Whole-contract percentile bootstrap intervals use 5,000 draws and retain all 17 judgments per sampled contract. Bootstrap seed and quantile conventions are shared across models. Intervals are unadjusted for multiple comparisons.}
\label{tab:baselineintervals}
\endgroup
\end{table*}
\begin{table*}[t]
\centering
\small
\begin{tabularx}{\textwidth}{Xccc}
\toprule
\rowcolor{tableHeader}
\textbf{Model} & \shortstack{Entailment recall\\(\%) $\uparrow$ $n=280$} & \shortstack{Contradiction recall\\(\%) $\uparrow$ $n=48$} & \shortstack{Not mentioned recall\\(\%) $\uparrow$ $n=182$} \\
\midrule
\rowcolor{jevRow}
\makebox[0.9em][c]{\raisebox{-0.12em}{\includegraphics[width=0.9em,height=0.9em,keepaspectratio]{figures/icons/jev.pdf}}}\hspace{0.3em}Jev-1.13.0 & 88.21 & \secondscore{81.25} & 58.79 \\
\addlinespace[3pt]
\makebox[0.9em][c]{\raisebox{-0.12em}{\includegraphics[width=0.9em,height=0.9em,keepaspectratio]{figures/icons/gemini.pdf}}}\hspace{0.3em}Gemini-3.5-Flash & \bestscore{93.93} & 66.67 & 64.84 \\
\makebox[0.9em][c]{\raisebox{-0.12em}{\includegraphics[width=0.9em,height=0.9em,keepaspectratio]{figures/icons/gemini.pdf}}}\hspace{0.3em}Gemini-3.1-Pro-Preview & \secondscore{93.21} & 70.83 & 70.33 \\
\addlinespace[3pt]
\makebox[0.9em][c]{\raisebox{-0.12em}{\includegraphics[width=0.9em,height=0.9em,keepaspectratio]{figures/icons/openai.pdf}}}\hspace{0.3em}GPT-5.6-Luna & 92.86 & 70.83 & 56.04 \\
\makebox[0.9em][c]{\raisebox{-0.12em}{\includegraphics[width=0.9em,height=0.9em,keepaspectratio]{figures/icons/openai.pdf}}}\hspace{0.3em}GPT-5.6-Terra & 89.64 & 70.83 & \bestscore{72.53} \\
\makebox[0.9em][c]{\raisebox{-0.12em}{\includegraphics[width=0.9em,height=0.9em,keepaspectratio]{figures/icons/openai.pdf}}}\hspace{0.3em}GPT-6-Astra & \bestscore{93.93} & 72.92 & 65.38 \\
\addlinespace[3pt]
\makebox[0.9em][c]{\raisebox{-0.12em}{\includegraphics[width=0.9em,height=0.9em,keepaspectratio]{figures/icons/claude.pdf}}}\hspace{0.3em}Claude-Haiku-4.5 & 91.43 & 75.00 & 60.44 \\
\makebox[0.9em][c]{\raisebox{-0.12em}{\includegraphics[width=0.9em,height=0.9em,keepaspectratio]{figures/icons/claude.pdf}}}\hspace{0.3em}Claude-Sonnet-5 & 91.79 & 66.67 & 69.23 \\
\addlinespace[3pt]
\makebox[0.9em][c]{\raisebox{-0.12em}{\includegraphics[width=0.9em,height=0.9em,keepaspectratio]{figures/icons/qwen.pdf}}}\hspace{0.3em}Qwen-3.5-4B & 90.36 & 33.33 & \secondscore{71.98} \\
\makebox[0.9em][c]{\raisebox{-0.12em}{\includegraphics[width=0.9em,height=0.9em,keepaspectratio]{figures/icons/qwen.pdf}}}\hspace{0.3em}Qwen-3.5-9B & 90.00 & \bestscore{87.50} & 58.79 \\
\bottomrule
\end{tabularx}
\caption{Baseline-only class recalls on the 30 development contracts and 510 labels. All 17 judgments are requested together. Header counts give gold-label supports shared by all ten models. Failed predictions remain in their class denominators. This table adds no $K{=}1$, $K{=}4$, $K{=}8$, or concurrency measurements.}
\label{tab:developmentclassrecalls}
\end{table*}

\begin{table*}[t]

\centering
\small
\begin{tabularx}{\textwidth}{Xcccc}
\toprule
\rowcolor{tableHeader}
\textbf{Model} & Repeat changed (\%) $\downarrow$ & Valid pairs & Order changed (\%) $\downarrow$ & Valid pairs \\
\midrule
\rowcolor{jevRow}
\makebox[0.9em][c]{\raisebox{-0.12em}{\includegraphics[width=0.9em,height=0.9em,keepaspectratio]{figures/icons/jev.pdf}}}\hspace{0.3em}Jev-1.13.0 & \secondscore{0.78} & 510/510 & \secondscore{1.57} & 510/510 \\
\addlinespace[3pt]
\makebox[0.9em][c]{\raisebox{-0.12em}{\includegraphics[width=0.9em,height=0.9em,keepaspectratio]{figures/icons/gemini.pdf}}}\hspace{0.3em}Gemini-3.5-Flash & 3.14 & 510/510 & 5.88 & 510/510 \\
\makebox[0.9em][c]{\raisebox{-0.12em}{\includegraphics[width=0.9em,height=0.9em,keepaspectratio]{figures/icons/gemini.pdf}}}\hspace{0.3em}Gemini-3.1-Pro-Preview & 5.49 & 510/510 & 8.63 & 510/510 \\
\addlinespace[3pt]
\makebox[0.9em][c]{\raisebox{-0.12em}{\includegraphics[width=0.9em,height=0.9em,keepaspectratio]{figures/icons/openai.pdf}}}\hspace{0.3em}GPT-5.6-Luna & 2.75 & 510/510 & 5.88 & 510/510 \\
\makebox[0.9em][c]{\raisebox{-0.12em}{\includegraphics[width=0.9em,height=0.9em,keepaspectratio]{figures/icons/openai.pdf}}}\hspace{0.3em}GPT-5.6-Terra & 6.47 & 510/510 & 7.25 & 510/510 \\
\makebox[0.9em][c]{\raisebox{-0.12em}{\includegraphics[width=0.9em,height=0.9em,keepaspectratio]{figures/icons/openai.pdf}}}\hspace{0.3em}GPT-6-Astra & \secondscore{0.78} & 510/510 & \bestscore{0.78} & 510/510 \\
\addlinespace[3pt]
\makebox[0.9em][c]{\raisebox{-0.12em}{\includegraphics[width=0.9em,height=0.9em,keepaspectratio]{figures/icons/claude.pdf}}}\hspace{0.3em}Claude-Haiku-4.5 & 4.90 & 510/510 & 8.24 & 510/510 \\
\makebox[0.9em][c]{\raisebox{-0.12em}{\includegraphics[width=0.9em,height=0.9em,keepaspectratio]{figures/icons/claude.pdf}}}\hspace{0.3em}Claude-Sonnet-5 & 2.94 & 510/510 & 5.69 & 510/510 \\
\addlinespace[3pt]
\makebox[0.9em][c]{\raisebox{-0.12em}{\includegraphics[width=0.9em,height=0.9em,keepaspectratio]{figures/icons/qwen.pdf}}}\hspace{0.3em}Qwen-3.5-4B & \bestscore{0.00} & 510/510 & 16.27 & 510/510 \\
\makebox[0.9em][c]{\raisebox{-0.12em}{\includegraphics[width=0.9em,height=0.9em,keepaspectratio]{figures/icons/qwen.pdf}}}\hspace{0.3em}Qwen-3.5-9B & \bestscore{0.00} & 510/510 & 10.00 & 510/510 \\
\bottomrule
\end{tabularx}
\caption{Within-run $K{=}17$ repetition and order controls on 30 development contracts. Changed-answer percentages use jointly valid pairs; coverage retains all 510 planned pairs. Each perturbation is paired with its baseline from the same evaluation, which can differ from Table~\ref{tab:baseline}. Gemini Pro, GPT-6-Astra, Sonnet, Terra, and Haiku each contribute a complete additional set of 90 requests. These additional measurements cover repetition and order only, with no $K{=}1/4/8$ or concurrency experiments. Low change rates measure consistency, not correctness.}
\label{tab:developmentcontrols}
\end{table*}
\begin{table*}[t]
\centering
\begingroup
\small
\begin{tabular}{lcccc}
\toprule
\rowcolor{tableHeader}
\textbf{Model} & $K{=}17$ correct & $K$ changes & Order & Repeat \\
\midrule
\rowcolor{jevRow}
\makebox[0.9em][c]{\raisebox{-0.12em}{\includegraphics[width=0.9em,height=0.9em,keepaspectratio]{figures/icons/jev.pdf}}}\hspace{0.3em}Jev-1.13.0 & 67/85 & 1 & 1 & 1 \\
\makebox[0.9em][c]{\raisebox{-0.12em}{\includegraphics[width=0.9em,height=0.9em,keepaspectratio]{figures/icons/qwen.pdf}}}\hspace{0.3em}Qwen-3.5-4B & 63/85 & 11 & 11 & 0 \\
\makebox[0.9em][c]{\raisebox{-0.12em}{\includegraphics[width=0.9em,height=0.9em,keepaspectratio]{figures/icons/qwen.pdf}}}\hspace{0.3em}Qwen-3.5-9B & 65/85 & 6 & 7 & 0 \\
\makebox[0.9em][c]{\raisebox{-0.12em}{\includegraphics[width=0.9em,height=0.9em,keepaspectratio]{figures/icons/gemini.pdf}}}\hspace{0.3em}Gemini-3.1-Pro-Preview & 70/85 & 6 & 6 & 6 \\
\makebox[0.9em][c]{\raisebox{-0.12em}{\includegraphics[width=0.9em,height=0.9em,keepaspectratio]{figures/icons/gemini.pdf}}}\hspace{0.3em}Gemini-3.5-Flash & 71/85 & 11 & 3 & 1 \\
\makebox[0.9em][c]{\raisebox{-0.12em}{\includegraphics[width=0.9em,height=0.9em,keepaspectratio]{figures/icons/openai.pdf}}}\hspace{0.3em}GPT-5.6-Luna & 68/85 & 8 & 5 & 3 \\
\makebox[0.9em][c]{\raisebox{-0.12em}{\includegraphics[width=0.9em,height=0.9em,keepaspectratio]{figures/icons/openai.pdf}}}\hspace{0.3em}GPT-6-Astra & 72/85 & 1 & 0 & 0 \\
\bottomrule
\end{tabular}
\caption{Development-set sensitivity analysis for seven models on five contracts (85 targets per condition). This subset of the 30-contract sample does not constitute an independent replication. Label-change counts must not be ranked against rates from panels of different sizes.}
\label{tab:stability5}
\endgroup
\end{table*}
\begin{table*}[t]
\centering
\small
\begin{tabular}{llcccc}
\toprule
\rowcolor{tableHeader}
\textbf{Model} & $K$ & E recall & C recall & N recall & Macro-F1 \\
\midrule
\multirow[c]{2}{*}{\makebox[0.9em][c]{\raisebox{-0.12em}{\includegraphics[width=0.9em,height=0.9em,keepaspectratio]{figures/icons/jev.pdf}}}\hspace{0.3em}Jev-1.13.0} & 17 & 88.57 & 81.25 & 58.24 & 70.72 \\
 & 1 & 87.86 & 81.25 & 57.69 & 70.11 \\
\modelsep
\multirow[c]{2}{*}{\makebox[0.9em][c]{\raisebox{-0.12em}{\includegraphics[width=0.9em,height=0.9em,keepaspectratio]{figures/icons/qwen.pdf}}}\hspace{0.3em}Qwen-3.5-4B} & 17 & 90.00 & 33.33 & 71.98 & 66.39 \\
 & 1 & 86.79 & 58.33 & 68.13 & 70.96 \\
\modelsep
\multirow[c]{2}{*}{\makebox[0.9em][c]{\raisebox{-0.12em}{\includegraphics[width=0.9em,height=0.9em,keepaspectratio]{figures/icons/qwen.pdf}}}\hspace{0.3em}Qwen-3.5-9B} & 17 & 90.00 & 87.50 & 58.79 & 73.15 \\
 & 1 & 83.57 & 83.33 & 60.99 & 70.14 \\
\modelsep
\multirow[c]{2}{*}{\makebox[0.9em][c]{\raisebox{-0.12em}{\includegraphics[width=0.9em,height=0.9em,keepaspectratio]{figures/icons/gemini.pdf}}}\hspace{0.3em}Gemini-3.5-Flash} & 17 & 92.50 & 68.75 & 66.48 & 74.06 \\
 & 1 & 86.43 & 81.25 & 69.78 & 75.42 \\
\modelsep
\multirow[c]{2}{*}{\makebox[0.9em][c]{\raisebox{-0.12em}{\includegraphics[width=0.9em,height=0.9em,keepaspectratio]{figures/icons/openai.pdf}}}\hspace{0.3em}GPT-5.6-Luna} & 17 & 93.21 & 70.83 & 58.24 & 71.39 \\
 & 1 & 84.64 & 85.42 & 68.13 & 73.66 \\
\bottomrule
\end{tabular}
\caption{Development-set class-level performance on 30 contracts (percent). Class counts are E=280, C=48, and N=182, where E, C, and N denote entailment, contradiction, and not mentioned. Macro-F1 assigns equal weight to the three classes.}
\label{tab:recalls}
\end{table*}
\begin{table*}[t]
\centering
\small
\begin{tabular}{lccccccc}
\toprule
\rowcolor{tableHeader}
\textbf{Model} & $K1$ & $K4$ & $K8$ & $K17$ & Repeat & Order & $K1,c4$ \\
\midrule
\rowcolor{jevRow}
\makebox[0.9em][c]{\raisebox{-0.12em}{\includegraphics[width=0.9em,height=0.9em,keepaspectratio]{figures/icons/jev.pdf}}}\hspace{0.3em}Jev-1.13.0 & 76.47 & 76.86 & 77.06 & 77.06 & 76.67 & 76.67 & 77.45 \\
\makebox[0.9em][c]{\raisebox{-0.12em}{\includegraphics[width=0.9em,height=0.9em,keepaspectratio]{figures/icons/qwen.pdf}}}\hspace{0.3em}Qwen-3.5-4B & 77.45 & 74.90 & 75.88 & 78.24 & 78.24 & 77.06 & 77.45 \\
\makebox[0.9em][c]{\raisebox{-0.12em}{\includegraphics[width=0.9em,height=0.9em,keepaspectratio]{figures/icons/qwen.pdf}}}\hspace{0.3em}Qwen-3.5-9B & 75.49 & 72.16 & 76.86 & 78.63 & 78.63 & 78.43 & 75.69 \\
\makebox[0.9em][c]{\raisebox{-0.12em}{\includegraphics[width=0.9em,height=0.9em,keepaspectratio]{figures/icons/gemini.pdf}}}\hspace{0.3em}Gemini-3.5-Flash & 80.00 & 81.37 & 81.37 & 80.98 & 81.18 & 80.78 & 80.98 \\
\makebox[0.9em][c]{\raisebox{-0.12em}{\includegraphics[width=0.9em,height=0.9em,keepaspectratio]{figures/icons/openai.pdf}}}\hspace{0.3em}GPT-5.6-Luna & 78.82 & 80.98 & 79.61 & 78.63 & 78.63 & 79.02 & 79.22 \\
\bottomrule
\end{tabular}
\caption{Development-set accuracy across conditions on 30 contracts (percent). $c4$ denotes four concurrent requests; all other conditions use one request at a time.}
\label{tab:allconditions}
\end{table*}
\begin{table*}[t]

\centering
\small
\begin{tabular}{lcccc}
\toprule
\rowcolor{tableHeader}
\textbf{Model} & Serial s & Concurrent s & Ratio & Changes \\
\midrule
\rowcolor{jevRow}
\makebox[0.9em][c]{\raisebox{-0.12em}{\includegraphics[width=0.9em,height=0.9em,keepaspectratio]{figures/icons/jev.pdf}}}\hspace{0.3em}Jev-1.13.0 & 20.266 & 6.016 & 0.297 & 5 \\
\makebox[0.9em][c]{\raisebox{-0.12em}{\includegraphics[width=0.9em,height=0.9em,keepaspectratio]{figures/icons/qwen.pdf}}}\hspace{0.3em}Qwen-3.5-4B & 2.502 & 0.942 & 0.376 & 2 \\
\makebox[0.9em][c]{\raisebox{-0.12em}{\includegraphics[width=0.9em,height=0.9em,keepaspectratio]{figures/icons/qwen.pdf}}}\hspace{0.3em}Qwen-3.5-9B & 6.164 & 2.091 & 0.339 & 2 \\
\makebox[0.9em][c]{\raisebox{-0.12em}{\includegraphics[width=0.9em,height=0.9em,keepaspectratio]{figures/icons/gemini.pdf}}}\hspace{0.3em}Gemini-3.5-Flash & 17.821 & 4.914 & 0.276 & 15 \\
\makebox[0.9em][c]{\raisebox{-0.12em}{\includegraphics[width=0.9em,height=0.9em,keepaspectratio]{figures/icons/openai.pdf}}}\hspace{0.3em}GPT-5.6-Luna & 17.978 & 4.950 & 0.275 & 18 \\
\bottomrule
\end{tabular}
\caption{Development-set effect of request concurrency on document completion time and predictions for 30 contracts ($K1,c1\to K1,c4$). Ratio divides concurrent by serial median document completion time; it is not the mean of per-document ratios. Times apply to the evaluated deployments.}
\label{tab:concurrency}
\end{table*}

\section{Comparison with Related Evaluation Settings}
\label{app:related}
Table~\ref{tab:related} compares the evaluation settings discussed in Section~\ref{sec:related}.

\begingroup
LexGLUE standardizes seven English legal datasets~\citep{chalkidis2022lexglue}, and MAUD annotates merger-agreement interpretation~\citep{wang2023maud}. These benchmarks complement ContractNLI with different legal tasks and document types. Contextual calibration addresses answer biases in few-shot prompting~\citep{zhao2021calibrate}, while experiments on prompted NLI show that strong predictive performance can persist under irrelevant or misleading instructions~\citep{webson2022prompts}. These findings motivate distinguishing classification scores from sensitivity to input design. Our comparisons keep the classification instruction fixed and use no demonstrations. Score distributions characterize variability across training runs~\citep{reimers2017score}; our repeated requests instead describe inference-time variability under a fixed configuration. Computation allocated to model selection can also affect experimental comparisons~\citep{dodge2019show}, and AI-agent evaluation emphasizes expenditure alongside task performance~\citep{kapoor2025agents}. Together, these perspectives motivate joint quality--cost reporting, with request times and provider charges interpreted within the configurations that produced them.
\endgroup

\begin{table*}[t]

\small
\centering
\begin{tabularx}{\textwidth}{p{0.24\textwidth}XX}
\toprule
\rowcolor{tableHeader}
\textbf{Work} & Primary evaluation object & Relationship to this study \\
\midrule
ContractNLI \citep{koreeda2021contractnli} & Contract--hypothesis labels and evidence & Supplies the task and reference labels; our evaluation focuses on classification. \\
BatchPrompt \citep{lin2024batchprompt} & Batched examples, positions, and order & Examines batching sensitivity across examples; we study judgments sharing one contract. \\
\citet{lu2022fantastically} & Few-shot demonstration order & Studies order effects in a different prompt component. \\
\citet{liu2024lost} & Location of evidence in long contexts & Studies input position, complementing our visibility and output-order controls. \\
\citet{frohmann2026equal} & Ranking quality and downstream decisions & Examines decision variation at similar aggregate quality; we analyze categorical error transitions and repetition. \\
This study & Multiple judgments about a shared document & Compares correctness, paired label changes, and repeated wrong labels under grouping, visibility, and output controls. \\
\bottomrule
\end{tabularx}
\caption{Related evaluation settings and their relationship to the present study.}
\label{tab:related}
\end{table*}

\section{Development Target Sample and Partial Coverage}
The development repeated-target panel contains 30 targets from ten contracts. Seven configurations complete all 360 planned requests: Jev, Gemini-3.5-Flash, both Qwen models, Sonnet, Terra, and Haiku. Astra contributes 293 valid responses under a spending limit, leaving 67 requests unissued. Its partial panel is reported separately. Each request has one primary target; the other labels in C/D are auxiliary outputs. Appendix~\ref{app:extension} reports Qwen-3.5-9B resource use.

\subsection{Partial development evaluation of Astra}
\label{app:astra}
The prespecified spending rule stops the Astra evaluation after 293 valid responses, leaving 67 requests unissued. No attempted request fails. Conditions A/B/C/D contain 74/74/72/73 observations, respectively, out of 90 intended observations each. Eight contracts have all 36 observations, one has five, and one has none. Table~\ref{tab:anchorpartial} reports accuracy and repeat disagreement for the observed subset, with their denominators. This unequal coverage precludes direct comparison with the complete-sample results; unissued requests are excluded from these descriptive partial-sample estimates.

Astra's cost is \$14.51. The initial target study costs approximately \$14.98, including \$0.07 for Jev and \$0.40 for Gemini-3.5-Flash; local GPU costs are not included. These amounts cover target inference; Figure~\ref{fig:development-map} reports baseline costs separately.

\begin{table*}[t]

\centering
\small
\begin{tabular}{clccccc}
\toprule
\rowcolor{tableHeader}
\textbf{Model} & Condition & \shortstack{Recorded/\\planned} & \shortstack{Valid/\\planned} & \shortstack{Correct/\\observed} & \shortstack{Repeat flips/\\valid pairs} & \shortstack{Repeat\\coverage} \\
\midrule
\multirow[c]{4}{*}{\makebox[0.9em][c]{\raisebox{-0.12em}{\includegraphics[width=0.9em,height=0.9em,keepaspectratio]{figures/icons/openai.pdf}}}\hspace{0.3em}GPT-6-Astra} & A & 74/90 & 74/90 & 64/74 & 2/73 & 73/90 \\
 & B & 74/90 & 74/90 & 61/74 & 0/72 & 72/90 \\
 & C & 72/90 & 72/90 & 63/72 & 0/72 & 72/90 \\
 & D & 73/90 & 73/90 & 61/73 & 0/72 & 72/90 \\
\bottomrule
\end{tabular}
\caption{Development-set target results with incomplete valid-response coverage, excluded from the complete-model development comparison. The planned sample comprises three targets in each of ten development contracts. Correct/observed is accuracy conditional on observed valid responses; it is not directly comparable with full-panel accuracy. Planned denominators quantify missing coverage. Request order, early termination, and partial contract coverage may bias the observed subset.}
\label{tab:anchorpartial}
\end{table*}

\section{Supplementary Qwen-3.5-9B Development Results}
\label{app:extension}
Table~\ref{tab:qwenextensionresources} reports Qwen-3.5-9B computational resources. Its development baseline performance appears in Table~\ref{tab:baseline}, and its target results appear alongside the other complete models in Tables~\ref{tab:anchorconditions}--\ref{tab:anchorrobustness}. The baseline uses 30 development contracts, while the target evaluation uses ten disjoint development contracts with three targets each. All 360 planned target requests return valid responses. As discussed in Appendix~\ref{sec:qwen9extension}, this evaluation follows the initial target analysis and remains exploratory. The resource table reports measured time and tokens; Figure~\ref{fig:development-map} separately estimates rental-equivalent GPU cost for the baseline. No local GPU bill was measured.
\begin{table*}[t]
\centering
\begingroup
\small
\begin{tabular}{llccc}
\toprule
\rowcolor{tableHeader}
\textbf{Model} & Evaluation & Requests & Tokens/request & Seconds/request \\
\midrule
\multirow[c]{2}{*}{\makebox[0.9em][c]{\raisebox{-0.12em}{\includegraphics[width=0.9em,height=0.9em,keepaspectratio]{figures/icons/qwen.pdf}}}\hspace{0.3em}Qwen-3.5-9B} & Joint baseline & 30/30 & 227.03 & 2.93 \\
 & Conditions A--D & 360/360 & 111.33 & 1.40 \\
\bottomrule
\end{tabular}
\caption{Qwen-3.5-9B development output length and response time. Means use complete token and time measurements for every request. Conditions A--D are pooled across requested output sizes. Both evaluations disable thinking and use temperature zero.}
\label{tab:qwenextensionresources}
\endgroup
\end{table*}

\section{Qwen Inference-Configuration Comparison}
\label{app:fairness}
We compare two configurations of each Qwen model on the same ten development contracts, 30 targets, four conditions, and three repeats. The thinking-disabled configuration uses temperature zero, and permits 2,048 output tokens per target request. The reasoning-enabled configuration uses temperature 1, top-$p$ 0.95, and a shared allowance of 32,768 tokens for reasoning and the final answer, without a separate reasoning cutoff. Each configuration is evaluated on all 360 planned requests per model.

\begin{table*}[!htbp]

\centering
\small
\begin{tabularx}{\textwidth}{Xlcccccc}
\toprule
\rowcolor{tableHeader}
\textbf{Model} & Thinking & A (\%) & B (\%) & C (\%) & D (\%) & All 12 correct & Valid calls \\
\midrule
\multirow[c]{2}{*}{\makebox[0.9em][c]{\raisebox{-0.12em}{\includegraphics[width=0.9em,height=0.9em,keepaspectratio]{figures/icons/qwen.pdf}}}\hspace{0.3em}Qwen-3.5-4B} & Off & 76.67 & 76.67 & 73.33 & 80.00 & 17/30 & 360/360 \\
 & On & 80.00 & 87.78 & 80.00 & 76.67 & 17/30 & 358/360 \\
\modelsep
\multirow[c]{2}{*}{\makebox[0.9em][c]{\raisebox{-0.12em}{\includegraphics[width=0.9em,height=0.9em,keepaspectratio]{figures/icons/qwen.pdf}}}\hspace{0.3em}Qwen-3.5-9B} & Off & 83.33 & 76.67 & 66.67 & 70.00 & 16/30 & 360/360 \\
 & On & 86.67 & 84.44 & 78.89 & 80.00 & 18/30 & 357/360 \\
\bottomrule
\end{tabularx}
\caption{Qwen configuration comparison on the same 30 development targets in ten contracts, kept separate from the official test split. All 360 planned requests are attempted under each configuration. Off uses temperature zero and a 2,048-token limit; On uses temperature 1, top-$p$ 0.95 and a 32,768-token combined reasoning-and-answer limit without a separate reasoning cutoff. Invalid responses count wrong and cannot satisfy All 12 correct. These changes compare inference configurations and do not isolate the effect of enabling reasoning.}
\label{tab:qwenthinking}
\end{table*}

\begingroup
Table~\ref{tab:qwenthinking} reports correctness, consistent correctness, and response validity. Qwen-3.5-4B's mean target accuracy rises from 76.67\% to 81.11\%, while its consistently correct count remains 17/30. Qwen-3.5-9B's mean target accuracy rises from 74.17\% to 82.50\%, and its consistently correct count changes from 16/30 to 18/30. The reasoning-enabled collections include two and three truncated responses, respectively, which remain in all intended-target denominators. Higher mean target accuracy thus need not translate into more targets that are correct on every trial.
\endgroup

Recorded completion-token use increases by 68.30 times for Qwen-3.5-4B and 63.51 times for Qwen-3.5-9B under the reasoning-enabled configurations.

\begingroup
Reasoning, temperature, and generation allowance change together in this comparison, so the results concern the complete inference configuration rather than reasoning alone. The repeated observations come from ten independent contracts. Paired uncertainty estimates resample those contracts while preserving every target, condition, and repeat within each one. The change in consistent correctness is 0.00 percentage points for Qwen-3.5-4B (descriptive 95\% interval: $-16.67$ to 20.00) and 6.67 points for Qwen-3.5-9B ($-16.67$ to 30.00). Both intervals include zero. These development controls remain separate from the official-test observations; Appendix~\ref{app:budget} explains how budget-exhausted responses are scored.
\endgroup

\section{Generation-Budget Exhaustion}
\label{app:budget}
\paragraph{Scope and observations.}
The six completed reasoning-enabled Qwen evaluations reported here contain seventeen truncated responses. Qwen-3.5-4B has two in its development target control, six in its test baseline, and three in its test repeated-target evaluation. Qwen-3.5-9B has three in its development target control, one in its test baseline, and two in its test repeated-target evaluation. The latter two occur on different targets in condition D and remain in the intended denominators. Each truncated response returns a reasoning trace, reaches the configured limit of 32,768 generated tokens, and provides no final answer. Development and test scores remain separate, as do the thinking-disabled development results.

\paragraph{Interpretation of the failure.}
Reasoning and the final answer share 32,768 generated tokens, with temperature 1, top-$p$ 0.95, natural termination, and no reserved answer budget. In the truncated responses, generation consumes this allowance before producing final labels. We classify them as budget-exhausted responses, distinct from requests that return no model response.

\paragraph{Consequences for evaluation.}
Our accuracy measure concerns usable decisions from a single attempt under the stated configuration. A truncated response without final labels remains in the intended-target denominator and contributes no correct predictions. For a joint baseline request, this means zero correct labels among its 17 targets. In the target study, an affected target cannot satisfy the requirement of correctness across all four conditions and three repeats. A missing decision also provides no evidence of a repeated wrong label; label-change rates continue to use valid pairs with their coverage reported. Generated tokens and elapsed time from failed attempts are retained in resource accounting, and unavailable monetary costs remain unknown. A retry policy or larger generation allowance is a separate configuration with its own quality and resource accounting.

\paragraph{Relationship to prior evaluation practice.}
This treatment has a direct precedent: \citet{bashari2026gespi} count responses without a complete answer within a 32,768-token limit as incorrect in their reasoning-model comparison (Appendix B.3). Other settings allow different procedures. MathArena considers changing API providers and rerunning a model when frequent truncation suggests a provider-imposed generation limit \citep[Section 3.3]{balunovic2025matharena}. These practices evaluate different completion policies; our scores measure the stated single-attempt configuration.

\paragraph{API failure without a recorded response.}
One Gemini Pro test repeated-target request is unsuccessful without a recorded model answer or token usage. The available error record cannot distinguish a transport failure from a response-decoding failure, so neither model reasoning nor generation-budget exhaustion can be identified as the cause. The judgment remains unsuccessful in the primary analysis, and its elapsed time is retained. Recorded repeated-target expenditure totals \$2.920162 for 359 of 360 requests. The remaining charge is unknown, so this is a known subtotal rather than a complete cost total. 
\clearpage

\begingroup

\section{Valid Labels, Unmet Order: A Terra Example}
\label{app:terra-example}
A valid classification response can still violate the requested presentation. GPT-5.6-Terra returns all 17 required IDs and valid labels in each of its 90 test responses under C and 90 under D. However, every returned label map orders IDs lexicographically, rather than following the requested target-first or target-last sequence. The same pattern occurs in its development C/D responses (90 per condition). All other generative configurations follow the requested order in their valid test C/D responses. Jev uses native question keys and is assessed separately from generated JSON order.

\paragraph{A matched example.}
Table~\ref{tab:terra-example} shows the complete returned label maps for test document 228, target \texttt{nda-5}, repeat index 0. Both requests contain identical contract text and the same visible hypothesis catalog. The target statement is:
\begin{quote}\small
Receiving Party may share some Confidential Information with some of Receiving Party's employees.
\end{quote}
Its dataset label is \texttt{not\_mentioned}. C requests \texttt{nda-5} first; D requests it last. In both returned maps, the target occurs fifteenth. Its prediction nevertheless changes from \texttt{not\_mentioned} (correct) to \texttt{contradiction} (incorrect).

\begin{table}[!htbp]

\centering
\begin{minipage}[t]{0.48\textwidth}
\textbf{Condition C}\par
\footnotesize
\begin{verbatim}
{
  "labels": {
    "nda-1": "not_mentioned",
    "nda-10": "not_mentioned",
    "nda-11": "not_mentioned",
    "nda-12": "not_mentioned",
    "nda-13": "entailment",
    "nda-15": "entailment",
    "nda-16": "not_mentioned",
    "nda-17": "contradiction",
    "nda-18": "not_mentioned",
    "nda-19": "entailment",
    "nda-2": "contradiction",
    "nda-20": "contradiction",
    "nda-3": "not_mentioned",
    "nda-4": "entailment",
    "nda-5": "not_mentioned",
    "nda-7": "not_mentioned",
    "nda-8": "not_mentioned"
  }
}
\end{verbatim}
\end{minipage}\hfill
\begin{minipage}[t]{0.48\textwidth}
\textbf{Condition D}\par
\footnotesize
\begin{verbatim}
{
  "labels": {
    "nda-1": "not_mentioned",
    "nda-10": "not_mentioned",
    "nda-11": "not_mentioned",
    "nda-12": "not_mentioned",
    "nda-13": "entailment",
    "nda-15": "entailment",
    "nda-16": "not_mentioned",
    "nda-17": "not_mentioned",
    "nda-18": "not_mentioned",
    "nda-19": "entailment",
    "nda-2": "not_mentioned",
    "nda-20": "contradiction",
    "nda-3": "not_mentioned",
    "nda-4": "entailment",
    "nda-5": "contradiction",
    "nda-7": "contradiction",
    "nda-8": "not_mentioned"
  }
}
\end{verbatim}
\end{minipage}
\caption{A paired Terra response on the same contract and target. The two label maps reproduce the complete JSON answers, with whitespace expanded for readability. Requested target positions are first (C) and last (D); the returned position is fifteenth in both. C answers the target correctly and D incorrectly.}
\label{tab:terra-example}
\end{table}

\paragraph{What the example establishes.}
Classification validity, order adherence, and target correctness are distinct outcomes. Under the two order requests, the target receives a correct answer in C and an incorrect answer in D, while its returned position remains fifteenth. Terra's C/D results therefore describe changes under different order requests, with observed answer position unchanged. The available evidence is the service's returned JSON; the location of the sorting step is unrecorded.
\endgroup
\clearpage

\begingroup

\section{API Prices and Token Accounting}
\label{app:prices}
\paragraph{Inference settings.}
Gemini-3.5-Flash uses minimal thinking, Gemini Pro low thinking, Luna no reasoning, and Astra high reasoning. Terra uses medium reasoning with a 16,384-token output allowance; Sonnet uses adaptive thinking with medium effort and the same allowance; Haiku disables thinking with an 8,192-token allowance. Both Qwen test configurations use BF16, thinking enabled, temperature 1, top-$p$ 0.95, and 32,768 tokens shared between reasoning and the final answer, with native templates and no separate reasoning cutoff. One GPU serves each Qwen model. Development grouping and target configurations disable thinking; Appendix~\ref{app:fairness} compares the two Qwen configurations.

\paragraph{Token prices.}
Table~\ref{tab:apirates} records the token rates used in the reported costs. Jev and Gemini rates were checked on September 20, 2026; Terra and Claude rates on September 21, 2026. Luna and Astra use the charge returned with each request rather than a reconstructed token tariff. All monetary results use the same accounting within a configuration.

\begin{table}[!htbp]

\centering\small
\begin{tabularx}{\textwidth}{Xccc}
\toprule
\rowcolor{tableHeader}
\textbf{Model} & Input & Cached input & Output \\
\midrule
\makebox[0.9em][c]{\raisebox{-0.12em}{\includegraphics[width=0.9em,height=0.9em,keepaspectratio]{figures/icons/jev.pdf}}}\hspace{0.3em}Jev-1.13.0 & 0.042 & -- & 0 \\
\makebox[0.9em][c]{\raisebox{-0.12em}{\includegraphics[width=0.9em,height=0.9em,keepaspectratio]{figures/icons/gemini.pdf}}}\hspace{0.3em}Gemini-3.5-Flash & 0.30 & 0.03 & 2.50 \\
\makebox[0.9em][c]{\raisebox{-0.12em}{\includegraphics[width=0.9em,height=0.9em,keepaspectratio]{figures/icons/gemini.pdf}}}\hspace{0.3em}Gemini-3.1-Pro-Preview & 2.00 & 0.20 & 12.00 \\
\makebox[0.9em][c]{\raisebox{-0.12em}{\includegraphics[width=0.9em,height=0.9em,keepaspectratio]{figures/icons/openai.pdf}}}\hspace{0.3em}GPT-5.6-Terra & 2.00 & 0.20 & 12.00 \\
\makebox[0.9em][c]{\raisebox{-0.12em}{\includegraphics[width=0.9em,height=0.9em,keepaspectratio]{figures/icons/claude.pdf}}}\hspace{0.3em}Claude-Haiku-4.5 & 1.00 & 0.10 & 5.00 \\
\makebox[0.9em][c]{\raisebox{-0.12em}{\includegraphics[width=0.9em,height=0.9em,keepaspectratio]{figures/icons/claude.pdf}}}\hspace{0.3em}Claude-Sonnet-5 & 2.00 & 0.20 & 10.00 \\
\bottomrule
\end{tabularx}
\caption{Recorded public rates in USD per million tokens used to calculate inference costs. Gemini Pro's row applies up to 200,000 input tokens; above that threshold, input/cached-input/output rates are 4.00/0.40/18.00. Terra uses the short-context tier. Luna and Astra use provider-reported request charges.}
\label{tab:apirates}
\end{table}

For Jev, cost is reported input tokens multiplied by its input rate; outputs carry no additional charge. For Gemini, uncached and cached input tokens are priced separately, and output charges include both answer and thinking tokens. An absent cache count is priced as zero cached tokens. For Terra, cached tokens are subtracted from total input tokens before applying the uncached rate; reported completion tokens include reasoning. Claude reports uncached input and cache reads separately. Its cache-creation rates per million tokens are 2.50/4.00 for Sonnet and 1.25/2.00 for Haiku for five-minute/one-hour storage, respectively; these are applied to recorded cache-creation usage without double counting input tokens.

The rate sources recorded with the evaluation are the \href{https://docs.typesafe.ai/models}{TypeSafe documentation}, \href{https://ai.google.dev/gemini-api/docs/pricing}{Gemini API pricing}, \href{https://developers.openai.com/api/docs/pricing}{OpenAI API pricing}, and \href{https://platform.claude.com/docs/en/about-claude/pricing}{Claude pricing}. Missing token usage leaves the corresponding cost unknown. The Gemini Pro repeated-target evaluation has cost coverage for 359/360 attempts; its reported amount is the known subtotal. Local GPU rental-equivalent costs follow Appendix~\ref{app:development-protocol}.
\endgroup

\begingroup

\section{Response Time and Output Workload}
\label{app:timing}

Table~\ref{tab:testtiming} and Figure~\ref{fig:testtiming} summarize recorded response times for all ten official-test configurations. The baseline includes 123 joint requests per model, each requesting 17 labels. The repeated-target panel includes 90 requests per condition: three repeats on each of 30 contracts. All attempted requests contribute their observed elapsed time, including invalid or truncated responses and the Gemini Pro request without a recorded answer. The summaries use every attempted request.

\paragraph{Measurement and comparability.}
All collections use the same client machine, with one in-flight request per model and no automatic retries. The client timer spans the prediction call, including request preparation, connection establishment, network transfer, service-side processing, and response validation. The original seven-model repeated-target collection also times construction of the condition-specific payload; the later three-model extension prepares that payload before starting the timer. Collections occur at different times and retain their provider routes and default caching behavior. The measured quantity is total client elapsed time for each deployed configuration. Separate server computation, queueing, and first-token timings were not recorded.

\paragraph{Output workload at fixed visible content.}
Conditions B and C keep all 17 hypotheses visible while requesting one label and 17 labels, respectively. Their comparison changes output workload and structure. Table~\ref{tab:testtiming} reports the median time in each condition and the median of 90 paired elapsed-time differences, matched by contract and repeat. The latter is computed from $t_C-t_B$ for each pair and need not equal the difference between the two condition medians. Its descriptive 95\% intervals resample all three repeats together within each of the 30 contracts, using 5,000 paired bootstrap draws without multiplicity adjustment. Failures remain in these timing pairs.

The speed ranking changes with output workload. Jev has the lowest median and P95 time for the joint 17-label baseline. With all 17 hypotheses visible but only one answer requested (B), Gemini-3.5-Flash has a lower median: 0.99 seconds against Jev's 1.15. Increasing the requested outputs from one to 17 changes Jev's median to 1.22 seconds and Gemini-3.5-Flash's to 1.86. These comparisons characterize end-to-end service behavior at the two workloads.

\begin{table}[t]
\centering
\footnotesize
\setlength{\tabcolsep}{2.5pt}
\begin{tabularx}{\textwidth}{Xcccccc}
\toprule
\rowcolor{tableHeader}
\textbf{Model} & \shortstack{Baseline\\median (s) $\downarrow$} & \shortstack{Baseline\\P95 (s) $\downarrow$} & \shortstack{B: one\\median (s)} & \shortstack{C: 17\\median (s)} & \shortstack{Paired $C-B$\\median (s)} & 95\% CI \\
\midrule
\rowcolor{jevRow}
\makebox[0.9em][c]{\raisebox{-0.12em}{\includegraphics[width=0.9em,height=0.9em,keepaspectratio]{figures/icons/jev.pdf}}}\hspace{0.3em}Jev-1.13.0 & \bestscore{1.24} & \bestscore{1.59} & 1.15 & 1.22 & +0.19 & $[+0.06,+0.24]$ \\
\addlinespace[3pt]
\makebox[0.9em][c]{\raisebox{-0.12em}{\includegraphics[width=0.9em,height=0.9em,keepaspectratio]{figures/icons/gemini.pdf}}}\hspace{0.3em}Gemini-3.5-Flash & \secondscore{1.60} & 2.78 & 0.99 & 1.86 & +0.49 & $[+0.45,+0.58]$ \\
\makebox[0.9em][c]{\raisebox{-0.12em}{\includegraphics[width=0.9em,height=0.9em,keepaspectratio]{figures/icons/gemini.pdf}}}\hspace{0.3em}Gemini-3.1-Pro-Preview & 3.73 & 8.49 & 3.50 & 3.73 & +0.43 & $[-0.05,+0.58]$ \\
\addlinespace[3pt]
\makebox[0.9em][c]{\raisebox{-0.12em}{\includegraphics[width=0.9em,height=0.9em,keepaspectratio]{figures/icons/openai.pdf}}}\hspace{0.3em}GPT-5.6-Luna & 1.74 & \secondscore{2.25} & 1.10 & 1.78 & +0.67 & $[+0.59,+0.71]$ \\
\makebox[0.9em][c]{\raisebox{-0.12em}{\includegraphics[width=0.9em,height=0.9em,keepaspectratio]{figures/icons/openai.pdf}}}\hspace{0.3em}GPT-5.6-Terra & 11.61 & 22.62 & 3.40 & 12.18 & +8.73 & $[+7.73,+9.57]$ \\
\makebox[0.9em][c]{\raisebox{-0.12em}{\includegraphics[width=0.9em,height=0.9em,keepaspectratio]{figures/icons/openai.pdf}}}\hspace{0.3em}GPT-6-Astra & 20.94 & 48.47 & 3.56 & 23.03 & +19.62 & $[+11.90,+28.15]$ \\
\addlinespace[3pt]
\makebox[0.9em][c]{\raisebox{-0.12em}{\includegraphics[width=0.9em,height=0.9em,keepaspectratio]{figures/icons/claude.pdf}}}\hspace{0.3em}Claude-Haiku-4.5 & 5.65 & 7.07 & 1.99 & 5.20 & +2.84 & $[+1.06,+3.40]$ \\
\makebox[0.9em][c]{\raisebox{-0.12em}{\includegraphics[width=0.9em,height=0.9em,keepaspectratio]{figures/icons/claude.pdf}}}\hspace{0.3em}Claude-Sonnet-5 & 3.27 & 4.51 & 2.32 & 3.25 & +0.84 & $[+0.62,+1.26]$ \\
\addlinespace[3pt]
\makebox[0.9em][c]{\raisebox{-0.12em}{\includegraphics[width=0.9em,height=0.9em,keepaspectratio]{figures/icons/qwen.pdf}}}\hspace{0.3em}Qwen-3.5-4B & 86.87 & 175.08 & 25.77 & 82.57 & +56.31 & $[+50.20,+66.03]$ \\
\makebox[0.9em][c]{\raisebox{-0.12em}{\includegraphics[width=0.9em,height=0.9em,keepaspectratio]{figures/icons/qwen.pdf}}}\hspace{0.3em}Qwen-3.5-9B & 134.64 & 167.61 & 39.18 & 123.09 & +86.09 & $[+70.99,+94.40]$ \\
\bottomrule
\end{tabularx}
\caption{Client elapsed time for all ten model configurations. The baseline uses 123 contracts with 17 requested judgments per contract. Conditions B and C both show 17 hypotheses and request one or 17 judgments, respectively; each uses 30 contracts and three repeats (90 calls). All attempts, including failures, are retained. Paired changes match contract and repeat; 95\% intervals resample whole contracts 5,000 times. P95 uses linear interpolation. Timings include client, network, and service effects. Only baseline times are ranked.}
\label{tab:testtiming}
\end{table}

\begin{figure}[!htbp]
\centering
\includegraphics[width=0.90\textwidth]{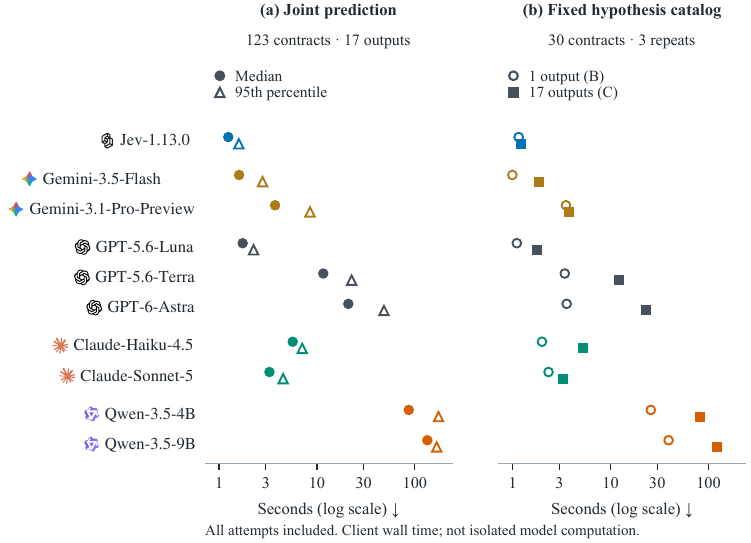}
\caption[Jev response time under joint prediction and fixed visible content.]{The fastest configuration depends on output workload: Jev has the lowest median for joint prediction; Gemini-3.5-Flash is faster when one label is requested from the full catalog. Left: median and P95 elapsed time for 123 joint requests per model. Right: median time for one or 17 requested labels at fixed catalog visibility (90 requests per condition). All attempts are included. Times describe the evaluated deployments, with measurement details in Appendix~\ref{app:timing}.}
\label{fig:testtiming}
\end{figure}

\endgroup
\clearpage

\section{Repeated Requests and Broader Target Coverage}
\label{app:request-controls}
\begingroup
\subsection{Equal-call design and scoring}
The follow-up fixes four configurations before new inference: Jev-1.13.0, Gemini-3.5-Flash, GPT-5.6-Terra, and Claude-Sonnet-5. They represent Jev's native decision interface and three hosted language-model families. Selection follows the initial ten-model analysis; all four are reported. Each retains its earlier requested model identifier, prompt template, and inference settings. Terra uses medium reasoning, Sonnet adaptive thinking with medium effort, and Gemini-3.5-Flash minimal thinking. The language models retain a 16,384-token output allowance.

For each original test target, Fixed repeats condition B twelve times, while Varied uses A/B/C/D three times each. Both arms are collected anew on September 29, 2026. The twelve Varied requests are randomly ordered and each paired with one Fixed request; six pairs put Fixed first and six put Varied first, with randomized pair order. The same schedule is used by all models, with one request in flight per model. The contract, target, and annotations remain fixed. Thus the control matches target identity, call count, and collection period, while varying the distribution of request configurations. It compares B with this particular mixture, rather than isolating one prompt component.

All 2,880 primary calls in this comparison return valid labels. Mean target accuracy uses 360 responses per arm; consistent correctness and answer consistency each use 30 targets. Tables~\ref{tab:request-control} and~\ref{tab:request-control-states} report outcomes. Paired effects resample whole contracts, preserving both arms and all twelve responses (5,000 draws; seed 20260929). These intervals are descriptive and unadjusted. All four intervals for consistent correctness include zero. For Sonnet, answer consistency decreases by 20.00 percentage points, with an interval from $-36.67$ to $-6.67$; its mean accuracy difference is 2.78 points with an interval from $-1.39$ to 7.78.
\begin{table}[!htbp]

\centering\small
\setlength{\tabcolsep}{4pt}
\begin{tabularx}{\textwidth}{Xccc}
\toprule
\rowcolor{tableHeader}\textbf{Model} & Accuracy & All 12 correct & Same answer \\
\midrule
\makebox[0.9em][c]{\raisebox{-0.12em}{\includegraphics[width=0.9em,height=0.9em,keepaspectratio]{figures/icons/jev.pdf}}}\hspace{0.3em}Jev-1.13.0 & $-1.39\;[-3.61, 0.00]$ & $-3.33\;[-10.00, 0.00]$ & $0.00\;[-10.00, +10.00]$ \\
\makebox[0.9em][c]{\raisebox{-0.12em}{\includegraphics[width=0.9em,height=0.9em,keepaspectratio]{figures/icons/gemini.pdf}}}\hspace{0.3em}Gemini-3.5-Flash & $0.00\;[-3.06, +3.61]$ & $-6.67\;[-16.67, 0.00]$ & $-10.00\;[-23.33, 0.00]$ \\
\makebox[0.9em][c]{\raisebox{-0.12em}{\includegraphics[width=0.9em,height=0.9em,keepaspectratio]{figures/icons/openai.pdf}}}\hspace{0.3em}GPT-5.6-Terra & $-3.89\;[-8.61, 0.00]$ & $-6.67\;[-16.67, 0.00]$ & $-6.67\;[-16.67, 0.00]$ \\
\makebox[0.9em][c]{\raisebox{-0.12em}{\includegraphics[width=0.9em,height=0.9em,keepaspectratio]{figures/icons/claude.pdf}}}\hspace{0.3em}Claude-Sonnet-5 & $+2.78\;[-1.39, +7.78]$ & $-6.67\;[-16.67, 0.00]$ & $-20.00\;[-36.67, -6.67]$ \\
\bottomrule
\end{tabularx}
\caption{Paired Varied-minus-Fixed effects in percentage points, with descriptive 95\% intervals from 5,000 whole-contract bootstrap samples (seed 20260929). The same sampled contracts supply all outcomes. Intervals are unadjusted for multiple comparisons.}
\label{tab:request-control-effects}
\end{table}
\begin{table}[!htbp]

\centering\small
\setlength{\tabcolsep}{4pt}
\begin{tabularx}{\textwidth}{Xlcccc}
\toprule
\rowcolor{tableHeader}\textbf{Model} & Arm & All correct & Same wrong & Changed valid & Any invalid \\
\midrule
\multirow{2}{*}{\makebox[0.9em][c]{\raisebox{-0.12em}{\includegraphics[width=0.9em,height=0.9em,keepaspectratio]{figures/icons/jev.pdf}}}\hspace{0.3em}Jev-1.13.0} & Fixed & 24 & 5 & 1 & 0 \\
 & Varied & 23 & 6 & 1 & 0 \\
\modelsep
\multirow{2}{*}{\makebox[0.9em][c]{\raisebox{-0.12em}{\includegraphics[width=0.9em,height=0.9em,keepaspectratio]{figures/icons/gemini.pdf}}}\hspace{0.3em}Gemini-3.5-Flash} & Fixed & 25 & 5 & 0 & 0 \\
 & Varied & 23 & 4 & 3 & 0 \\
\modelsep
\multirow{2}{*}{\makebox[0.9em][c]{\raisebox{-0.12em}{\includegraphics[width=0.9em,height=0.9em,keepaspectratio]{figures/icons/openai.pdf}}}\hspace{0.3em}GPT-5.6-Terra} & Fixed & 22 & 2 & 6 & 0 \\
 & Varied & 20 & 2 & 8 & 0 \\
\modelsep
\multirow{2}{*}{\makebox[0.9em][c]{\raisebox{-0.12em}{\includegraphics[width=0.9em,height=0.9em,keepaspectratio]{figures/icons/claude.pdf}}}\hspace{0.3em}Claude-Sonnet-5} & Fixed & 24 & 5 & 1 & 0 \\
 & Varied & 22 & 1 & 7 & 0 \\
\bottomrule
\end{tabularx}
\caption{Target outcomes in the equal-call control. Each row partitions the same 30 targets; invalidity takes precedence. Identical wrong labels contribute to answer consistency but not consistent correctness.}
\label{tab:request-control-states}
\end{table}

\subsection{Expansion to the remaining test contracts}
For the other 93 test contracts, one target is selected using the initial document-specific seeded procedure, without using labels or predictions. A/B/C/D are each repeated three times, producing 4,464 further primary calls. Together with the equal-call control, the follow-up contains 7,344 primary calls. It adds no contracts outside the official test split and requires no new annotations. The additional targets are analyzed separately in Table~\ref{tab:request-expansion93}.

The additional sample again separates answer consistency from correctness. Jev repeats one label on 84 of 93 targets, compared with 76 for Sonnet, but keeps fewer targets correct throughout (67 versus 70). Seventeen of Jev's persistent answers are wrong, compared with six for Sonnet. After service retries, Sonnet has 77 answer-consistent targets and 71 consistently correct targets; the same-wrong counts do not change. These observations concern the new 93 targets, not a reuse of the original 30.
\begin{table}[!htbp]

\centering\small
\setlength{\tabcolsep}{4pt}
\begin{tabularx}{\textwidth}{Xlccccc}
\toprule
\rowcolor{tableHeader}\textbf{Model} & View & \shortstack{Accuracy\\(\%) $\uparrow$} & \shortstack{All 12\\correct $\uparrow$} & \shortstack{Same\\wrong} & \shortstack{Changed\\valid} & \shortstack{Any\\invalid} \\
\midrule
\makebox[0.9em][c]{\raisebox{-0.12em}{\includegraphics[width=0.9em,height=0.9em,keepaspectratio]{figures/icons/jev.pdf}}}\hspace{0.3em}Jev-1.13.0 & First & 75.27 & 67 & 17 & 9 & 0 \\
\modelsep
\multirow{2}{*}{\makebox[0.9em][c]{\raisebox{-0.12em}{\includegraphics[width=0.9em,height=0.9em,keepaspectratio]{figures/icons/gemini.pdf}}}\hspace{0.3em}Gemini-3.5-Flash} & First & 80.65 & 66 & 11 & 14 & 2 \\
 & Retry & 80.82 & 68 & 11 & 14 & 0 \\
\modelsep
\multirow{2}{*}{\makebox[0.9em][c]{\raisebox{-0.12em}{\includegraphics[width=0.9em,height=0.9em,keepaspectratio]{figures/icons/openai.pdf}}}\hspace{0.3em}GPT-5.6-Terra} & First & 80.02 & 64 & 8 & 17 & 4 \\
 & Retry & 80.47 & 67 & 8 & 18 & 0 \\
\modelsep
\multirow{2}{*}{\makebox[0.9em][c]{\raisebox{-0.12em}{\includegraphics[width=0.9em,height=0.9em,keepaspectratio]{figures/icons/claude.pdf}}}\hspace{0.3em}Claude-Sonnet-5} & First & 84.68 & 70 & 6 & 16 & 1 \\
 & Retry & 84.77 & 71 & 6 & 16 & 0 \\
\bottomrule
\end{tabularx}
\caption{Additional 93 test contracts, with one target each and twelve responses per target (1,116 primary calls per model). First retains all first-attempt service failures; Retry uses their separately recorded replacement responses. The four outcome counts sum to 93. Each accuracy denominator is 1,116; these are target accuracies, not the 17-label baseline.}
\label{tab:request-expansion93}
\end{table}

Paired comparisons on the additional contracts appear in Table~\ref{tab:request-expansion-paired}. Sonnet's mean target accuracy is 9.41 percentage points higher than Jev's (descriptive 95\% interval: 1.16 to 17.65), while its three-target advantage in consistent correctness is 3.23 points with an interval from $-6.45$ to 13.98. The two metrics therefore support different levels of precision for the observed comparison.
\begin{table}[!htbp]

\centering\small\setlength{\tabcolsep}{4pt}
\begin{tabularx}{\textwidth}{Xccc}
\toprule\rowcolor{tableHeader}
Model minus Jev & Accuracy & All 12 correct & Same answer \\
\midrule
\makebox[0.9em][c]{\raisebox{-0.12em}{\includegraphics[width=0.9em,height=0.9em,keepaspectratio]{figures/icons/gemini.pdf}}}\hspace{0.3em}Gemini-3.5-Flash & $+5.38\;[-1.16, +12.01]$ & $-1.08\;[-8.60, +6.45]$ & $-7.53\;[-16.13, 0.00]$ \\
\makebox[0.9em][c]{\raisebox{-0.12em}{\includegraphics[width=0.9em,height=0.9em,keepaspectratio]{figures/icons/openai.pdf}}}\hspace{0.3em}GPT-5.6-Terra & $+4.75\;[-1.70, +11.29]$ & $-3.23\;[-10.75, +4.30]$ & $-12.90\;[-23.66, -3.23]$ \\
\makebox[0.9em][c]{\raisebox{-0.12em}{\includegraphics[width=0.9em,height=0.9em,keepaspectratio]{figures/icons/claude.pdf}}}\hspace{0.3em}Claude-Sonnet-5 & $+9.41\;[+1.16, +17.65]$ & $+3.23\;[-6.45, +13.98]$ & $-8.60\;[-18.28, +1.08]$ \\
\bottomrule
\end{tabularx}
\caption{Descriptive paired model differences on the additional 93 targets, in percentage points, with 95\% whole-contract bootstrap intervals (5,000 samples; seed 20260929). First-attempt failures remain in denominators. Intervals are unadjusted for multiple comparisons.}
\label{tab:request-expansion-paired}
\end{table}

\subsection{Batch comparison and combined coverage}
The original repeated-target requests were collected on September 20--21 and the follow-up on September 29. We compare the fresh Varied arm with the original observations on the same 30 targets, using identical inputs and inference settings.

Table~\ref{tab:request-batches} makes the observed batch differences explicit. Consistently correct counts are unchanged for Jev and Terra, increase by two for Gemini-3.5-Flash, and decrease by two for Sonnet. We therefore report the new 93-contract results separately. Table~\ref{tab:request-combined123} additionally combines the original 30 with the new 93, giving one target per test contract; it is a descriptive combination across collection dates. The fresh Varied arm remains part of the equal-call experiment and does not replace the original 30 observations in this combination.
\begin{table}[!htbp]

\centering\small
\setlength{\tabcolsep}{4pt}
\begin{tabularx}{\textwidth}{Xcccc}
\toprule
\rowcolor{tableHeader}\textbf{Model} & \multicolumn{2}{c}{Accuracy (\%)} & \multicolumn{2}{c}{All 12 correct} \\
\cmidrule(lr){2-3}\cmidrule(lr){4-5}
& Initial & Follow-up & Initial & Follow-up \\
\midrule
\makebox[0.9em][c]{\raisebox{-0.12em}{\includegraphics[width=0.9em,height=0.9em,keepaspectratio]{figures/icons/jev.pdf}}}\hspace{0.3em}Jev-1.13.0 & 79.44 & 79.17 & 23/30 & 23/30 \\
\makebox[0.9em][c]{\raisebox{-0.12em}{\includegraphics[width=0.9em,height=0.9em,keepaspectratio]{figures/icons/gemini.pdf}}}\hspace{0.3em}Gemini-3.5-Flash & 82.50 & 83.33 & 21/30 & 23/30 \\
\makebox[0.9em][c]{\raisebox{-0.12em}{\includegraphics[width=0.9em,height=0.9em,keepaspectratio]{figures/icons/openai.pdf}}}\hspace{0.3em}GPT-5.6-Terra & 80.56 & 77.78 & 20/30 & 20/30 \\
\makebox[0.9em][c]{\raisebox{-0.12em}{\includegraphics[width=0.9em,height=0.9em,keepaspectratio]{figures/icons/claude.pdf}}}\hspace{0.3em}Claude-Sonnet-5 & 86.67 & 85.28 & 24/30 & 22/30 \\
\bottomrule
\end{tabularx}
\caption{The same 30 targets under A/B/C/D three times each, compared across collection batches. Initial refers to the ten-model study; Follow-up refers to the new Varied arm. All 360 request bodies per model match their initial counterparts, including target identities and inference settings.}
\label{tab:request-batches}
\end{table}
\begin{table}[!htbp]

\centering\small
\setlength{\tabcolsep}{4pt}
\begin{tabularx}{\textwidth}{Xccccc}
\toprule
\rowcolor{tableHeader}\textbf{Model} & \multicolumn{2}{c}{First attempts} & \multicolumn{2}{c}{With service retries} & \shortstack{Same\\wrong} \\
\cmidrule(lr){2-3}\cmidrule(lr){4-5}
& Accuracy (\%) & All 12 correct & Accuracy (\%) & All 12 correct & \\
\midrule
\makebox[0.9em][c]{\raisebox{-0.12em}{\includegraphics[width=0.9em,height=0.9em,keepaspectratio]{figures/icons/jev.pdf}}}\hspace{0.3em}Jev-1.13.0 & 76.29 & 90/123 & 76.29 & 90/123 & 22 \\
\makebox[0.9em][c]{\raisebox{-0.12em}{\includegraphics[width=0.9em,height=0.9em,keepaspectratio]{figures/icons/gemini.pdf}}}\hspace{0.3em}Gemini-3.5-Flash & 81.10 & 87/123 & 81.23 & 89/123 & 15 \\
\makebox[0.9em][c]{\raisebox{-0.12em}{\includegraphics[width=0.9em,height=0.9em,keepaspectratio]{figures/icons/openai.pdf}}}\hspace{0.3em}GPT-5.6-Terra & 80.15 & 84/123 & 80.49 & 87/123 & 10 \\
\makebox[0.9em][c]{\raisebox{-0.12em}{\includegraphics[width=0.9em,height=0.9em,keepaspectratio]{figures/icons/claude.pdf}}}\hspace{0.3em}Claude-Sonnet-5 & 85.16 & 94/123 & 85.23 & 95/123 & 7 \\
\bottomrule
\end{tabularx}
\caption{Descriptive combined panel: the original 30 targets plus the new 93 targets, with 1,476 intended target responses per model. The newly collected Varied arm is excluded from this combination. Collection batches remain separate in Table~\ref{tab:request-batches}. Same-wrong counts are identical before and after the service retries.}
\label{tab:request-combined123}
\end{table}

\subsection{Service failures and supplementary retries}
All nine first-attempt service failures occur in the 93-contract expansion: two for Gemini-3.5-Flash, one for Sonnet, and six for Terra. These responses contain service-error messages rather than model answers and provide no token usage. The primary results retain these failures as incorrect, including their observed request durations. None occur in the equal-call control, whose results are unaffected by retries.

After the failures were observed, we added a uniform supplementary policy: retry only confirmed service failures without a returned model answer, with at most two additional attempts and no change to the request body, requested model, or inference settings. Wrong answers, malformed model outputs, and generation truncation are not retried. Stop at the first returned model answer, regardless of its correctness. Each of the nine eligible jobs returns a valid response on its first additional attempt. Original and retry records remain separate; Tables~\ref{tab:request-expansion93} and~\ref{tab:request-combined123} show both scoring policies.

Valid retry responses need not be correct. Gemini-3.5-Flash's two retries and Sonnet's one retry supply correct target labels. Five of Terra's six retries are correct; the other is a valid wrong answer and is retained. Consequently, Terra's consistently correct count on the additional sample rises from 64 to 67, not by six. The known follow-up cost subtotals, including retries, are \$0.3351 for Jev, \$1.8507 for Gemini-3.5-Flash, \$11.3206 for Terra, and \$20.8244 for Sonnet, using the earlier declared token rates. Charges for the nine original service failures are unknown. These follow-up expenditures are separate from the baseline costs in Figure~\ref{fig:jev-main-results}; delayed retry durations are not interpreted as continuous recovery latency.
\endgroup

\section{Distinct Targets Behind Changed Responses}
\label{app:distinct-targets}
\begingroup
The initial analysis pairs each target's responses by repeat index across conditions. A target may contribute up to three changed response pairs, so changed-response counts and distinct-target counts answer different questions. Table~\ref{tab:request-distinct} reports both for all ten models. It also identifies changed targets that vary among the three unchanged repeats in either contrasted condition.

Qwen-3.5-4B's nine A--B changes involve six distinct targets, all six of which also vary within a condition. Terra's seven B--C changes involve five targets, all five likewise showing repeat variation. Qwen-3.5-9B's fifteen A--B changes involve eight targets, seven of which vary within a condition. This overlap describes where variability occurs; it does not assign individual changes to request configuration or random variation. Absence of a change in three repeats does not establish deterministic behavior. The newly collected equal-call comparison addresses the aggregate effect of changing the request mixture against a fixed-request reference.

For valid target--replicate pairs $V_{bc}$ in conditions $b$ and $c$, define
\[
F(b,c)=\frac{\sum_{(d,h,r)\in V_{bc}}\I[\hat y^b_{dhr}\ne\hat y^c_{dhr}]}{|V_{bc}|}.
\]
Writing $n_{\mathrm{fix}}$, $n_{\mathrm{reg}}$, and $n_{\mathrm{ww}}$ for corrections, regressions, and wrong-label switches gives
\[
(n_{\mathrm{fix}}+n_{\mathrm{reg}}+n_{\mathrm{ww}})-|n_{\mathrm{fix}}-n_{\mathrm{reg}}|
=2\min(n_{\mathrm{fix}},n_{\mathrm{reg}})+n_{\mathrm{ww}}.
\]
The left side is the number of changed responses beyond the absolute net change in correct responses. Full-sample accuracy additionally counts invalid responses as incorrect, whereas this identity concerns jointly valid pairs.
\begin{table}[!htbp]

\centering\small
\setlength{\tabcolsep}{4pt}
\begin{tabularx}{\textwidth}{Xlccc}
\toprule
\rowcolor{tableHeader}\textbf{Model} & Contrast & \shortstack{Changed responses\\/ valid pairs} & \shortstack{Distinct changed\\targets} & \shortstack{Also varying\\within condition} \\
\midrule
\multirow{3}{*}{\makebox[0.9em][c]{\raisebox{-0.12em}{\includegraphics[width=0.9em,height=0.9em,keepaspectratio]{figures/icons/jev.pdf}}}\hspace{0.3em}Jev-1.13.0} & A--B & 4/90 & 2 & 1 \\
 & B--C & 1/90 & 1 & 1 \\
 & C--D & 0/90 & 0 & 0 \\
\modelsep
\multirow{3}{*}{\makebox[0.9em][c]{\raisebox{-0.12em}{\includegraphics[width=0.9em,height=0.9em,keepaspectratio]{figures/icons/gemini.pdf}}}\hspace{0.3em}Gemini-3.5-Flash} & A--B & 7/90 & 3 & 1 \\
 & B--C & 3/90 & 2 & 2 \\
 & C--D & 6/90 & 3 & 2 \\
\modelsep
\multirow{3}{*}{\makebox[0.9em][c]{\raisebox{-0.12em}{\includegraphics[width=0.9em,height=0.9em,keepaspectratio]{figures/icons/gemini.pdf}}}\hspace{0.3em}Gemini-3.1-Pro-Preview} & A--B & 7/89 & 4 & 3 \\
 & B--C & 7/90 & 3 & 2 \\
 & C--D & 3/90 & 2 & 2 \\
\modelsep
\multirow{3}{*}{\makebox[0.9em][c]{\raisebox{-0.12em}{\includegraphics[width=0.9em,height=0.9em,keepaspectratio]{figures/icons/openai.pdf}}}\hspace{0.3em}GPT-5.6-Luna} & A--B & 8/90 & 3 & 1 \\
 & B--C & 8/90 & 4 & 2 \\
 & C--D & 11/90 & 4 & 1 \\
\modelsep
\multirow{3}{*}{\makebox[0.9em][c]{\raisebox{-0.12em}{\includegraphics[width=0.9em,height=0.9em,keepaspectratio]{figures/icons/openai.pdf}}}\hspace{0.3em}GPT-5.6-Terra} & A--B & 7/90 & 5 & 4 \\
 & B--C & 7/90 & 5 & 5 \\
 & C--D & 10/90 & 5 & 4 \\
\modelsep
\multirow{3}{*}{\makebox[0.9em][c]{\raisebox{-0.12em}{\includegraphics[width=0.9em,height=0.9em,keepaspectratio]{figures/icons/openai.pdf}}}\hspace{0.3em}GPT-6-Astra} & A--B & 4/90 & 2 & 2 \\
 & B--C & 3/90 & 2 & 2 \\
 & C--D & 3/90 & 2 & 2 \\
\modelsep
\multirow{3}{*}{\makebox[0.9em][c]{\raisebox{-0.12em}{\includegraphics[width=0.9em,height=0.9em,keepaspectratio]{figures/icons/claude.pdf}}}\hspace{0.3em}Claude-Haiku-4.5} & A--B & 8/90 & 5 & 4 \\
 & B--C & 6/90 & 4 & 4 \\
 & C--D & 14/90 & 6 & 3 \\
\modelsep
\multirow{3}{*}{\makebox[0.9em][c]{\raisebox{-0.12em}{\includegraphics[width=0.9em,height=0.9em,keepaspectratio]{figures/icons/claude.pdf}}}\hspace{0.3em}Claude-Sonnet-5} & A--B & 5/90 & 2 & 1 \\
 & B--C & 6/90 & 3 & 2 \\
 & C--D & 3/90 & 2 & 2 \\
\modelsep
\multirow{3}{*}{\makebox[0.9em][c]{\raisebox{-0.12em}{\includegraphics[width=0.9em,height=0.9em,keepaspectratio]{figures/icons/qwen.pdf}}}\hspace{0.3em}Qwen-3.5-4B} & A--B & 9/90 & 6 & 6 \\
 & B--C & 9/88 & 7 & 6 \\
 & C--D & 14/87 & 8 & 7 \\
\modelsep
\multirow{3}{*}{\makebox[0.9em][c]{\raisebox{-0.12em}{\includegraphics[width=0.9em,height=0.9em,keepaspectratio]{figures/icons/qwen.pdf}}}\hspace{0.3em}Qwen-3.5-9B} & A--B & 15/90 & 8 & 7 \\
 & B--C & 7/90 & 5 & 5 \\
 & C--D & 12/88 & 7 & 7 \\
\bottomrule
\end{tabularx}
\caption{Initial ten-model study. Changed responses include up to three observations of one target. The last column counts changed targets with at least two different valid labels among the three unchanged repeats in either contrasted condition; it is a subset of the preceding column. Each model--contrast comparison contains 30 distinct targets.}
\label{tab:request-distinct}
\end{table}
\endgroup
\clearpage

\section{Majority Voting from Recorded Answers}
\label{app:voting}
\begingroup
\subsection{Retrospective design}
This retrospective analysis uses completed follow-up responses, with rules fixed before computing these summaries. The contemporaneous 30-target control and additional 93 targets remain separate; no new calls or annotations are added.

Let $S$ be a selected subset of $k$ recorded responses for one target. Strict-majority voting returns label $\ell$ only when more than $k/2$ of its response slots contain $\ell$. Otherwise it returns no decision. Invalid responses supply no label but remain in the denominator $k$, and a missing voted decision counts as incorrect. Gold labels score the final decision; they do not choose subsets or break ties. 

Fixed vote-3 averages all $\binom{12}{3}=220$ subsets from twelve B responses. Diverse vote-3 averages $\binom{4}{3}\,3^3=108$ subsets, choosing three different conditions and one repeat from each. Every response has the same inclusion probability within its arm. Single averages its twelve constituent responses. The sensitivity analysis additionally averages all 220 unconstrained triples and all $\binom{12}{5}=792$ five-response subsets; those can contain several responses from one condition. These summaries describe expected performance over subsets of the recorded finite panel.

\subsection{Paired voting gains}
We first average subset outcomes within each target, then bootstrap whole contracts 5,000 times (seed 20261007). Each resample preserves the compared strategies and all responses for a target. Intervals use the 2.5th and 97.5th percentiles with linear interpolation, are descriptive, and are unadjusted for multiple comparisons. Subsets are not independent contracts; intervals degenerate when all observed target differences are zero.
\begin{table}[H]

\centering\small
\setlength{\tabcolsep}{4pt}
\begin{tabularx}{\textwidth}{l l c c >{\centering\arraybackslash}X c}
\toprule
\rowcolor{tableHeader}
\textbf{Model} & Arm & Single & Vote-3 & $\Delta$ [95\% interval] & Coverage \\
\midrule
\multirow{2}{*}{\makebox[0.9em][c]{\raisebox{-0.12em}{\includegraphics[width=0.9em,height=0.9em,keepaspectratio]{figures/icons/jev.pdf}}}\hspace{0.3em}Jev-1.13.0} & Fixed & 80.56 & 80.15 & -0.40 [-1.21, +0.00] & 100.00 \\
 & Varied & 79.17 & 80.00 & +0.83 [+0.00, +2.50] & 100.00 \\
\modelsep
\multirow{2}{*}{\makebox[0.9em][c]{\raisebox{-0.12em}{\includegraphics[width=0.9em,height=0.9em,keepaspectratio]{figures/icons/gemini.pdf}}}\hspace{0.3em}Gemini-3.5-Flash} & Fixed & 83.33 & 83.33 & +0.00 [+0.00, +0.00] & 100.00 \\
 & Varied & 83.33 & 84.44 & +1.11 [-0.56, +3.33] & 100.00 \\
\modelsep
\multirow{2}{*}{\makebox[0.9em][c]{\raisebox{-0.12em}{\includegraphics[width=0.9em,height=0.9em,keepaspectratio]{figures/icons/openai.pdf}}}\hspace{0.3em}GPT-5.6-Terra} & Fixed & 81.67 & 81.76 & +0.09 [-1.38, +1.47] & 100.00 \\
 & Varied & 77.78 & 77.96 & +0.19 [-1.76, +2.22] & 100.00 \\
\modelsep
\multirow{2}{*}{\makebox[0.9em][c]{\raisebox{-0.12em}{\includegraphics[width=0.9em,height=0.9em,keepaspectratio]{figures/icons/claude.pdf}}}\hspace{0.3em}Claude-Sonnet-5} & Fixed & 82.50 & 82.91 & +0.41 [+0.00, +1.23] & 100.00 \\
 & Varied & 85.28 & 85.86 & +0.59 [-1.98, +3.09] & 100.00 \\
\bottomrule
\end{tabularx}
\caption{Voting on the contemporaneous 30-target control. Accuracy and coverage are percentages; $\Delta$ is vote accuracy minus the corresponding single-call mean, in percentage points. Coverage counts subsets returning a strict-majority label. Descriptive paired intervals resample the 30 contracts, not individual subsets.}
\label{tab:voting-control-effects}
\end{table}
\begin{table}[H]

\centering\small
\setlength{\tabcolsep}{4pt}
\begin{tabularx}{\textwidth}{X l c c c c c}
\toprule
\rowcolor{tableHeader}
\textbf{Model} & Arm & Corrections & Regressions & Gain & Loss & Equal \\
\midrule
\multirow{2}{*}{\makebox[0.9em][c]{\raisebox{-0.12em}{\includegraphics[width=0.9em,height=0.9em,keepaspectratio]{figures/icons/jev.pdf}}}\hspace{0.3em}Jev-1.13.0} & Fixed & 0.05 & 0.45 & 0 & 1 & 29 \\
 & Varied & 0.83 & 0.00 & 1 & 0 & 29 \\
\modelsep
\multirow{2}{*}{\makebox[0.9em][c]{\raisebox{-0.12em}{\includegraphics[width=0.9em,height=0.9em,keepaspectratio]{figures/icons/gemini.pdf}}}\hspace{0.3em}Gemini-3.5-Flash} & Fixed & 0.00 & 0.00 & 0 & 0 & 30 \\
 & Varied & 1.76 & 0.65 & 2 & 1 & 27 \\
\modelsep
\multirow{2}{*}{\makebox[0.9em][c]{\raisebox{-0.12em}{\includegraphics[width=0.9em,height=0.9em,keepaspectratio]{figures/icons/openai.pdf}}}\hspace{0.3em}GPT-5.6-Terra} & Fixed & 2.04 & 1.95 & 3 & 2 & 25 \\
 & Varied & 2.28 & 2.10 & 3 & 3 & 24 \\
\modelsep
\multirow{2}{*}{\makebox[0.9em][c]{\raisebox{-0.12em}{\includegraphics[width=0.9em,height=0.9em,keepaspectratio]{figures/icons/claude.pdf}}}\hspace{0.3em}Claude-Sonnet-5} & Fixed & 0.55 & 0.14 & 1 & 0 & 29 \\
 & Varied & 2.62 & 2.04 & 3 & 2 & 25 \\
\bottomrule
\end{tabularx}
\caption{What three-call voting changes on 30 targets. Corrections and regressions are expected percentages relative to a uniformly selected constituent call, averaged across all subsets and targets; their difference equals the accuracy gain. Gain/Loss/Equal count targets with positive/negative/zero change in expected accuracy. They count targets once, not the number of subsets.}
\label{tab:voting-transitions}
\end{table}

Relative to a uniformly selected constituent response, voting corrects a wrong answer or regresses from a correct answer. Their expected difference equals the accuracy gain (Table~\ref{tab:voting-transitions}). Terra's Fixed voting corrects 2.04\% and regresses 1.95\%, leaving only a 0.09-point gain.

\subsection{Resource accounting and sensitivity}
\begin{table}[H]

\centering\small
\setlength{\tabcolsep}{4pt}
\begin{tabularx}{\textwidth}{X l c c c}
\toprule
\rowcolor{tableHeader}
\textbf{Model} & Arm & Single & Vote-3 & Pooled vote-5 \\
\midrule
\multirow{2}{*}{\makebox[0.9em][c]{\raisebox{-0.12em}{\includegraphics[width=0.9em,height=0.9em,keepaspectratio]{figures/icons/jev.pdf}}}\hspace{0.3em}Jev-1.13.0} & Fixed & 0.000134 & 0.000403 & 0.000671 \\
 & Varied & 0.000196 & 0.000587 & 0.000979 \\
\modelsep
\multirow{2}{*}{\makebox[0.9em][c]{\raisebox{-0.12em}{\includegraphics[width=0.9em,height=0.9em,keepaspectratio]{figures/icons/gemini.pdf}}}\hspace{0.3em}Gemini-3.5-Flash} & Fixed & 0.000810 & 0.002430 & 0.004049 \\
 & Varied & 0.001052 & 0.003155 & 0.005259 \\
\modelsep
\multirow{2}{*}{\makebox[0.9em][c]{\raisebox{-0.12em}{\includegraphics[width=0.9em,height=0.9em,keepaspectratio]{figures/icons/openai.pdf}}}\hspace{0.3em}GPT-5.6-Terra} & Fixed & 0.002866 & 0.008599 & 0.014331 \\
 & Varied & 0.006736 & 0.020207 & 0.033679 \\
\modelsep
\multirow{2}{*}{\makebox[0.9em][c]{\raisebox{-0.12em}{\includegraphics[width=0.9em,height=0.9em,keepaspectratio]{figures/icons/claude.pdf}}}\hspace{0.3em}Claude-Sonnet-5} & Fixed & 0.010960 & 0.032881 & 0.054802 \\
 & Varied & 0.011693 & 0.035080 & 0.058467 \\
\bottomrule
\end{tabularx}
\caption{Expected inference cost (USD per target decision; lower is better) in the fresh 30-target control, using recorded reference costs. The cost of every selected call is included, including auxiliary labels requested by C/D; these labels receive no cost credit here. Within each arm, uniform inclusion makes three- and five-call costs exactly three and five times the single-call mean. Fixed and Varied have different workloads and costs. These values concern target requests, not the original joint baseline.}
\label{tab:voting-cost}
\end{table}
Expected voting cost sums all selected calls. Uniform inclusion makes a $k$-call strategy cost $k$ times its arm's single-call mean. Costs differ across arms; auxiliary C/D labels receive no cost credit. Recorded durations do not measure parallel latency or throughput.
All 2,880 control calls are valid, and no control subset abstains. On the additional contracts, primary results retain nine service failures; the supplementary view uses their recorded retries (Appendix~\ref{app:request-controls}). Unknown original charges prevent complete voting-policy costs for affected models, including after successful retries.
\begin{table}[H]

\centering\small
\setlength{\tabcolsep}{4pt}
\begin{tabularx}{\textwidth}{l l c >{\centering\arraybackslash}X c >{\centering\arraybackslash}X}
\toprule
\rowcolor{tableHeader}
\textbf{Model} & Arm & Vote-3 & $\Delta$ [95\% interval] & Vote-5 & $\Delta$ [95\% interval] \\
\midrule
\multirow{2}{*}{\makebox[0.9em][c]{\raisebox{-0.12em}{\includegraphics[width=0.9em,height=0.9em,keepaspectratio]{figures/icons/jev.pdf}}}\hspace{0.3em}Jev-1.13.0} & Fixed & 80.15 & -0.40 [-1.21, +0.00] & 80.00 & -0.56 [-1.67, +0.00] \\
 & Varied & 79.58 & +0.41 [+0.00, +1.23] & 79.85 & +0.68 [+0.00, +2.05] \\
\modelsep
\multirow{2}{*}{\makebox[0.9em][c]{\raisebox{-0.12em}{\includegraphics[width=0.9em,height=0.9em,keepaspectratio]{figures/icons/gemini.pdf}}}\hspace{0.3em}Gemini-3.5-Flash} & Fixed & 83.33 & +0.00 [+0.00, +0.00] & 83.33 & +0.00 [+0.00, +0.00] \\
 & Varied & 83.97 & +0.64 [-0.35, +1.86] & 84.22 & +0.88 [-0.71, +2.80] \\
\modelsep
\multirow{2}{*}{\makebox[0.9em][c]{\raisebox{-0.12em}{\includegraphics[width=0.9em,height=0.9em,keepaspectratio]{figures/icons/openai.pdf}}}\hspace{0.3em}GPT-5.6-Terra} & Fixed & 81.76 & +0.09 [-1.38, +1.47] & 81.94 & +0.28 [-2.37, +2.78] \\
 & Varied & 77.82 & +0.04 [-1.78, +1.86] & 77.98 & +0.20 [-2.47, +2.83] \\
\modelsep
\multirow{2}{*}{\makebox[0.9em][c]{\raisebox{-0.12em}{\includegraphics[width=0.9em,height=0.9em,keepaspectratio]{figures/icons/claude.pdf}}}\hspace{0.3em}Claude-Sonnet-5} & Fixed & 82.91 & +0.41 [+0.00, +1.23] & 83.18 & +0.68 [+0.00, +2.05] \\
 & Varied & 85.79 & +0.51 [-0.99, +2.05] & 85.88 & +0.61 [-1.74, +3.01] \\
\bottomrule
\end{tabularx}
\caption{Unconstrained voting sensitivity analysis on 30 targets. Unlike the primary diverse-three rule, every size-three or size-five subset is eligible, including multiple responses from the same condition. Entries give accuracy (\%) and its change from that arm\textquotesingle s single-call mean (percentage points).}
\label{tab:voting-pooled}
\end{table}
\begin{table}[H]

\centering\small
\setlength{\tabcolsep}{4pt}
\begin{tabularx}{\textwidth}{l l c c >{\centering\arraybackslash}X c}
\toprule
\rowcolor{tableHeader}
\textbf{Model} & View & Single & Diverse-3 & $\Delta$ [95\% interval] & Pooled-5 \\
\midrule
\makebox[0.9em][c]{\raisebox{-0.12em}{\includegraphics[width=0.9em,height=0.9em,keepaspectratio]{figures/icons/jev.pdf}}}\hspace{0.3em}Jev-1.13.0 & First & 75.27 & 74.25 & -1.02 [-2.24, +0.15] & 74.29 \\
\modelsep
\multirow{2}{*}{\makebox[0.9em][c]{\raisebox{-0.12em}{\includegraphics[width=0.9em,height=0.9em,keepaspectratio]{figures/icons/gemini.pdf}}}\hspace{0.3em}Gemini-3.5-Flash} & First & 80.65 & 81.00 & +0.36 [-1.16, +1.88] & 80.93 \\
 & Retry & 80.82 & 81.00 & +0.18 [-1.34, +1.70] & 80.93 \\
\modelsep
\multirow{2}{*}{\makebox[0.9em][c]{\raisebox{-0.12em}{\includegraphics[width=0.9em,height=0.9em,keepaspectratio]{figures/icons/openai.pdf}}}\hspace{0.3em}GPT-5.6-Terra} & First & 80.02 & 79.37 & -0.65 [-1.88, +0.52] & 79.25 \\
 & Retry & 80.47 & 79.49 & -0.98 [-2.14, +0.14] & 79.25 \\
\modelsep
\multirow{2}{*}{\makebox[0.9em][c]{\raisebox{-0.12em}{\includegraphics[width=0.9em,height=0.9em,keepaspectratio]{figures/icons/claude.pdf}}}\hspace{0.3em}Claude-Sonnet-5} & First & 84.68 & 84.05 & -0.63 [-1.91, +0.63] & 84.06 \\
 & Retry & 84.77 & 84.05 & -0.72 [-2.00, +0.54] & 84.06 \\
\bottomrule
\end{tabularx}
\caption{Voting on the additional 93 targets. First uses original attempts; Retry applies only the nine already recorded service retries. Missing answers receive no vote but retain their slot in the majority denominator. Accuracy is in percent and $\Delta$ in percentage points. None of the diverse-three gain intervals excludes zero. Unknown original failure charges prevent a complete cost estimate for the affected models in either view.}
\label{tab:voting-expansion}
\end{table}

\endgroup

\section{Error Profiles and Contract Examples}
\label{app:diagnostics}
\begingroup
\subsection{Gold classes and persistent errors}
The following diagnostics use first attempts on the additional 93 targets. Each contract supplies one target and twelve response slots. Class accuracy averages correctness over those slots. Class-macro accuracy first computes each gold class's accuracy and then weights the three classes equally; it is distinct from Macro-F1 in the original baseline. All-correct and Same-wrong counts require twelve valid predictions. Wrong or invalid response slots remain in the error denominator.
\begin{table}[H]

\centering\small
\setlength{\tabcolsep}{4pt}
\begin{tabularx}{\textwidth}{X l c c c c}
\toprule
\rowcolor{tableHeader}
\textbf{Model} & Gold class & Targets & Accuracy $\uparrow$ & All correct $\uparrow$ & Same wrong $\downarrow$ \\
\midrule
\multirow{3}{*}{\makebox[0.9em][c]{\raisebox{-0.12em}{\includegraphics[width=0.9em,height=0.9em,keepaspectratio]{figures/icons/jev.pdf}}}\hspace{0.3em}Jev-1.13.0} & Entailment & 38 & 86.84 & 31 & 3 \\
 & Contradiction & 13 & 92.31 & 12 & 1 \\
 & Not mentioned & 42 & 59.52 & 24 & 13 \\
\modelsep
\multirow{3}{*}{\makebox[0.9em][c]{\raisebox{-0.12em}{\includegraphics[width=0.9em,height=0.9em,keepaspectratio]{figures/icons/gemini.pdf}}}\hspace{0.3em}Gemini-3.5-Flash} & Entailment & 38 & 89.69 & 32 & 3 \\
 & Contradiction & 13 & 77.56 & 8 & 1 \\
 & Not mentioned & 42 & 73.41 & 26 & 7 \\
\modelsep
\multirow{3}{*}{\makebox[0.9em][c]{\raisebox{-0.12em}{\includegraphics[width=0.9em,height=0.9em,keepaspectratio]{figures/icons/openai.pdf}}}\hspace{0.3em}GPT-5.6-Terra} & Entailment & 38 & 84.21 & 31 & 3 \\
 & Contradiction & 13 & 84.62 & 10 & 1 \\
 & Not mentioned & 42 & 74.80 & 23 & 4 \\
\modelsep
\multirow{3}{*}{\makebox[0.9em][c]{\raisebox{-0.12em}{\includegraphics[width=0.9em,height=0.9em,keepaspectratio]{figures/icons/claude.pdf}}}\hspace{0.3em}Claude-Sonnet-5} & Entailment & 38 & 90.79 & 33 & 1 \\
 & Contradiction & 13 & 66.03 & 5 & 1 \\
 & Not mentioned & 42 & 84.92 & 32 & 4 \\
\bottomrule
\end{tabularx}
\caption{Gold-class diagnostics on the additional 93 targets, with first-attempt failures retained. Accuracy averages twelve responses per target. All correct and Same wrong count targets. The class distribution is 38 entailments, 13 contradictions, and 42 not-mentioned labels.}
\label{tab:diagnostic-classes}
\end{table}
\begin{table}[H]

\centering\small
\setlength{\tabcolsep}{4pt}
\begin{tabularx}{\textwidth}{X c c c c c}
\toprule
\rowcolor{tableHeader}
\textbf{Model} & Accuracy & Class macro & Same wrong & Error slots & Share (\%) \\
\midrule
\makebox[0.9em][c]{\raisebox{-0.12em}{\includegraphics[width=0.9em,height=0.9em,keepaspectratio]{figures/icons/jev.pdf}}}\hspace{0.3em}Jev-1.13.0 & 75.27 & 79.56 & 17 & 276 & 73.91 \\
\makebox[0.9em][c]{\raisebox{-0.12em}{\includegraphics[width=0.9em,height=0.9em,keepaspectratio]{figures/icons/gemini.pdf}}}\hspace{0.3em}Gemini-3.5-Flash & 80.65 & 80.22 & 11 & 216 & 61.11 \\
\makebox[0.9em][c]{\raisebox{-0.12em}{\includegraphics[width=0.9em,height=0.9em,keepaspectratio]{figures/icons/openai.pdf}}}\hspace{0.3em}GPT-5.6-Terra & 80.02 & 81.21 & 8 & 223 & 43.05 \\
\makebox[0.9em][c]{\raisebox{-0.12em}{\includegraphics[width=0.9em,height=0.9em,keepaspectratio]{figures/icons/claude.pdf}}}\hspace{0.3em}Claude-Sonnet-5 & 84.68 & 80.58 & 6 & 171 & 42.11 \\
\bottomrule
\end{tabularx}
\caption{Persistent-error concentration on 93 targets. Accuracy and Class macro are percentages; the latter weights the three gold classes equally. Same wrong counts targets, Error slots counts wrong or invalid answers among 1,116 attempts, and Share is $12\times\mathrm{Same\ wrong}/\mathrm{Error\ slots}$. Repeated errors on one target are not treated as independent contracts.}
\label{tab:diagnostic-error-share}
\end{table}
\begin{table}[H]

\centering\small
\setlength{\tabcolsep}{4pt}
\begin{tabularx}{\textwidth}{X l l c}
\toprule
\rowcolor{tableHeader}
\textbf{Model} & Gold & Repeated prediction & Targets \\
\midrule
\multirow[c]{4}{*}{\makebox[0.9em][c]{\raisebox{-0.12em}{\includegraphics[width=0.9em,height=0.9em,keepaspectratio]{figures/icons/jev.pdf}}}\hspace{0.3em}Jev-1.13.0} & C & E & 1 \\
 & E & C & 3 \\
 & N & C & 8 \\
 & N & E & 5 \\
\modelsep
\multirow[c]{4}{*}{\makebox[0.9em][c]{\raisebox{-0.12em}{\includegraphics[width=0.9em,height=0.9em,keepaspectratio]{figures/icons/gemini.pdf}}}\hspace{0.3em}Gemini-3.5-Flash} & C & E & 1 \\
 & E & C & 3 \\
 & N & C & 1 \\
 & N & E & 6 \\
\modelsep
\multirow[c]{5}{*}{\makebox[0.9em][c]{\raisebox{-0.12em}{\includegraphics[width=0.9em,height=0.9em,keepaspectratio]{figures/icons/openai.pdf}}}\hspace{0.3em}GPT-5.6-Terra} & C & E & 1 \\
 & E & C & 2 \\
 & E & N & 1 \\
 & N & C & 1 \\
 & N & E & 3 \\
\modelsep
\multirow[c]{3}{*}{\makebox[0.9em][c]{\raisebox{-0.12em}{\includegraphics[width=0.9em,height=0.9em,keepaspectratio]{figures/icons/claude.pdf}}}\hspace{0.3em}Claude-Sonnet-5} & C & N & 1 \\
 & E & C & 1 \\
 & N & E & 4 \\
\bottomrule
\end{tabularx}
\caption{Confusions among same-wrong targets in the additional sample. E: entailment; C: contradiction; N: not mentioned. Each contract is counted once. These predictions are wrong on all twelve requests; they cannot be corrected by voting over subsets of these recorded answers.}
\label{tab:diagnostic-confusion}
\end{table}

The gold-label profile changes the relative ordering of class-macro and ordinary mean accuracy. This comparison identifies a dependency on the class mixture, not a preferred deployment distribution. The persistent-error share uses $12$ error slots per same-wrong target; it measures the concentration of observed errors rather than providing twelve independent pieces of evidence about that target.

\subsection{Hypothesis types and contract length}
\begin{table}[H]

\centering\small
\setlength{\tabcolsep}{4pt}
\begin{tabularx}{\textwidth}{l c >{\centering\arraybackslash}X >{\centering\arraybackslash}X >{\centering\arraybackslash}X >{\centering\arraybackslash}X}
\toprule
\rowcolor{tableHeader}
\textbf{Hypothesis} & $n$ & \makebox[0.9em][c]{\raisebox{-0.12em}{\includegraphics[width=0.9em,height=0.9em,keepaspectratio]{figures/icons/jev.pdf}}}\hspace{0.3em}Jev-1.13.0 & \makebox[0.9em][c]{\raisebox{-0.12em}{\includegraphics[width=0.9em,height=0.9em,keepaspectratio]{figures/icons/gemini.pdf}}}\hspace{0.3em}Gemini-3.5-Flash & \makebox[0.9em][c]{\raisebox{-0.12em}{\includegraphics[width=0.9em,height=0.9em,keepaspectratio]{figures/icons/openai.pdf}}}\hspace{0.3em}GPT-5.6-Terra & \makebox[0.9em][c]{\raisebox{-0.12em}{\includegraphics[width=0.9em,height=0.9em,keepaspectratio]{figures/icons/claude.pdf}}}\hspace{0.3em}Claude-Sonnet-5 \\
\midrule
\texttt{nda-1} & 9 & 39.81 / 3 & 37.04 / 3 & 53.70 / 1 & 76.85 / 1 \\
\texttt{nda-10} & 4 & 100.00 / 0 & 100.00 / 0 & 70.83 / 0 & 100.00 / 0 \\
\texttt{nda-11} & 6 & 100.00 / 0 & 100.00 / 0 & 100.00 / 0 & 100.00 / 0 \\
\texttt{nda-12} & 10 & 100.00 / 0 & 100.00 / 0 & 92.50 / 0 & 100.00 / 0 \\
\texttt{nda-13} & 3 & 72.22 / 0 & 94.44 / 0 & 100.00 / 0 & 100.00 / 0 \\
\texttt{nda-15} & 6 & 15.28 / 4 & 91.67 / 0 & 80.56 / 0 & 87.50 / 0 \\
\texttt{nda-16} & 5 & 61.67 / 1 & 48.33 / 2 & 50.00 / 1 & 55.00 / 2 \\
\texttt{nda-17} & 6 & 37.50 / 3 & 22.22 / 3 & 55.56 / 2 & 52.78 / 1 \\
\texttt{nda-18} & 7 & 100.00 / 0 & 100.00 / 0 & 100.00 / 0 & 98.81 / 0 \\
\texttt{nda-19} & 4 & 100.00 / 0 & 100.00 / 0 & 100.00 / 0 & 100.00 / 0 \\
\texttt{nda-2} & 7 & 85.71 / 1 & 85.71 / 1 & 85.71 / 1 & 82.14 / 0 \\
\texttt{nda-20} & 6 & 50.00 / 3 & 79.17 / 0 & 62.50 / 1 & 69.44 / 1 \\
\texttt{nda-3} & 2 & 100.00 / 0 & 100.00 / 0 & 95.83 / 0 & 66.67 / 0 \\
\texttt{nda-4} & 6 & 83.33 / 1 & 83.33 / 1 & 83.33 / 1 & 83.33 / 1 \\
\texttt{nda-5} & 6 & 83.33 / 1 & 83.33 / 1 & 83.33 / 1 & 87.50 / 0 \\
\texttt{nda-7} & 3 & 100.00 / 0 & 94.44 / 0 & 72.22 / 0 & 75.00 / 0 \\
\texttt{nda-8} & 3 & 100.00 / 0 & 100.00 / 0 & 100.00 / 0 & 100.00 / 0 \\
\bottomrule
\end{tabularx}
\caption{Hypothesis-specific diagnostics on 93 targets. Each model cell gives mean target accuracy (\%) / number of same-wrong targets. The 17 groups contain only 2--10 contracts each; all groups and denominators are shown. Hypothesis texts appear in Table~\ref{tab:diagnostic-hypothesis-texts}.}
\label{tab:diagnostic-hypotheses}
\end{table}
\begin{table}[H]

\centering\small
\setlength{\tabcolsep}{4pt}
\begin{tabularx}{\textwidth}{l X}
\toprule
\rowcolor{tableHeader}
\textbf{ID} & Original hypothesis \\
\midrule
\texttt{nda-1} & All Confidential Information shall be expressly identified by the Disclosing Party. \\
\texttt{nda-10} & Receiving Party shall not disclose the fact that Agreement was agreed or negotiated. \\
\texttt{nda-11} & Receiving Party shall not reverse engineer any objects which embody Disclosing Party's Confidential Information. \\
\texttt{nda-12} & Receiving Party may independently develop information similar to Confidential Information. \\
\texttt{nda-13} & Receiving Party may acquire information similar to Confidential Information from a third party. \\
\texttt{nda-15} & Agreement shall not grant Receiving Party any right to Confidential Information. \\
\texttt{nda-16} & Receiving Party shall destroy or return some Confidential Information upon the termination of Agreement. \\
\texttt{nda-17} & Receiving Party may create a copy of some Confidential Information in some circumstances. \\
\texttt{nda-18} & Receiving Party shall not solicit some of Disclosing Party's representatives. \\
\texttt{nda-19} & Some obligations of Agreement may survive termination of Agreement. \\
\texttt{nda-2} & Confidential Information shall only include technical information. \\
\texttt{nda-20} & Receiving Party may retain some Confidential Information even after the return or destruction of Confidential Information. \\
\texttt{nda-3} & Confidential Information may include verbally conveyed information. \\
\texttt{nda-4} & Receiving Party shall not use any Confidential Information for any purpose other than the purposes stated in Agreement. \\
\texttt{nda-5} & Receiving Party may share some Confidential Information with some of Receiving Party's employees. \\
\texttt{nda-7} & Receiving Party may share some Confidential Information with some third-parties (including consultants, agents and professional advisors). \\
\texttt{nda-8} & Receiving Party shall notify Disclosing Party in case Receiving Party is required by law, regulation or judicial process to disclose any Confidential Information. \\
\bottomrule
\end{tabularx}
\caption{Original ContractNLI hypothesis texts, unchanged from the released dataset.}
\label{tab:diagnostic-hypothesis-texts}
\end{table}
\begin{table}[H]

\centering\small
\setlength{\tabcolsep}{4pt}
\begin{tabularx}{\textwidth}{X l c c c}
\toprule
\rowcolor{tableHeader}
\textbf{Model} & Length group & Accuracy $\uparrow$ & All correct $\uparrow$ & Same wrong $\downarrow$ \\
\midrule
\multirow{3}{*}{\makebox[0.9em][c]{\raisebox{-0.12em}{\includegraphics[width=0.9em,height=0.9em,keepaspectratio]{figures/icons/jev.pdf}}}\hspace{0.3em}Jev-1.13.0} & Short & 81.18 & 24 & 5 \\
 & Middle & 75.00 & 22 & 5 \\
 & Long & 69.62 & 21 & 7 \\
\modelsep
\multirow{3}{*}{\makebox[0.9em][c]{\raisebox{-0.12em}{\includegraphics[width=0.9em,height=0.9em,keepaspectratio]{figures/icons/gemini.pdf}}}\hspace{0.3em}Gemini-3.5-Flash} & Short & 84.95 & 23 & 2 \\
 & Middle & 73.66 & 20 & 5 \\
 & Long & 83.33 & 23 & 4 \\
\modelsep
\multirow{3}{*}{\makebox[0.9em][c]{\raisebox{-0.12em}{\includegraphics[width=0.9em,height=0.9em,keepaspectratio]{figures/icons/openai.pdf}}}\hspace{0.3em}GPT-5.6-Terra} & Short & 79.30 & 20 & 3 \\
 & Middle & 83.33 & 23 & 3 \\
 & Long & 77.42 & 21 & 2 \\
\modelsep
\multirow{3}{*}{\makebox[0.9em][c]{\raisebox{-0.12em}{\includegraphics[width=0.9em,height=0.9em,keepaspectratio]{figures/icons/claude.pdf}}}\hspace{0.3em}Claude-Sonnet-5} & Short & 87.90 & 23 & 1 \\
 & Middle & 79.84 & 22 & 4 \\
 & Long & 86.29 & 25 & 1 \\
\bottomrule
\end{tabularx}
\caption{Length groups on the additional 93 contracts, with 31 targets each. Whitespace-separated word counts define Short (232--1138), Middle (1150--1754), and Long (1861--6267). The bins are shared by all models. Accuracy is in percent; the other columns count targets. These descriptive groups also differ in hypothesis and gold-label composition.}
\label{tab:diagnostic-lengths}
\end{table}
Contract length is the number of whitespace-separated words in the full unchanged text, rather than a model-specific token count. Sorting by length and then document ID gives three groups of 31 contracts. The same groups are used for all models; their word ranges appear in Table~\ref{tab:diagnostic-lengths}. Hypothesis groups and length groups are descriptive, with no subgroup significance testing. The small hypothesis denominators and different label mixtures do not support attributing an error to a single property of the contract.

\clearpage
\subsection{Illustrative cases}
\label{app:contract-cases}
The cases below use original ContractNLI IDs and annotations. Excerpts are verbatim portions of the input contracts, and predictions reproduce recorded target labels. Cases are selected for illustration after the analysis; they do not estimate prevalence or supply a model's explanation. The companion case file contains the exact target labels and source identifiers.

\paragraph{A correction and a regression hidden by a score.}
In the initial Qwen-3.5-4B A--B comparison, four corrections and four regressions cancel, leaving 68/90 correct responses in each condition despite nine changed labels. Two of those paired responses make the cancellation concrete (repeat index 0 in each contract):
\begin{itemize}[leftmargin=*,itemsep=4pt]
\item Contract 21, \texttt{nda-13}: ``Receiving Party may acquire information similar to Confidential Information from a third party.'' Its gold label is entailment. The contract includes the exclusion ``which is lawfully obtained by the Receiving Party from a third party and is not subject to a similar restriction on disclosure;''. Qwen changes from not mentioned in A to entailment in B, correcting its answer.
\item Contract 408, \texttt{nda-4}: ``Receiving Party shall not use any Confidential Information for any purpose other than the purposes stated in Agreement.'' Its gold label is entailment. The contract states that the parties ``will not use or disclose in any way any of such Confidential Information without the prior express written consent of the Buyer Parties.'' Qwen changes from entailment in A to not mentioned in B, losing a correct answer.
\end{itemize}
The direction of change differs despite unchanged evidence and target within each pair. Both targets also vary among unchanged repeats, so the examples show observed cancellation rather than assigning it exclusively to the added hypothesis catalog.

\paragraph{A repeated wrong answer that voting preserves.}
Contract 85, \texttt{nda-15}, asks whether ``Agreement shall not grant Receiving Party any right to Confidential Information.'' The released gold label is not mentioned. A use restriction in the contract reads: ``The Receiving Party agrees not to use the Information in any other way than set forth in this Agreement.'' The target concerns the grant of rights, rather than simply the restriction on use; the excerpt identifies the distinction being judged.

Jev returns contradiction on all twelve A/B/C/D requests. Gemini-3.5-Flash and Sonnet each return not mentioned on all twelve. Terra returns not mentioned on eleven first attempts and has one service failure; its recorded retry returns not mentioned. Every three- or five-answer subset of Jev's recorded answers therefore votes for contradiction. Its high agreement on this target supplies no correction of the dataset-scored error.
\endgroup

\end{document}